\documentclass{article}

\usepackage[preprint,nonatbib]{neurips_2022}

\usepackage[utf8]{inputenc} % allow utf-8 input
\usepackage[T1]{fontenc}    % use 8-bit T1 fonts
\usepackage{hyperref}       % hyperlinks
\usepackage{url}            % simple URL typesetting
\usepackage{booktabs}       % professional-quality tables
\usepackage{amsfonts}       % blackboard math symbols
\usepackage{nicefrac}       % compact symbols for 1/2, etc.
\usepackage{microtype}      % microtypography
\usepackage{xcolor}         % colors
\usepackage{graphicx}
\usepackage{caption}
\usepackage{subfigure}
\usepackage{booktabs} % for professional tables
\usepackage{multirow}
\usepackage{wrapfig}
\usepackage{enumitem}
\usepackage{algorithm}
\usepackage[noend]{algpseudocode}
\usepackage{eqnarray}
\usepackage{makecell}
\usepackage{amssymb}
\usepackage{amsmath}
\usepackage{amsthm}
\usepackage{bbm}

\newtheorem{theorem}{Theorem}%[section]

\newcommand{\BK}[1]{\textcolor{red}{BK: #1}}
\newcommand{\JH}[1]{\textcolor{magenta}{JH: #1}}
\newcommand{\AM}[1]{\textcolor{cyan}{AM: #1}}

\usepackage{color}

\newtheoremstyle{TheoremRep}
        {\topsep}{\topsep}              %%% space between body and thm
        {\itshape}                      %%% Thm body font
        {}                              %%% Indent amount (empty = no indent)
        {\bfseries}                     %%% Thm head font
        {.}                             %%% Punctuation after thm head
        { }                             %%% Space after thm head
        {\thmname{#1}\thmnote{ \bfseries #3}}%%% Thm head spec
\theoremstyle{TheoremRep}

\newcommand{\ps}{\ensuremath{P}}

\title{On Certifying and Improving \\ Generalization to Unseen Domains}

\author{Akshay Mehra\textsuperscript{1}, Bhavya Kailkhura\textsuperscript{2}, Pin-Yu Chen\textsuperscript{3} and Jihun Hamm\textsuperscript{1}\\
{\small \textsuperscript{1}Tulane University \quad \textsuperscript{2}Lawrence Livermore National Laboratory \quad \textsuperscript{3}IBM Research}\\ 
{\tt\small\{amehra, jhamm3\}@tulane.edu, kailkhura1@llnl.gov, pin-yu.chen@ibm.com}\\
}

\begin{document}

\maketitle

\begin{abstract}
Domain Generalization (DG) aims to learn models whose performance remains high on unseen domains encountered at test-time by using data from multiple related source domains. 
Many existing DG algorithms reduce the divergence between source distributions in a representation space to potentially align the unseen domain close to the sources. 
This is motivated by the analysis that explains generalization to unseen domains using distributional distance (such as the Wasserstein distance) to the sources.
However, due to the openness of the DG objective, it is challenging to evaluate DG algorithms comprehensively using a few benchmark datasets.
In particular, we demonstrate that the accuracy of the models trained with DG methods varies significantly across unseen domains, generated from popular benchmark datasets.
This highlights that the performance of DG methods on a few benchmark datasets may not be representative of their performance on unseen domains in the wild.
To overcome this roadblock, we propose a universal certification framework based on distributionally robust optimization (DRO) that can efficiently certify the worst-case performance of any DG method. 
This enables a data-independent evaluation of a DG method complementary to the empirical evaluations on benchmark datasets. 
Furthermore, we propose a training algorithm that can be used with any DG method to provably improve their certified performance.
Our empirical evaluation demonstrates the effectiveness of our method at significantly improving the worst-case loss (i.e., reducing the risk of failure of these models in the wild) without incurring a  significant performance drop on benchmark datasets.

\end{abstract}

%%%%%%%%%%%%%%%%%%%%%%%%%%%%%%%%%%%%%%%%%%%%%%%%%%%%%%%%%%%%%%%%%%%%%%%%%%%%%%%%%%%%%%%%%%%%%%%%%%%%%%%%%%%%%%%%%%%%%%%%%%%%%%%%%%%%%%%%%%%%%%%%%%%%%%%%%%%%%%%%%%%%
\section{Introduction}
\begin{figure}[tb]
  \centering
  \subfigure[G2DM \cite{albuquerque2019generalizing}]{\includegraphics[width=0.24\columnwidth]{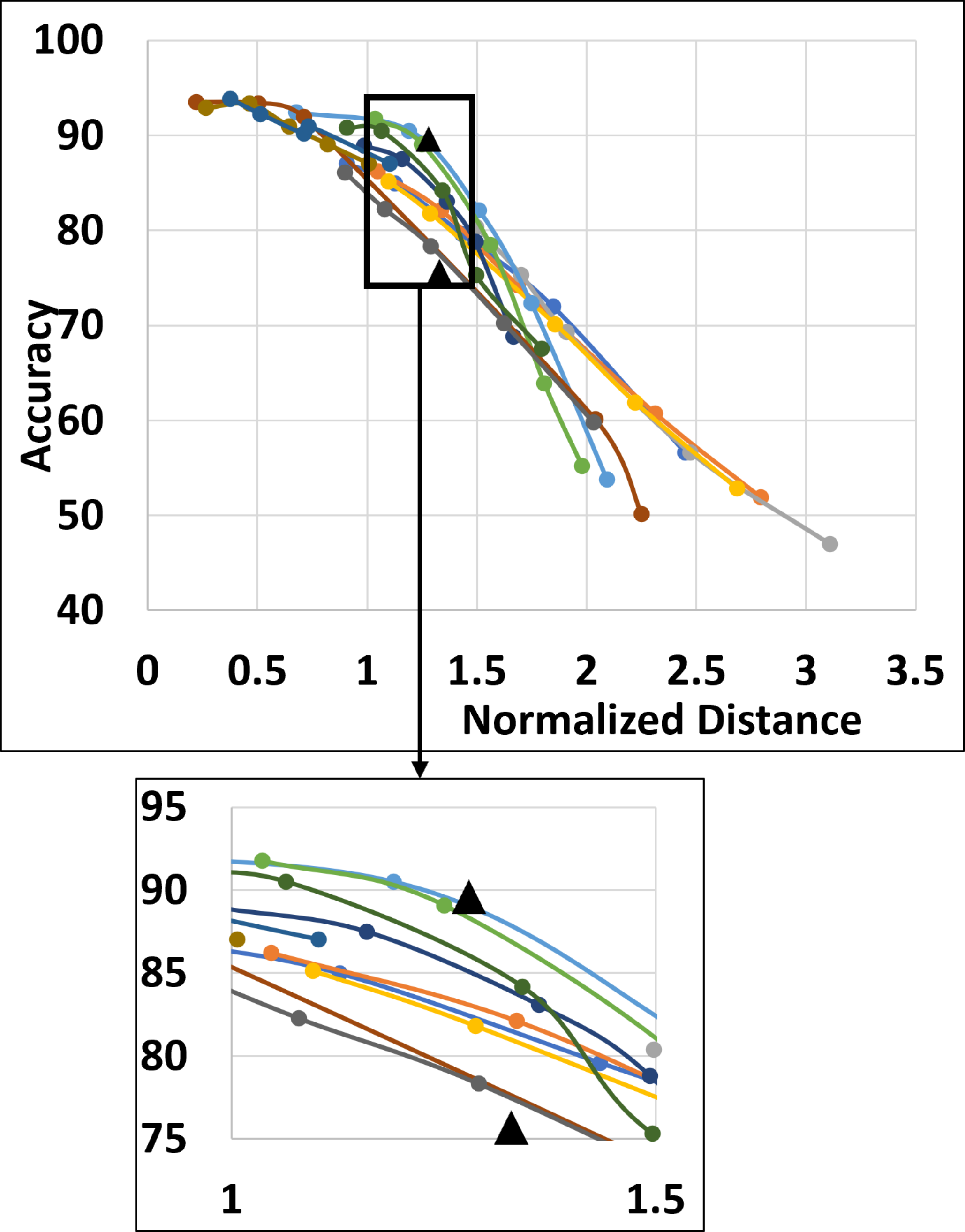}}
  \subfigure[CDAN \cite{long2018conditional}]{\includegraphics[width=0.24\columnwidth]{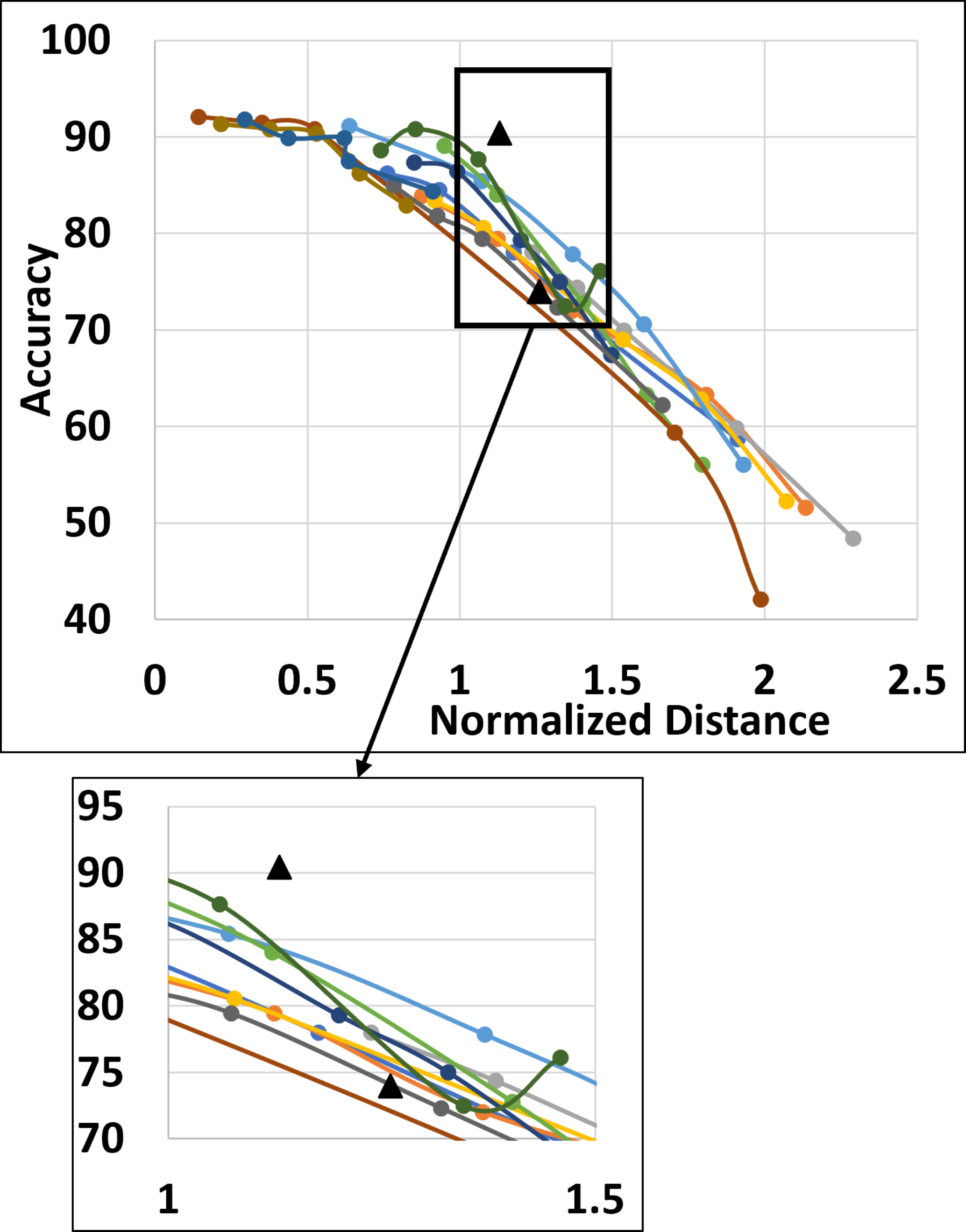}}
  \subfigure[WM (see Sec.~\ref{sec:certification})]{\includegraphics[width=0.24\columnwidth]{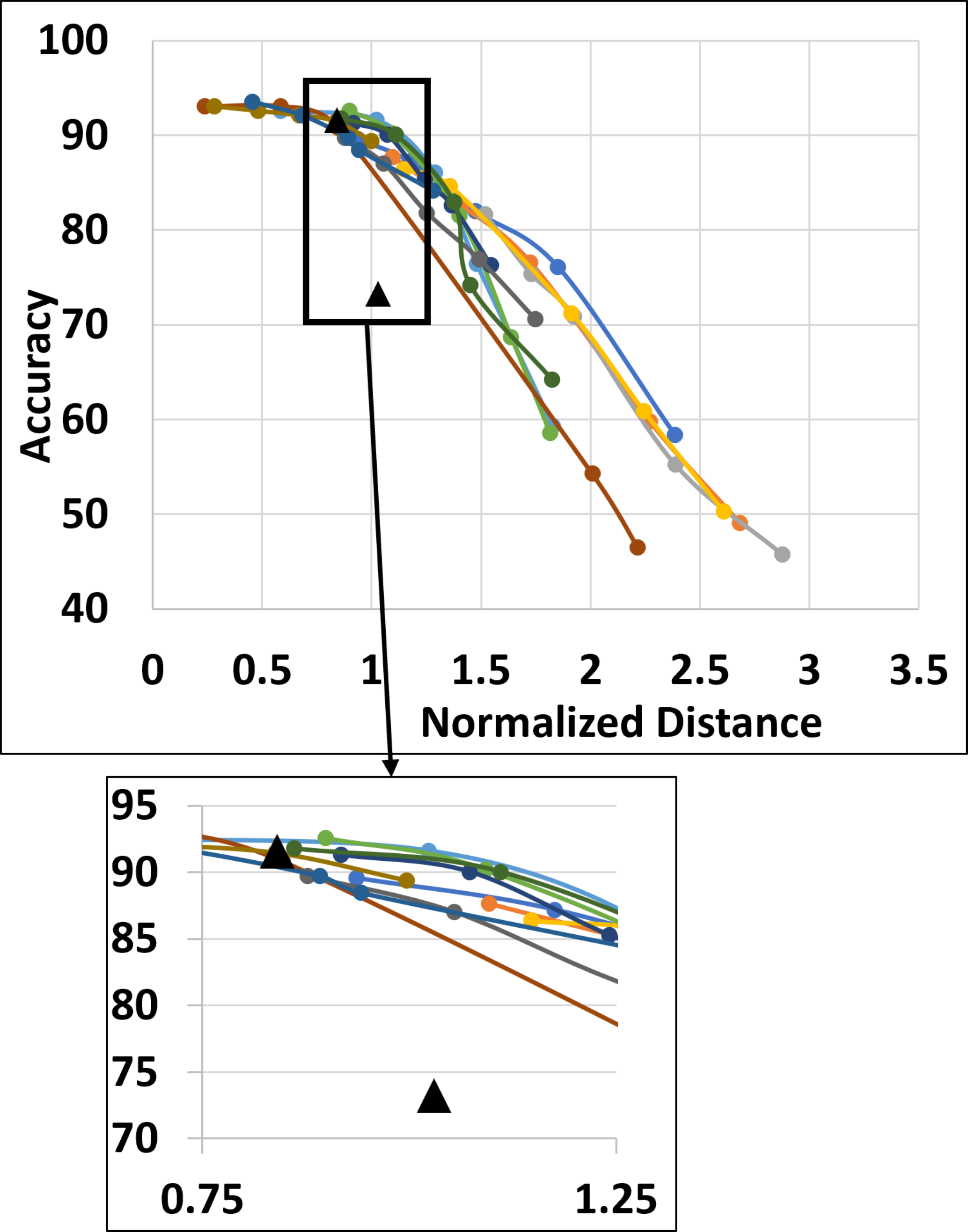}}
  \subfigure[VREX \cite{krueger2021out}]{\includegraphics[width=0.24\columnwidth]{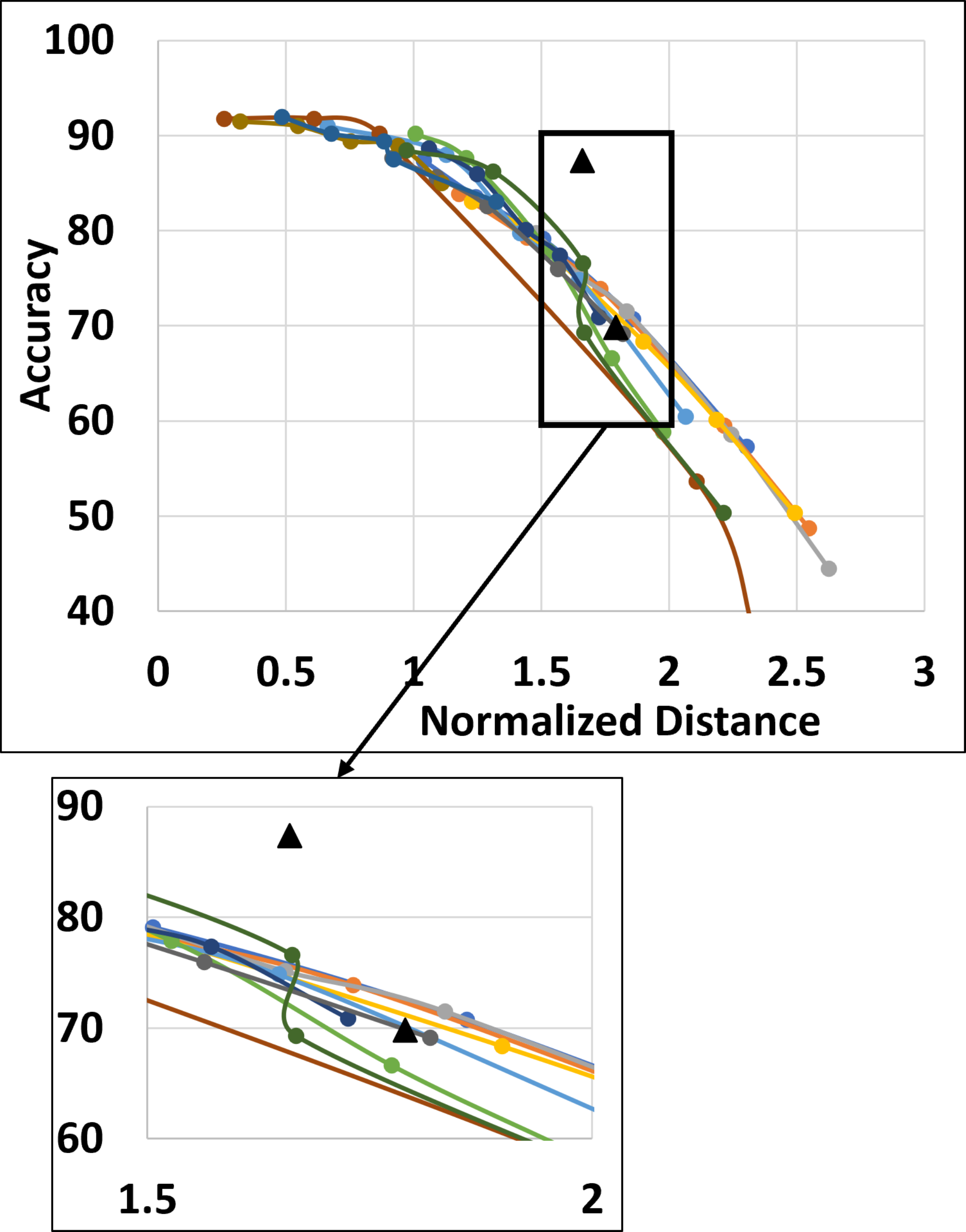}}
   \includegraphics[width=0.75\textwidth]{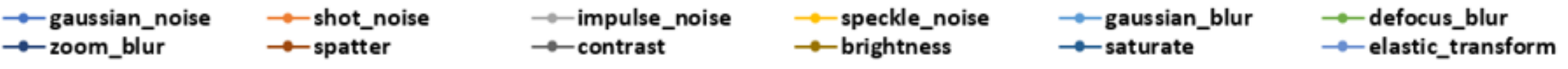}
  \caption{(Best viewed in color.)
   {\bf Hardness of empirically evaluating the generalization performance of DG methods:} 
   The high variability in the accuracy at the same distance from the sources demonstrates that the performance of a DG method on a few domains is not representative of their performance on other unseen domains.
   The triangles denote unseen benchmark distributions (Photos and Art from PACS \cite{li2017deeper} here (left to right)) and lines denote distributions under common corruptions.
   } 
  \label{fig:high_variability_of_dg_a}
\end{figure}

%%%%%%%%%%%%%%%%%%%%%%%%%%%%%%%%%%%%%%%%%%%%%%%%%%%%%%%%%%%%%%%%%%%%%%%%%%%%%%%%%%%%%%%%%%%%%%%%%%%%%%%%%%%%%%%%%%%%%%%%%%%%%%%%%%%%%%%%%%%%%%%%%%%%%%%%%%%%%%%%%%%%
%Machine learning models are being increasingly used in applications where they can encounter data from domains unseen during training.
A major challenge in machine learning is to develop models that can generalize well to domains unseen during training.
%\PYB{One of the ultimate goals in machine learning is to develop models that can generalize well to domains unseen during training.} changed
%training on all variations that a model may encounter at test-time is prohibitive both in terms of time and cost. 
For example, a self-driving car or a drone deployed in the wild can encounter complex spatial or temporal changes such as object variations, camera blur, unseen weather changes, etc., which are not seen during training.
%For example, a self-driving car or a drone deployed in the wild may encounter complex spatial or temporal variations such as weather changes which were not present in the training data.
This domain shift at test-time leads to drastic degradation in the performance of the models and increases the risk of deploying these models in the real world \cite{bulusu2020anomalous, hendrycks2019benchmarking}. 
%the object detection performance of a self-driving car or a drone deployed in wild may drastically degrade due to  .
%\JH{If you want to list examples, there should be more mission-critical examples. Also cite papers.} added
The area of Domain Generalization (DG) \cite{wang2021generalizing} studies this crucial problem and aims to develop models whose performance remains high on unseen domains (or variations). 
%Many works in this area focus on learning a representation space such that features of data from diverse source domains become indistinguishable in this space . This can achieved by using adversarial training with discriminators \cite{ganin2016domain,albuquerque2019generalizing} or explicitly minimizing divergence measures such as the Wasserstein distance between distributions in the representation space.
%\JH{Introduce DG ideas/methods first. } already mentioned in the next line
Analyses in recent works have demonstrated that the distributional distance between the source and the unseen domains is a key metric to predict the generalization performance of the model on unseen domains \cite{sinha2017certifying,kumar2022certifying,kuhn2019wasserstein,gao2016distributionally, diffenderfer2021winning, staib2019distributionally,bental2013robust,weber2022certifying,ben2007analysis,albuquerque2019generalizing}.
%that generalization to unseen domains at test-time is possible if the unseen domain is close to the source domain under different divergence measures such as the Wasserstein distance~ \cite{sinha2017certifying,kumar2022certifying,kuhn2019wasserstein,gao2016distributionally, diffenderfer2021winning}, maximum mean discrepancy~ \cite{staib2019distributionally}, $f$-divergence~ \cite{bental2013robust,weber2022certifying}, $\mathcal{H}$-divergence~ \cite{ben2007analysis,albuquerque2019generalizing}.
To ensure that the unseen domain lies close to the source domains 
%Since generalization to arbitrary unseen domains is not possible 
some works in DG impose additional assumptions \cite{blanchard2011generalizing,albuquerque2019generalizing,krueger2021out,kumar2022certifying,weber2022certifying} on the unseen domains whereas others learn a representation space in which the divergence between multiple source domains can be reduced while ensuring high performance on these domains \cite{albuquerque2019generalizing,ganin2015unsupervised,zhao2018adversarial,zhang2019bridging}. % (we focus on this class of DG methods).
%Since distributions are fixed in the input space 
%\PYB{why? can we say distribution shifts can be measures in input as well as representation spaces?} modified
%, many works in DG have proposed to learn a representation space where the divergence between multiple source domains  can be reduced while ensuring high performance on these domains~ \cite{albuquerque2019generalizing,ganin2015unsupervised,zhao2018adversarial,zhang2019bridging}.
By learning such a representation space, these methods aim to potentially reduce the divergence between source and unseen domains, thereby guaranteeing generalization.
However, due to the openness of the DG goal, it is challenging to evaluate the quality of the representation learned by DG algorithms comprehensively only through empirical evaluations.
Furthermore, it is unclear if the performance of a DG algorithm on benchmark datasets is representative of its performance in the wild.
%However, since models trained with DG methods will be deployed in the wild, we are interested in examining whether the performance of DG methods on benchmark datasets is representative of their performance on unseen domains.
%A caveat with this approach is that an ideal DG model needs to be robust to any valid domain, including the ones unseen in any previous benchmarks.
%Due to the inherent difficulty of DG problem, we critically examine whether the performance of DG methods on benchmark datasets is representative of their performance on unseen domains.
%based on benchmar datasets How \emph{generalizable} are existing DG methods in terms of maintaining high performance achieved on benchmark datasets to equidistance yet truly unseen domains 
%%%%%%%%%To study this, we focus on DG methods that align the distributions in the representation space and learn models trained with two methods (G2DM\cite{albuquerque2019generalizing} and WM (see Sec.~\ref{sec:certification})).
%use them learn a representation space by minimizing divergence between 
%using two source domains from benchmark datasets (R-MNIST \cite{ghifary2015domain} and PACS \cite{li2017deeper}). 
%%%%%%%%%\JH{Don't limit the methods/data yet!!!!!! You have to get more results for the supplementary material. A review will easily reject this paper with the lack of experiments as an excuse.}
In Fig.~\ref{fig:high_variability_of_dg_a},
%and Fig.~\ref{fig:high_variability_of_dg_b} in the Appendix, 
we demonstrate these difficulties by considering a number of possible target domains such as Arts and Photos from PACS \cite{li2017deeper}, having variations in the image style
%, VLCS \cite{fang2013unbiased}), rotation angle (R-MNIST \cite{ghifary2015domain}) 
as well as various corrupted versions \cite{hendrycks2019benchmarking,mu2019mnist} of domains in PACS. We plot the performance of various DG methods versus the distance of these domains from the sources in the representation space.
%To examine this we consider a number of possible target domains such as variations in the image style (PACS \cite{li2017deeper}, VLCS \cite{fang2013unbiased}), rotation angle (R-MNIST \cite{ghifary2015domain}) as well as various corrupted versions \cite{hendrycks2019benchmarking,mu2019mnist} of the target domains, and plot the performance of DG methods w.r.t. the distance of the domains from the sources in the representation space.
%We learn the representation space by minimizing the divergence between two source domains from benchmark datasets (R-MNIST \cite{ghifary2015domain} and PACS \cite{li2017deeper}) and use the remaining domains as unseen domains as well as synthetically crafted new domains using natural corruptions. 
%To evaluate how the performance of DG methods change on unseen domains in relation to their distance from the source domain, in Fig.~\ref{fig:high_variability_of_dg_a} we plot the accuracy of the models on unseen domains w.r.t their distance from the sources in the representation space.
%We learn the representation space by minimizing the divergence between two source domains from benchmark datasets (R-MNIST \cite{ghifary2015domain} and PACS \cite{li2017deeper}) and use the remaining domains as unseen domains as well as synthetically crafted new domains using natural corruptions. 
%plot the accuracy of these models on various unseen domains including those present in benchmark datasets as well as synthetically crafted. 
Although, the accuracy of DG methods degrades quickly as the unseen domain becomes farther from the source (as expected), what is surprising is the high variability in the accuracy of the models even on unseen domains equidistant from the sources (e.g., the accuracy varies from 90\%-75\% at a normalized distance of 1.2 for G2DM \cite{albuquerque2019generalizing} in Fig.~\ref{fig:high_variability_of_dg_a}). 
%Thus, performance on a few benchmark domains is not representative of the model's performance on other unseen domains. % even those at the same distance.% lying at the same distance. 
%on unseen distributions lying at the same distance i.e.  trained with a DG method, has high variability in its accuracy (5-7\%) on unseen domains lying at the same distance from the sources in the representation space. 
Moreover, the performance of a DG method relative to another method is not consistent over different unseen domains, making it difficult to assess the progress of DG as a field
%quite vary significantly across different unseen domains.
%i.e. a method may be better than the other in one domain but have the opposite trend for another domains 
(see Sec.~\ref{sec:corruptions} for a detailed discussion).
%can vary substantially based on distance as well as  , at any given distance, the accuracy of the models trained with a different DG methods can be significantly different.
%This high variability in the accuracy of the models trained with DG methods, 
%makes it hard to assess their generalization performance  
%This
%This highlight that the \emph{performance of a DG method on a few benchmark datasets is not representative of the its
%performance on other unseen domains }
This highlights the \emph{insufficiency of empirical evaluation on benchmark datasets to reliably assess the true generalization performance of DG methods}
%fundamental difficulty in gauging the performance of a DG method on unseen domains solely based on their performance on a few benchmark datasets} 
and motivates the need for a more rigorous and reliable method for evaluation of DG algorithms.

%Let $\epsilon_{adv}$ be the distance from the source to the adv distribution (see text for definition) $\epsilon_{adv}=D(P_S,P_{S,adv})$, for which the accuracy is 0. Two conclusion from this are, 1) The worst-case accuracy beyond $\epsilon_{adv}$ is therefore 0, and 2) any distribution closer than $\epsilon'$ will be no worse than $P_{S,adv}$, that is $P_{S,adv}$ is the closest absolute worst distribution. These observations hold in either the input space or the representation space.
%LEFT: For a usual non-robust model, no more guarantees can be made. A new target distribution $P_{T}$ may or may nor work well beyond $\epsilon'$.
%Suppose we have a $\epsilon$-distributionally robust classifier. ($\min_{\theta} \max_{P_T\;s.t\;D(P_S,P_T)\leq \epsilon} E_{P_T}[l(f(\theta),y)]=a$.)
%RIGHT: The loss is guaranteed to be small ($\leq a$) for any new domain $P_{T_1}$ that is close to the source ($D(P_S,P_{T_1})\leq \epsilon$). 
%Beyond this limit, a target domain $P_{T_2}$ may have a small loss by coincidence while another close domain $P_{T_3}$ may have a large loss again by coincidence.
%}
%In Fig.~\ref{fig:high_variability_of_dg_a}, we demonstrate that such an evaluation is insufficient to assess the  performance of DG methods when deployed in the real world. 

%\BK{To solve this fundamental issue with DG evaluation}
Thus,
%To solve this fundamental issue with DG evaluation, 
we propose a universal, target-independent, and computationally-efficient certification framework that quantifies the risk of deploying a DG model in the real world in terms of its worst-case performance. 
%\PYB{We should motivate the notion of "gap" between worst-case loss and perceived loss on a test dataset, and mention low gap means lower risk (less likely to fail) when a model is deployed in the real world.} added below
%\JH{Mention that we are proposing an ADDITIONALdata-independent method of evaluation, not a replacement of empirical evaluation.}
%\BK{colloquially, certification quantifies the risk of deploying DG models in the real world in terms of worst-case performance.}
%At a colloquial level, our certification method quantifies the risk of deploying a DG model in the real world in terms of its worst-case performance.
%by providing the worst-case loss incurred by a model at any distance in the representation space.
%Our target-independent certification procedure makes use of the distance from the source distribution and certifies the worst case loss of the model at this distance. % using distributionally robust optimization.
%Our target-independent certification method, Cert-DG, certifies the worst-case loss of the model at a distance from the source distribution.
%to certify the worst-case loss incurred by a DG method only a particular distance from the source domains in the representation space. 
Our certification framework, \emph{Cert-DG}, considers the space of probability distributions with Wasserstein distance as a metric and computes the loss of the worst-case distribution in a ball around the source distributions (see Fig.~\ref{fig:method_explanation}) in the representation space. 
Building on existing works on the strong duality of distributionally robust optimization  (DRO)~\cite{gao2016distributionally,sinha2017certifying,bental2013robust} we show the ease of computing the dual optimal value with common losses including cross-entropy, hinge, and 0-1 loss.
%A small gap between the certified loss and empirical loss of a model on a benchmark dataset implies that the model's performance on this benchmark dataset is representative of its performance on other unseen distributions lying at the same distance.\JH{This sounds out of the blue so I removed it.} 
One issue with measuring distances in the representation space is that each representation space has an arbitrary scale, making it difficult to compare the
certificates across models trained with different methods, data, or initializations.
%the certified losses at different distances across DG methods.
To remedy this, we propose a distance normalization approach, for a fair model comparison that measures all distances relative to a reference distribution in the representation space.
We propose to use the distribution consisting of the closest misclassified points of the source domains in the representation space as the reference distribution (see Sec~\ref{sec:certification}), because of its existence in models trained with any method, its intuitive meaning, and ease of computation. 
%\JH{Emphasize why its a NATURAL and computationally sensible choice as a reference. Details car be referred to Sec 3.} 
%This allows us to compare the certificates from our framework across models trained with different methods/initializations. 
%scaling allows us to add meaning to representation space distances which can be scaled arbitrarily across different DG methods.

%\JH{Need a better motivation for DRO. Mention the remaining gap between worst-case and empirical loss, since the models are not trained for the worst-case loss but for empirical loss of training sources. Refer to later figures. }
For models trained with DG methods, we observe a large gap between the certified loss and the empirical loss on benchmark and synthetic domains (see Fig.~\ref{fig:certification_vanilla_wm_rotatedmnist}). 
Therefore, to improve the certified loss of the models trained with DG methods, we also propose a DRO-based iterative training algorithm, \emph{DR-DG}, that augments the training data with samples incurring high loss under the current model.
DR-DG can be used to improve the certified loss of models trained with any DG method and effectively reduces the worst-case loss over a large distance from the sources 
%in the representation space 
with computational complexity similar to adversarial training \cite{madry2017towards,volpi2018generalizing,sinha2017certifying}. 
Our results demonstrate a significant improvement in the certified loss of the models trained with DR-DG with a minor decrease in the accuracy of the models on the source domains. 
The reduction in the gap between the empirical and the certified loss of the models suggests that DR-DG trained models are more generalizable to unseen domains as well their performance on benchmark datasets is representative of their performance in the wild.  
%Thereby making DG models more generalizable to unseen domains in comparison to models trained without DRO training.
%Moreover, our DRO training makes the performance of DG methods on benchmark datasets more representative of their generalization to unseen domains due to a reduction in the gap between the loss and worst-case loss.
Thus, our certification framework, which explicitly provides performance guarantees based on distance, and our algorithm DR-DG, which trains to minimize the worst-case loss, are efficient and effective ways for certifying and improving the generalization of models to unseen domains. 
%Thus, the generality of our certification framework suggests that the worst-case loss at different distances in the representation space should be used as a metric to report success of future DG methods. 
%Moreover, the effectiveness and simplicity of our DRO-based training makes it an appealing choice to improve the generalization of any existing or future DG method. 
%\JH{Mention computational complexity, c.f. adv training, etc.}
%\JH{Mention in summary how our proposed framework can IMPACT the research or usage in practice.}

Our main contributions are as follows.%{\bf Contributions}
%\JH{Where is the emphasis on the universality, computational efficiency, exact certifiability, data-independent etc????? Gotta be very specific.}
%\JH{Also try to write each in the strongest possible tone but objectively.}
\begin{itemize}[leftmargin=0.75cm]
    \item We highlight a fundamental limitation with the evaluation of DG methods using a few benchmark datasets, that could lead to a biased evaluation and an incorrect understanding of their generalization to other unseen domains.
    %of the same distance. to guarantee generalization of models to unseen domains. 
    %\PYB{which could lead to biased evaluation and incorrect understanding of generalization on other distributions of the same distance. }
    %the insufficiency of evaluating DG methods on a few benchmark datasets to guarantee generalization to unseen domains by demonstrating high variability in their accuracy on benchmark and synthetic distributions.
    %on distributions lying at a particular distance in the representation space.  
    \item %\PYB{To achieve provably reliable DG,} 
    To achieve provably reliable DG, we propose a target-independent certification framework that can certify the worst-case performance of any DG method solely based on the distance in the representation space. We show its computational efficiency with common losses including cross-entropy, hinge, and 0-1 loss.
    %a certificate based on worst-case loss the model in a Wasserstein ball around the source distributions in the representation space.
    %to consistently evaluate the generalization of DG methods to unseen domains. Our certificates provide a guarantee on the worst-case performance of a DG method on distributions within a Wasserstein ball around the source distributions in the representation space.
    \item We present a simple intuitive distance normalization technique that allows comparison of any performance metric across different models 
    %learned using different architectures, training methods, or data 
    in relation to distances in the representation space. 
    %to effectively compare distances in the representation space learned using different architectures, training methods, or data. The scaling uses the distribution consisting of closest adversarial examples of the points in the source domains. 
    %The scaling uses the efficiently computable which computes representation space distance normalized?standardized distance that allows direct comparison of different models arising from .
    %Mention average distance to closest point of decision boundary.
    \item We propose a computationally efficient algorithm that improves the generalization performance of models on unseen domains by reducing the worst-case loss at different distances.%, demonstrated by ample examples. 
    %trained with DG methods  guarantees of models trained with different DG methods based on DRO that can be used in conjunction with any existing DG method to provably improves their generalization to unseen domains. 

    %\item We empirically show that many DA and DG approaches are vulnerable to training-time or test-time attacks.
    %Specifically, we present new attack methods based on bilevel optimization to show the failure of . 
    %\item We present a unifying view of analytical works on accuracy/robustness bounds and recent DA/DG works both of which based on Wasserstein/optimal transport distance of distributions.
    %\item We present an algorithm for certifying the accuracy of DA and DG with a DRO approach
    %\item We present results ...
\end{itemize}
\section{Related Work}

%%%%%%%%%%%%%%%%%%%%%%%%%%%%%%%%%%%%%%%%%%%%%%%%%%%%%%%%%%%%%%%%%%%%%%%%%%%%%%%%%%%%%%%%%%%%%%%%%%%%%%%%%%%%%%%%%%%%%%%%%%%%%%%%%%%%%%%%%%%%%%%%%%%%%%%%%%%%%%%%%%%%
{\bf Domain generalization and domain adaptation:} The key problem addressed by works in these areas is to improve the performance of the models when training and test distributions are different. 
Analytical works in these areas \cite{ben2007analysis,ben2010theory,mansour2009domain,shen2018wasserstein,mehra2021understanding,zhao2019learning,johansson2019support, blanchet2019quantifying} have shown that model's generalization performance remains high under distribution shifts when the training (source) and test (target) distributions are close under certain divergence measures. Distributional divergence measures studied in previous works include the Wasserstein distance \cite{sinha2017certifying,kumar2022certifying,kuhn2019wasserstein,gao2016distributionally}, maximum mean discrepancy~\cite{staib2019distributionally}, $f$-divergence~\cite{bental2013robust,weber2022certifying}, $\mathcal{H}$-divergence~\cite{ben2007analysis,albuquerque2019generalizing}.
%Since the target distribution can lie far away from the source distribution(s) in the input space several works impose additional assumptions on the target distribution. 
Since generalization to arbitrary domains is not possible, previous works make additional assumptions on the unseen domains such as \cite{blanchard2011generalizing} assumes that the source and target distributions are derived from the same hyper-distribution, \cite{albuquerque2019generalizing,krueger2021out} assumes that the target distributions belong to the convex hull of the source distribution(s), \cite{kumar2022certifying} considers shifts generated by different parameterized transformations.
Another line of work considers learning a representation space by minimizing different divergence measures between the source distributions \cite{albuquerque2019generalizing,zhang2021quantifying,ganin2016domain,zhao2018adversarial,qiao2020learning,gulrajani2020search}. 
%the distance between distributions can be minimized and if the unseen target distributions aligns close to the sources in this space, performance of the model on the unseen distribution will remain high.
%Due to lack of any target domain information in the domain generalization setting, recent works have aims to improve the performance of the model to unseen domains, recent works have explored different assumptions on the to an arbitrary distributions is not possible, recent works \cite{blanchard2011generalizing,,, ,sehwag2021robust,} have shown th
%This has led to \JH{Is it accurate?} the development of many algorithms \cite{albuquerque2019generalizing,zhang2021quantifying,ganin2016domain,zhao2018adversarial,qiao2020learning,gulrajani2020search} that learn a representation space in which the divergence between the source(s) and target distributions can be minimized. 
Yet another line of research in DG learns a representation space that disentangles \cite{arjovsky2019invariant,zhang2021towards,dittadi2020transfer,montero2020role} domain-specific features from the domain invariant features such that predictors learned on top of these invariant features also become domain invariant. 
Another popular approach for DG uses data augmentation approaches \cite{hendrycks2019augmix,wang2021augmax,kireev2021effectiveness,calian2021defending,sun2021certified} that make the models invariant to common/natural variations of the data from the source.

{\bf Certified robustness:} Point-wise certification guarantees on the performance of the classifier has been extensively studied in the area of certified adversarial robustness \cite{szegedy2013intriguing,madry2017towards,lecuyer2019certified,li2019certified,wong2018provable,raghunathan2018semidefinite,salman2019provably,zhai2020macer,sinha2017certifying,mehra2021robust}, where the certification outputs a radius around a test point within which classifier predictions remain constant. 
Our work is different from these works since we are concerned with distributional robustness guarantees for DG, i.e., a certification that allows us to quantify the generalization performance on an unseen distribution rather than certifying the instance-wise performance (see Appendix~\ref{app:discussion_pointwise} for a comparison between point-wise and distributional robustness). 
Recently, certified robustness has gained attention in the context of certifying the performance of the classifier (in a distributional sense) on bounded distribution shifts \cite{kumar2022certifying,weber2022certifying,sehwag2021robust}. 

{\bf Distributionally robust optimization:} DRO has been extensively used in machine learning to quantify the performance of a model on distributions belonging to different uncertainty sets \cite{bental2013robust, mohajerin2018data, duchi2016statistics,zhao2018data,gao2016distributionally,bertsimas2018data}. Previous works have considered the uncertainty sets to be the Wasserstein balls \cite{sinha2017certifying,gao2016distributionally, shafieezadeh2019regularization,cranko2021generalised,kuhn2019wasserstein,blanchet2019quantifying, volpi2018generalizing,qiao2020learning}, $f$-divergence balls  \cite{bental2013robust,duchi2016statistics,hashimoto2018fairness,hu2018does,weber2022certifying} around the source distribution as well as shifts over different groups  \cite{sagawa2019distributionally}. 
Our certification framework considers the uncertainty set to be the Wasserstein ball around the source distributions in the representation space. 
\section{A framework for provable domain generalization}
%%%%%%%%%%%%%%%%%%%%%%%%%%%%%%%%%%%%%%%%%%%%%%%%%%%%%%%%%%%%%%%%%%%%%%%%%%%%%%%%%%%%%%%%%%%%%%%%%%%%%%%%%%%%%%%%%%%%%%%%%%%%%%%%%%%%%%%%%%%%%%%%%%%%%%%%%%%%%%%%%%%%
\label{sec:certification}

{\bf Notation:} Let $\mathcal{X}$ denote the data domain, $\mathcal{Y}$ denote the labels, and $P_{S/T}(x,y)$ be the joint distributions of features and labels for the source/target domain. Let $\mathcal{D}_{S/T}$ denotes examples drawn from $P_{S/T}(x,y)$.
When there are multiple sources, $P_S^i(x,y)$ denotes the $i^{th}$ source.
Let $g:\mathcal{X}\rightarrow\mathcal{Z}$ be the representation map of an input $x$ to its features in the representation space $\mathcal{Z}$ and let $h:\mathcal{Z}\rightarrow\mathcal{Y}$ be the classifier on top of $\mathcal{Z}$.
A distribution on $\mathcal{Z}\times \mathcal{Y}$ can be a new distribution or a push-forward distribution ($g_{\sharp}\mathcal{P}$) of an input-space distribution, which can be distinguished from the context.% We use the same notation as the distinction will be clear from the context. %do nto , or a new distribution.  a  in the representation space and use $P_S$, $P_S^i$, $P_T^i$, etc., to denote push-forward distributions, with an exception of $P_{S_{adv}}$ which is a distribution in the representation space.

\vspace{-0.15cm}
\subsection{Domain Generalization via minimizing Wasserstein Distance}

Consider the space of joint probability distributions $\Pi(P, Q)$ with marginal distributions $P$ and $Q$, and suppose $c$ is the cost of transporting mass from $(x_1,y_1)$ to $(x_2,y_2)$. The Optimal Transport (OT) distance \cite{villani2009optimal,peyre2019computational,alvarez2020geometric} is the minimum expected cost \(\inf_{\pi \in \Pi(P,Q)} \mathbb{E}_{\pi}[c((x_1,y_1),(x_2,y_2))]\), denoted by $OT_c(P,Q)$. 
%\(OT_c =\min_{\gamma \in \Pi(\mu_1,\mu_2)}<\gamma, C>_F\), for finite empirical distributions (discrete case) where $C\geq0$ is a matrix with pairwise costs and $<\cdot,\cdot>_F$ denotes the Frobenius dot product.
When the cost $c$ is a $\ell_2$-norm, OT distance is also known as the type-2 Wasserstein distance: $W_2(P, Q) = (\inf_{\pi \in \Pi(P, Q)}\mathbb{E}_{(x_1,y_1),(x_2,y_2)\in\pi}[c((x_1,y_1),(x_2,y_2))^2])^\frac{1}{2}$.
Following \cite{sinha2017certifying,volpi2018generalizing}, we measure $OT_c$ between distributions in the representation space $\mathcal{Z}$ with the cost $c((z_1,y_1),(z_2,y_2)) = \|z_1- z_2\|^2_2 + \infty \cdot I[y_1\neq y_2]$, which allows couplings only between points of the same class. 
With this choice of $c$, we have $OT_c=W_2^2$. 
Previous works have shown that the Wasserstein distance can be computed efficiently using linear programming \cite{peyre2019computational,flamary2021pot}, entropic regularization, \cite{cuturi2013sinkhorn} or the dual formulation \cite{genevay2016stochastic,shen2018wasserstein,arjovsky2017wasserstein} for small-scale problems and specific norms. For large-scale problems, regularized version of OT is often used \cite{li2018deep,genevay2018learning}.

%\begin{figure}
\begin{wrapfigure}[27]{r}{0.47\textwidth}
\vspace{-0.45cm}
\small
\centering
\includegraphics[width=0.55\linewidth]{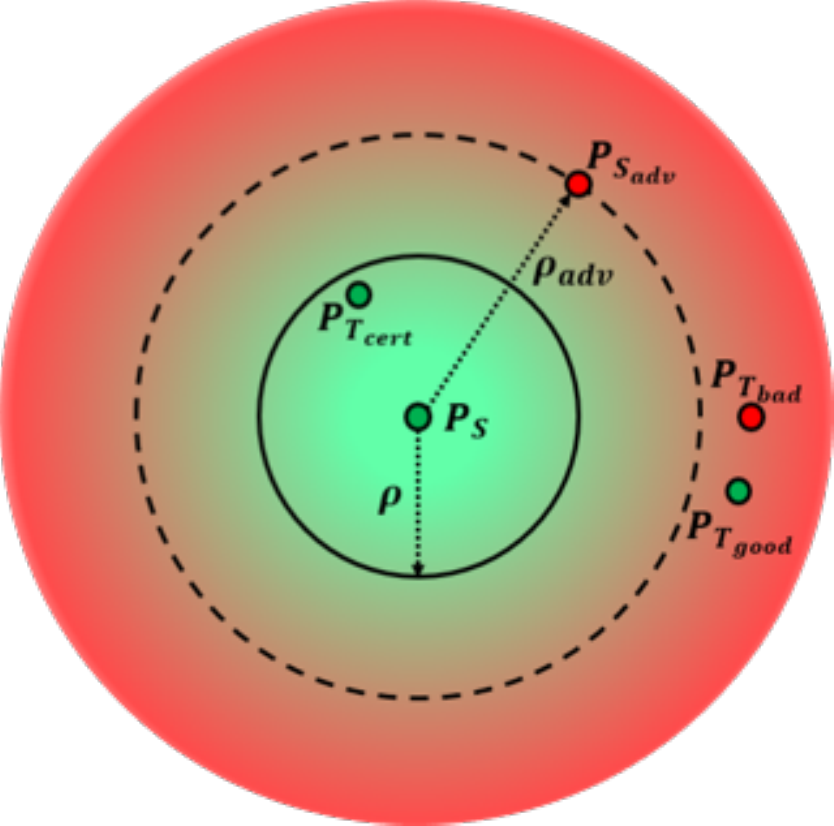}
\caption{The space of probability distributions with Wasserstein distance as a metric. 
$P_S$ and $P_T$ are the sources and the unseen target distributions. 
The color indicates the performance of the given model on a distribution (Green: good, Red: bad).
Naively, the empirical performance (solid color on each point) of a target $P_T$ is only loosely correlated with its distance from $P_S$.
On the other hand, the worst-case performance (gradually changing color from the center to outside) can be efficiently certified for any distance $\rho$ and is independent of targets. \\
The adversarial distribution $P_{S_{adv}}$ is a unique reference point whose distance from the source serves as a unit to normalize distances outside of which the performance becomes very low.%\JH{If out of space, this can go away}
}
\label{fig:method_explanation}
\end{wrapfigure}
%\end{figure}
Many DG methods minimize different divergence measures to align source distributions in the representation space \cite{albuquerque2019generalizing,staib2019distributionally} (see Appendix~\ref{app:dg_algorithms} for a review of DG).
As concrete examples, G2DM~\cite{albuquerque2019generalizing} and CDAN~\cite{long2018conditional} learn a representation space by minimizing the classification loss on multiple source domains and uses one-vs-all type discrimination losses to minimize the divergence between different source domains.
Following these works, we propose a Wasserstein Matching (WM) based algorithm which uses one-vs-all type Wasserstein distance-based loss to minimize the divergence between the source distributions (see Appendix~\ref{app:dg_algorithms} for details). 
The WM method performs competitively with G2DM, CDAN, and other state-of-the-art DG methods on benchmark datasets.
To compute and optimize Wasserstein distances between distributions we follow \cite{damodaran2018deepjdot} which computes the distance cost matrix per batch and then optimizes the coupling for that batch. 
One challenge with measuring distances in the representation space is that the representation map learned by different DG methods or even the same method for different model initialization has arbitrary scales and therefore the distances are not directly comparable across models or methods. 
%which makes the comparison of the different performance metrics (accuracy, loss, certified loss) of the model at a particular distance in the representation space inconsistent.
To address this, we propose a distance normalization technique that uses the distance between the source $P_S$ and a unique reference distribution $P_{S_{adv}}$ as the unit length in the representation space.
This distribution $P_{S_{adv}}$ consists of points $(z', y)$ generated similarly to the CW-attack \cite{sehwag2021robust,carlini2017adversarial}: for each $z$ from the source, $z'$ is the closest misclassified point ($h(z') \neq y$). 
%This procedure is also used in \cite{sehwag2021robust} for a different purpose.
Using this, we report all distances in this paper as the normalized distance $\frac{W_2(P_S,\cdot)}{\rho_{adv}:=W_2(P_S,P_{S_{adv}})}$. 
The unit distance $\rho_{adv}$ (see Fig.~\ref{fig:method_explanation}) is dependent on the learned representation map and the classification boundary. As the adversarial distribution $P_{S_{adv}}$ is the closest distribution whose accuracy is zero (since all points $z'$ are misclassified by construction), the normalized distance of 1 provides a sense of distance from the source where the generalization performance is expected to be low in general (and in particular the accuracy is zero for the adversarial distribution $P_{S_{adv}}$). 
%see Fig.~\ref{fig:method_explanation})

\begin{algorithm}[t] 
\caption{Cert-DG: Certifying Domain Generalization} 
\label{alg:certification}
{\bf Input}: Radius $\rho$, Representation map $g_\theta$, Classifier $h_\zeta$, Source data $\mathcal{D}_{S}$, \\ $T_1$, $T_2=1$, $\alpha$, $\beta$, $\gamma_{init}, \gamma_{min}, \gamma_{max}$. \\ %= \{(x_i, y_i)\}_{i=1}^{N_S}$ (for UDA) OR $\mathcal{D}_{S} = \{\mathcal{D}_{S}^{j}\}_{j=1}^{N}$, $\mathcal{D}_{S}^j = \{(x^j_i, y^j_i)\}_{i=1}^{N_S^j}$ (for DG).\\
{\bf Output}: $l_{worst}$ (worst-case loss within a ball of radius $\rho$)
\begin{algorithmic}
\State{{\bf Init:} $\gamma \leftarrow \gamma_{init}$, $\mathcal{D}_{} \leftarrow (g_\theta(\mathcal{X}_{S}), \mathcal{Y}_S)$} 
%\Comment{Initialize the worst-case distribution with $\mathcal{D}_{S}$}
\For{$m=1,\;\cdots\;,T_1$}
    \For{a batch ${(x, y)} \sim \mathcal{D}_\mathrm{S}$ and corresponding batch $z \sim \mathcal{D}_{}$ of size $n$}
        %\State{}
    	\For{$t=1,\cdots,T_2$} 
    	    \For{$i=1,\cdots,n$} 
        	    \State{$\phi_\gamma \leftarrow \ell(h_\zeta(z^i), y^i) - \gamma \|z^i-g_\theta(x^i)\|_2^2$} 
        	    \Comment{Compute $\phi_\gamma$}
                \State{$z^i \leftarrow z^i + \alpha \nabla_{z} \phi_\gamma $}
                \Comment{Update $z$ and store back in $\mathcal{D}$}
            \EndFor
            \State{$\gamma \leftarrow \gamma -  \beta\{\rho^2 - \frac{1}{n}\sum_{i=1}^{n} \|z^i-g_\theta(x^i)\|_2^2 \}$} \Comment{Update $\gamma$ using the gradient}
            \State{$\gamma \leftarrow \mathrm{Clip}(\gamma, \gamma_{min}, \gamma_{max})$}\Comment{Clip the value of $\gamma$}
        \EndFor
    \EndFor
\EndFor
%\State{$l_{worst} \leftarrow \{\gamma \rho^2 + \mathbb{E}_{(x,y)\sim\mathcal{D}_{S}, z \sim \mathcal{T}_{dro}}[\phi_\gamma((x,y,z))]\}$} 
%\Comment{Return the worst-case loss in the $\rho$ ball} 
\State{$\ell_{worst} \leftarrow \{\gamma \rho^2 + \frac{1}{N}\sum_{(x,y)\sim\mathcal{D}_{S}, z \sim \mathcal{D}}[\phi_\gamma((x,y,z))]\}$} 
\Comment{Return the worst-case loss in the $\rho$ ball} 
\end{algorithmic}
\end{algorithm}

\subsection{Certifying Domain Generalization}
%\BK{A lot of figure refs are missing (Fig. ??)}
As demonstrated in Figs.~\ref{fig:high_variability_of_dg_a} and~\ref{fig:high_variability_of_dg_b} (in Appendix), the empirical loss of various distributions has high variability and is not a deterministic function of the distance to the source. 
Thus, to provide a data-independent measure of DG performance, we propose to use the worst-case loss at a particular distance, as used in distributionally robust optimization \cite{bental2013robust, mohajerin2018data, duchi2016statistics,zhao2018data,gao2016distributionally,bertsimas2018data}.
%\JH{1. Define worst-case loss is, and explain why this notion is so useful. Explain discrete vs continuous distinctions}
Considering the space of joint distributions on $\mathcal{X}\times \mathcal{Y}$ with 2-Wasserstein distance as a metric \cite{peyre2019computational} (see Fig.~\ref{fig:method_explanation}). 
The worst-case loss of any distribution $P$ whose distance is at most $\rho$ from source distribution $Q$ is then
\begin{equation}
\label{Eq:dro}
\sup_{P: W_2(P, Q) \leq \rho}\;\mathbb{E}_{(x,y) \sim P}[\ell(h(g(x)), y)].
\end{equation}
%\JH{2. Explain surrogate loss, strong duality, citing mostly Gao}
This (primal) optimization problem is an infinite-dimensional problem over a convex set $P$, which makes it difficult to solve it directly, in general.
However, the optimal value can be computed through a finite-dimensional optimization problem through the following equality
\begin{equation}
\label{Eq:strong_duality}
\sup_{P:W_2(P, Q)\leq\rho}\mathbb{E}_P[\ell(h(g(x)), y)] = \inf_{\gamma \geq 0}\{\gamma \rho^2 + \mathbb{E}_{(x_0,y_0) \sim Q}[\phi_\gamma(x_0,y_0)]\},\;\;\mathrm{where}
\end{equation}
\begin{equation}\label{eq:surrogate_loss}
\phi_{\gamma}(x_0,y_0):= \sup_{x\in\mathcal{X}}\;\{\ell(h(g(x)),y_0) - \gamma c((g(x_0),y_0),(g(x),y_0))\}
%\phi_{\gamma}(x_0,y_0):= \sup_{x\in\mathcal{X}}\;\{l(h(g(x)),y_0) - \gamma \|g(x_0)-g(x)\|^2\} 
\end{equation}
%\AM{Should we use norm in place of c here as well?} 
is the Moreau envelope or the surrogate function \cite{sinha2017certifying,volpi2018generalizing}. 
The equality in Eq.~\ref{Eq:strong_duality} is called the strong duality and is proven under different assumptions on $c$, $\ell$
and whether the nominal distribution $Q$ is continuous, empirical ($\frac{1}{N} \sum_i \delta(z-z_i)$, where $N$ denotes number of points in the empirical distribution) or both~ \cite{gao2016distributionally,sinha2017certifying,mohajerin2018data,zhao2018data,blanchet2019quantifying}.
%\JH{3. Explain Cert-DG exactly and Explain how it's different from Sinha or others}
To certify the data-independent worst-case loss of any DG method, we solve Eq.~\ref{Eq:strong_duality} using the algorithm presented in Alg.~\ref{alg:certification} called Cert-DG.
The proposed algorithm is different from existing works in the following ways. 
%are key differences of the proposed algorithm different from existing works.
 \cite{sinha2017certifying,volpi2018generalizing} proposed to solve the Lagrangian relaxed penalized version while we solve the  non-relaxed problem by minimizing over the dual variable $\gamma$ (Eq.~\ref{Eq:strong_duality}) as well as computing the surrogate loss (Eq.~\ref{eq:surrogate_loss}).
More importantly, since DG methods align the distributions in the representation space (one layer before the logit/softmax layer \cite{albuquerque2019generalizing,long2018conditional}), we solve Eq.~\ref{Eq:strong_duality} directly in this space using the following surrogate loss 
\begin{equation}\label{eq:rep-surrogate}
\phi_{\gamma}(z_0,y_0):= \sup_{z\in\mathcal{Z}}\;\{\ell(Wz+b,y_0) - \gamma \|z-z_0\|_2^2\}.   
\end{equation}
Since the maximization is in the representation space (see Appendix~\ref{app:discussion} for a discussion on input space vs representation space), we do not require assumptions such as Lipschitzness of $\nabla_z \ell(z;\theta)$ and $\nabla_\theta \ell(z;\theta)$, as made in \cite{sinha2017certifying}, which are prohibitive for the certification of deep neural nets (and can be NP-hard with ReLU networks, as shown in \cite{sinha2017certifying}).
For Eq.~\ref{eq:rep-surrogate} to have a finite optimal value that is also computationally easy to find, we do not necessarily require differentiability or even continuity of $\ell$ \cite{gao2016distributionally}. This allows us to use various types of losses such as cross-entropy, hinge, and 0/1-loss. 
To solve Eq.~\ref{eq:rep-surrogate}, we propose to use SGD for cross-entropy and modified hinge loss. For 0/1-loss the surrogate loss is given in the closed-form (see Appendix~\ref{app:other_losses} for the details for different losses). 
For these losses, the problem in Eq.~\ref{Eq:strong_duality} is a saddle point problem that is convex in the scalar $\gamma$ and is strongly concave in $z$ (for hinge loss and for cross-entropy loss with $\gamma > \gamma_0$, for some $\gamma_0$). 
Thus, the problem can be solved efficiently by alternating SGD as proposed in Alg.~\ref{alg:certification}. For example, certifying a model in the experiments at a distance $\rho$ using a sample of 1000 points takes only about a minute on a GPU enabled machine.
%due to the fact that the optimization is taking place in the representation space (see Appendix~\ref{app:discussion} for a discussion on input space vs representation space).
Thus, we propose a \emph{computationally efficient method of certifying the worst-case loss of any deep NN-based DG method at an arbitrary distance $\rho$ in the representation space, that is compatible with different types of loss functions.}

\begin{algorithm}[t] 
\caption{DR-DG: Distributionally-Robust Domain Generalization} 
\label{alg:dro_training}
{\bf Input}: $F$, $T_1$, $T_2$, $\alpha$, $\beta$, $\eta$, $\gamma_{init}, \gamma_{min}, \gamma_{max}$, $\mathcal{D}_{S} = \{\mathcal{D}_{S}^{j}\}_{j=1}^{N_S}$, $\mathcal{D}_{S}^{j}=\{(x^j_i, y^j_i)\}_{i=1}^{N_S^j}$, $\ell_{dg}$\\
{\bf Output}: Representation model parameters $\theta$, Classifier parameters $\zeta$.
\begin{algorithmic}
\State{{\bf Init:} Initialize parameters of the representation $g_\theta$ and classification $h_\zeta$ models.}
\State{{\bf Init:} $\gamma \leftarrow \gamma_{init}$, $\mathcal{D}_{} \leftarrow g_\theta(\mathcal{X}_{S})$}
%\Comment{Initialize the worst-case distribution with $\mathcal{D}_{S}$}
\For{$m=1,\;\cdots\;,T_1$}
    \State{$\mathcal{D}_{adv} \leftarrow \mathrm{GenAdvDist}(\mathcal{D}_{S}, g_\theta, h_\zeta)$} \Comment{Generate adversarial distribution via CW-attack \cite{carlini2017adversarial}}
    \State{$\rho \leftarrow $F$ \cdot W_2(\mathcal{D}_{adv}, (g_\theta(\mathcal{X}_{S}), \mathcal{Y}_S))$} \Comment{Initialize the ball size $\rho$ }
    \For{a batch ${(x, y)} \sim \mathcal{D}_\mathrm{S}$ and corresponding batch $z \sim \mathcal{D}_{}$ of size $n$}
    	\For{$t=1,\cdots,T_2$} 
    	    \For{$i=1,\cdots,n$} 
        	    \State{$\phi_\gamma \leftarrow \ell(h_\zeta(z^i), y^i) - \gamma \|z^i - g_\theta(x^i)\|_2^2$} 
        	    \Comment{Compute $\phi_\gamma$}
                \State{$z^i \leftarrow z^i + \alpha \nabla_{z} \phi_\gamma $}
                \Comment{Update and store $z$}
            \EndFor
            \State{$\gamma \leftarrow \gamma -  \beta\{\rho^2 - \frac{1}{n}\sum_{i=1}^{n} \|z^i-g_\theta(x^i)\|_2^2 \}$} \Comment{Update $\gamma$ using the gradient}
            \State{$\gamma \leftarrow \mathrm{Clip}(\gamma, \gamma_{min}, \gamma_{max})$}\Comment{Clip the value of $\gamma$}
        \EndFor
    
        %\State{\# Compute the robust surrogate $\phi_\gamma$}
        %\State{$\{(x_{rob}^i, y^i)\}_{i=1}^n \leftarrow \{(x^i, y^i)\}_{i=1}^n$}
    	%\For{$t=1,\cdots,T_2$}
        %    \State{$x^i_{rob} \leftarrow x^i_{rob} + \alpha \nabla_{x} \{\ell(h_\zeta(g_\theta(x^i_{rob})), y^i) - \gamma c_{g_\theta}((x^i_{rob}, y^i), (x^i, y^i))\} $}
        %\EndFor
        \State{}
        \State{$\ell_{dro} \leftarrow \frac{1}{n}\sum_{i=1}^{n}\ell(h_\zeta(z^i), y^i)$}
        \Comment{Loss on the worst-case distribution in $\rho$-ball}
        %\State{$\ell_{Orthogonal} \leftarrow \|\zeta^T\zeta - \mathbb{I}\|_F$}
        %\Comment{Orthogonality constraint on Classification model params}
        %\State{}
        %\State{$\ell_{clean} \leftarrow \frac{1}{n}\sum_{i=1}^{n}\ell(h_\zeta(g_\theta(x^i)), y^i)$}
        %\Comment{Loss on the source distribution}
        %\State{\JH{Let's remove UDA stuff}}
        %\If{Task = UDA}
        %    \State{$\ell_{crossdomain} \leftarrow W_2^2(D_S, D_T)$}
        %\ElsIf{Task = DG} 
        %\State{$\ell_{dg} \leftarrow \sum_{i=1}^{N}\sum_{j=i+1}^{N}W_2^2(\mathcal{D}^i_S, \mathcal{D}^j_S)$} \Comment{Loss used by the given DG algorithm.}
        %\EndIf
        \State{$\ell_{total} \leftarrow \ell_{dro} + \ell_{dg}$}
        \Comment{$\ell_{dg}$ is the loss of a given DG method.}
        %\ell_{Orthogonal
        %\State{\JH{What's $\zeta$ and $\theta$? Also, please high level descriptions: Added}}
        
        \State{$\theta \leftarrow \theta - \eta\nabla_\theta\ell_{total}$} \Comment{Update Representation model params}
        \State{$\zeta \leftarrow \zeta - \eta\nabla_\zeta\ell_{total}$} \Comment{Update Classification model params}
    \EndFor
\EndFor
\end{algorithmic}
\end{algorithm}

%%%%%%%%%%%%%%%%%%%%%%%%%%%%%%%%%%%%%%%%%%%%%%%%%%%%%%%%%%%%%%%%%%%%%%%%%%%%%%%%%%%%%%%%%%%%%%%
\subsection{Improving Domain Generalization with DRO}
%%%%%%%%%%%%%%%%%%%%%%%%%%%%%%%%%%%%%%%%%%%%%%%%%%%%%%%%%%%%%%%%%%%%%%%%%%%%%%%%%%%%%%%%%%%%%%%

The results of using our certification (see Fig.~\ref{fig:certification_vanilla_wm_rotatedmnist} and Figs.~\ref{fig:loss_before_after_wm_g2dm}, ~\ref{fig:loss_before_after_cdan_vrex} in the Appendix) on models trained with Vanilla DG methods show a large gap between the worst-case performance and performance on benchmark tasks.
This gap exists since the vanilla models are not optimized to learn a representation space where the loss of the worst-case distribution is small at any given distance from the sources. 
%sn analogy .. randomize 
Thus, to improve the worst-case performance of a given DG algorithm, we propose to perform distributionally-robust model training (\(\min_{\zeta,\theta,\gamma\geq0} \{\gamma \rho^2 + \mathbb{E}_Q[\phi_\gamma]\}\)) in addition to minimizing losses of specific DG algorithm. 
Our algorithm takes the losses of specific DG algorithms as input and optimizes those along with the DRO loss. 
Our proposed algorithm, DR-DG, is presented in Alg.~\ref{alg:dro_training}.
Similar to Cert-DG, DR-DG can be used in conjunction with any existing DG method to improve their worst-case loss at various distances. 
%\AM{Any loss can be used with DR-DG. Explain ldro does not change the behavior of the DG algorithm.}
DR-DG improves the worst-case loss by alternating between two steps.
In the first step, we generate the data which approximates the loss the worst-case distribution by solving the dual problem in Eq.~\ref{Eq:strong_duality} over a mini-batch. 
The size of the Wasserstein ball $\rho$ is computed relative to the distance to the adversarial distribution $\rho_{adv}$ (see Fig.~\ref{fig:method_explanation}) i.e. $\rho = F \cdot \rho_{adv}$, where $F$ is a factor specified as an input parameter.
In the second step, we minimize the objectives of the DG algorithm and additionally minimize the loss of the worst-case samples generated in the first step. 
In the experiments, we demonstrate that this simple training procedure leads to a significant reduction in the worst-case loss of the model.
This makes models trained with our method certifiably more generalizable to unseen domains than their vanilla counterparts.
To run the optimization end-to-end with SGD, the choice of loss needs to be restricted to (a.e.) differentiable losses. However, separate from the training procedure, the trained model can be certified with any loss (see Fig.~\ref{fig:certification_before_after_dro}). 
In practice, the additional objective of generating and training on samples that approximate the loss of the worst-case distribution only adds a small overhead to vanilla DG training. The overall run-time increase is proportional to the number of maximization steps executed per batch ($T_2$ in Alg.~\ref{alg:dro_training}). For $T_2$=20 on PACS with WM and G2DM the run-time increases by mere $\sim$3 seconds per epoch in our experiments.
Lastly, while we are minimizing the worst-case loss of the empirical distribution in practice, it also minimizes the worst-case loss of the unknown true distribution as shown in Theorem 3 \cite{sinha2017certifying}: 
\(
\sup_{P: W_2(P,{Q})\leq \rho} \mathbb{E}_Q[\ell] \leq \gamma \rho^2 + \mathbb{E}_{\hat{Q}}[\phi_{\gamma}] + O(n^{-1/2})
\) with high probability, where $\hat{Q}$ is the $n$-sample empirical training (source) distribution of $Q$.
Thus, DR-DG reduces the worst-case loss and improves its generalization on unseen domains in a principled way.

\if 0
By the ehe way, \cite{sinha2017certifying} in Sec 3.1. discusses the data-dependent upper bound
\[
\sup_{P: W_2(P,Q)\leq \rho} E_P[l] \leq \gamma \rho + E_{\hat{P}}[\phi_{\gamma}]+O(n^{-1/2})
\]
with high probability, where $\hat{P}$ is an IID sample of $N$ point from $P$. The additional error term is due to this finite-sample error. We could adopt this viewpoint, but we will consider $\epsilon$-ball around the empirical distribution to simplify discussions. 

Maybe related? \cite{peyre2019computational, dudley69}
$E[|W_p(P,Q)-W_p(\hat{P},\hat{Q})|] = O(N^{-1/d})$ for $d>2$ and $p\geq 1$ for $\mathbb{R}^d$ and measure supported on bounded domain.
\fi

\if0
\subsubsection*{Algorithm for DRO}
We are solving a Lagragian Dual problem $\inf_{\gamma \geq 0}\{\gamma \rho^2 + \mathbb{E}_Q[\phi_\gamma(\theta;Z)]\}$.
For the dual variable $\gamma$, it's a convex function since it is a pointwise supremum of an affine function (i.e., Lagrangian). 
\fi

%\vspace{-0.35cm}
%%%%%%%%%%%%%%%%%%%%%%%%%%%%%%%%%%%%%%%%%%%%%%%%%%%%%%%%%%%%%%%%%%%%%%%%%%%%%%%%%%%%%%%%%%%%%%%%%%%%%%%%%%%%%%%%%%%%%%%%%%%%%%%%%%%%%%%%%%%%%%%%%%%%%%%%%%%%%%%%%%%%
\section{Experiments}
%%%%%%%%%%%%%%%%%%%%%%%%%%%%%%%%%%%%%%%%%%%%%%%%%%%%%%%%%%%%%%%%%%%%%%%%%%%%%%%%%%%%%%%%%%%%%%%%%%%%%%%%%%%%%%%%%%%%%%%%%%%%%%%%%%%%%%%%%%%%%%%%%%%%%%%%%%%%%%%%%%%%
In this section, we present empirical evaluations of Cert-DG and DR-DG. % on using  along with an evaluation of models trained with DR-DG DRO using Cert-DG as well as standard benchmark datasets. \JH{This sentence is problematic}
As mentioned in Sec.~\ref{sec:certification}, we use various DG algorithms that learn a representation space by minimizing the divergence between the source distributions during training. 
%The choice of the domains used for training is arbitrary, and to demonstrate that the performance of DG methods on benchmark datasets may not be representative of methods generalization peris not enough to guarantee generalization we reserve other domains for various evaluations of DG methods. to unseen domains  we reuse we chose to train using just two domains and test on the rest of the domains evaluation.
In particular, we use Wasserstein matching (WM), which explicitly reduces the Wasserstein distance between the source distributions in the representation space (see Appendix~\ref{app:dg_algorithms}), G2DM \cite{albuquerque2019generalizing} and CDAN \cite{long2018conditional} which use discriminators to align the source distributions. We also present results on another popular DG algorithm VREX \cite{krueger2021out} in the Appendix.
We use benchmark datasets Rotated MNIST (R-MNIST) \cite{ghifary2015domain}, PACS \cite{li2017deeper} and VLCS \cite{fang2013unbiased}. Dataset details are present in Appendix~\ref{app:experimental_details}.
%Specifically, R-MNIST consists of 6 image domains rotated by multiples of $15^{\circ}$ (angles ranging from $0^{\circ}$ to $75^{\circ}$) while PACS consists of four image domains of different styles (Photos, Art, Cartoon, and Sketch).
Similar to \cite{gulrajani2020search}, we use a convolutional neural network for R-MNIST and fine-tune a  ResNet50 model pre-trained on Imagenet for PACS and VLCS. 
%Results on other DG algorithms and datasets are presented in Appendix~\ref{app:additional_experiments}.
For all our experiments,  we arbitrarily chose two domains for training ($0^{\circ}$ and $15^{\circ}$ for R-MNIST, Cartoon and Sketch for PACS, Caltech101 and SUN09 for VLCS), while reserving others for various evaluations. 
%For experiments on other datasets and algorithms see Appendix~\ref{app:additional_experiments}.
We used the Python OT library \cite{flamary2021pot} to measure Wasserstein distances. As described in Sec.~\ref{sec:certification}, we use the $W_2$ Wasserstein distance in our work. During the training of WM, we use batch-wise computation of Wasserstein distance similar to \cite{damodaran2018deepjdot}. 
To be able to compare the distances across different representation spaces, as shown in Sec.~\ref{sec:certification}, we normalize the distance by the distance between the source and its adversarial distribution in the representation space. The distance computation and Alg.~\ref{alg:dro_training} rely on the computation of the adversarial distribution. We use the Cleverhans \cite{papernot2018cleverhans} implementation of the CW attack \cite{carlini2017towards} in the representation space for computing this distribution. 
%\JH{No, this choice is not arbitrary and it's our contribution!!! What other distribution can you use naturally? Give me one} The choice of measuring representation space distance using the adversarial distribution is arbitrary and is based on computational efficiency of finding this distribution with existing methods. However, any other distribution can be used in place of $S_{adv}$. 
Additional results on Cert-DG and DR-DG are in Appendix~\ref{app:additional_experiments} along with experimental details in Appendix~\ref{app:experimental_details}. Our codes are available at \url{https://github.com/akshaymehra24/CertifiableDG}.

\subsection{Performance of DG algorithms on unseen distributions}
\label{sec:corruptions}
To understand how the accuracy of the models trained with DG methods changes on the unseen domains in relation to the distance of the unseen domains in the representation space, we evaluate the models on several benchmark and synthetic domains.
For models trained with R-MNIST, we use, $30^\circ, 45^\circ, 60^\circ, 75^\circ$ as unseen domains, and for models trained with PACS, we use Art and Photo as unseen domains.
We generate additional unseen target domains by adding natural corruptions \cite{hendrycks2019benchmarking,mu2019mnist} to test data of the source domains. For R-MNIST we use corruptions from MNIST-C \cite{mu2019mnist} and for PACS we used corruptions from Imagenet-C \cite{hendrycks2018benchmarking} (see Appendix~\ref{app:fig_1_corruptions}).
%changes when faced with unseen domains at rest time in relation to the distance  generalization of current DG methods, we evaluate their performance on different unseen distributions which are obtained by modifying the test sets of the data from the source domains.
%We use natural corruptions \cite{hendrycks2019benchmarking,mu2019mnist} to create various distributions as substitutes for possible distributions that may be encountered when the models are deployed in the real world scenarios. 
We chose to use common corruptions to create the unseen domains since they resemble some of the variations that may be encountered at test time. We control  the severity level to create a number of unseen domains whose distance changes with severity. 
%of the corruptions we can several unseen domains. ability to change the parameters of the corruptions allows us to create distributions of varying distances from the source in the representation space learned by DG methods.  
%The results of the performance of models on these domains are presented in  Fig~\ref{fig:high_variability_of_dg_a}. 
As observed in Fig~\ref{fig:high_variability_of_dg_a} and~\ref{fig:high_variability_of_dg_b}, accuracy of DG methods on distributions lying at similar distances in the representation space from the source has high variability. 
In particular, the smaller-range plots show that the accuracy of these models varies by about 5-7\% on different unseen domains. 
%, we see that their performance can change by about 5-7\% even when these distributions are close to the sources. 
This variability in the accuracy is not only observed for corrupted distributions but also on benchmark distributions, where the distributions of Art and Photo lie at a similar distance to the sources in the representation space but have about 15\% performance variation. 
%is not particular to the distributions we crafted but also happens on benchmark distributions as seen in  
Moreover, the results also highlight the limitation of evaluating DG methods only on benchmark datasets since the performance of the models on these datasets is not representative of the model's performance on unseen domains. In particular, the high performance on Photos domain in Fig~\ref{fig:high_variability_of_dg_a}, does not imply that the model's performance on other domains at that distance will also be high.
Lastly, across algorithms performance of models can be different in different domains. For instance on R-MNIST (Fig.~\ref{fig:high_variability_of_dg_b} in Appendix), WM is good on scaling compared to all other methods but performs worse on shot noise.
These differences in the performance of the models trained with DG methods make it difficult to assess the generalization of these models to unseen domains by only empirical evaluation and motivates the need for a principled certification framework for DG. 
%This high variability in the performance of the DG methods makes it difficult to consistently evaluate their generalization to new domains as well as makes it difficult to compare different DG methods{}. This demonstrates the need for a certification procedure for evaluating the generalization of DG.

\subsection{Certifying the worst-case loss of models trained with DG methods}
\label{sec:cert_vanilla}
\begin{wrapfigure}[14]{r}{0.65\textwidth}
\vspace{-0.35cm}
\centering
\includegraphics[width=0.48\linewidth]{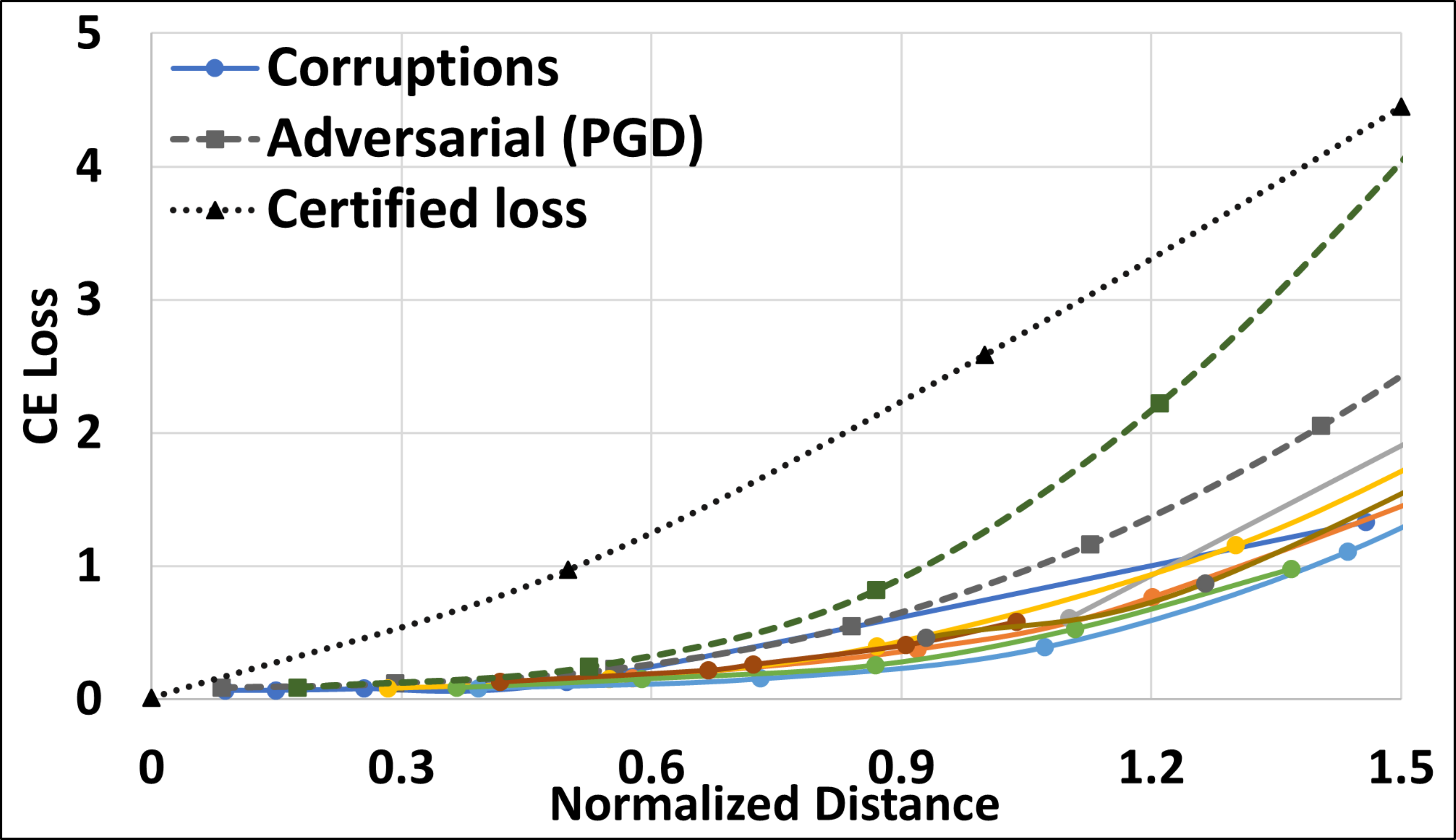}
\includegraphics[width=0.48\linewidth]{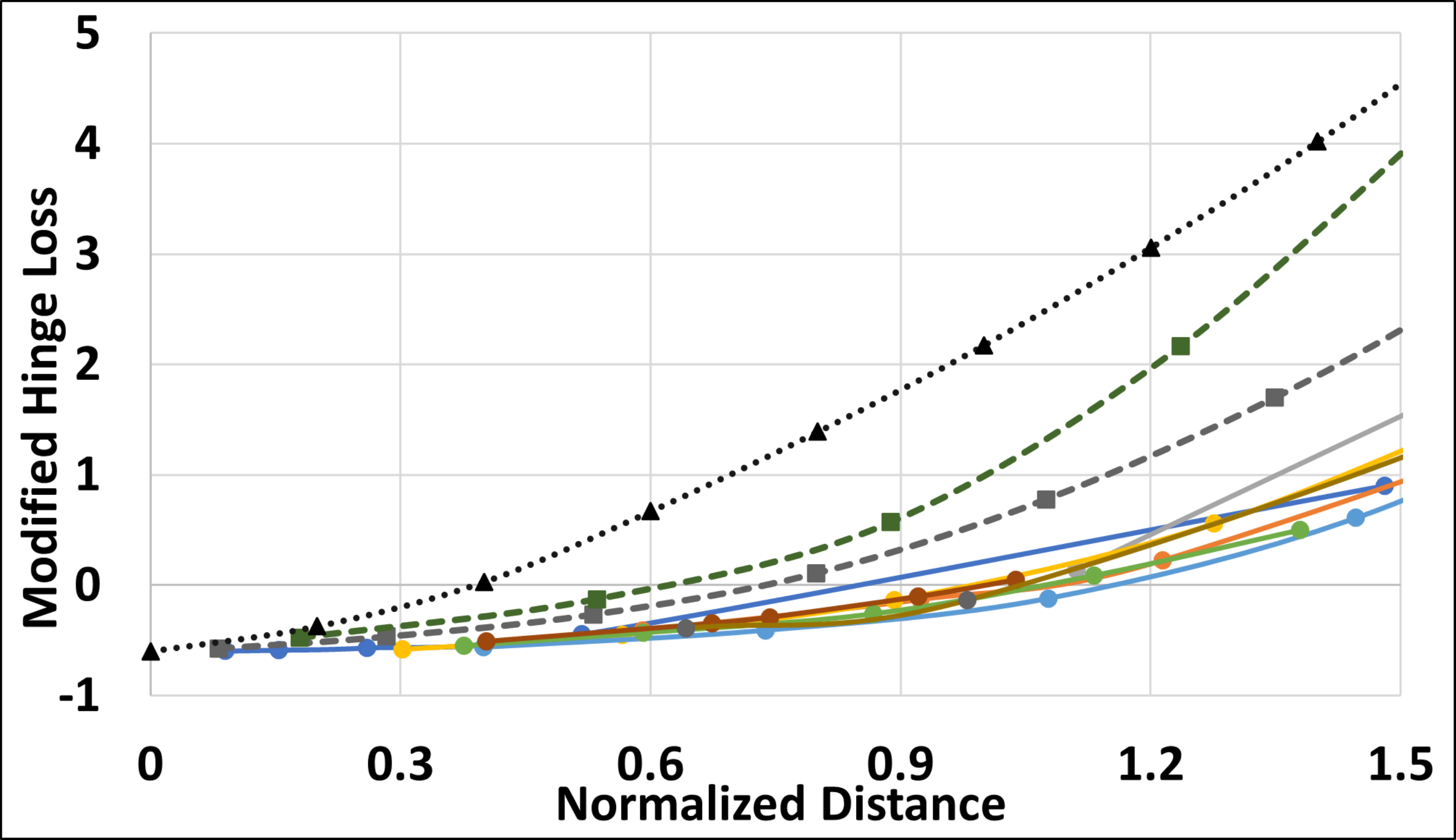}
\caption{Comparison of certified loss (Cross-entropy (left) and modified hinge loss (right)) and empirical loss on corrupted and adversarial domains of a model trained with WM on R-MNIST. The large gap between the certified and empirical losses indicates the insufficiency of evaluation of DG methods on a few unseen domains to guarantee generalization on other domains.  
%\AM{Mention equation of DRO loss}
%were used.
%Compatibility of our certification procedure with different loss functions such as the cross-entropy and the modified hinge loss. The certified loss provides a meaningful measure of the performance of the models on unseen natural and adversarial domains.
%\BK{provide detail on data/method}
}
\label{fig:certification_vanilla_wm_rotatedmnist}
\end{wrapfigure}

%High accuracy of DG methods on a few benchmark distributions is insufficient to guarantee their performance on unseen distributions even if they lie at the same distance as the benchmark distributions.
%\AM{Do we need certification results in the input space?}
%\begin{figure}[tb]
In this section, we use Cert-DG, Alg.~\ref{alg:certification}, to compute the worst-case loss of the DG methods at different distances from the source distribution in the representation space.
As seen in Fig.~\ref{fig:certification_vanilla_wm_rotatedmnist}, the high worst-case loss, even close to the source, indicates the existence of distributions that can potentially lead to a drastically different performance than that expected from the evaluation on benchmark datasets. 
%\end{figure}
In addition to corrupted domains, we also plot adversarial domains, in Fig.~\ref{fig:certification_vanilla_wm_rotatedmnist}, crafted by PGD \cite{madry2017towards} attacks in the input as well as the representation space (see Appendix~\ref{app:adversarial_attacks}). 
The large gap between the certified loss and the empirical loss indicates the existence of distributions beyond corruptions and adversarial examples, which could degrade the performance of the models.
Since the worst-case distribution computed by Cert-DG can contain points distorted by different amounts, its loss can be higher than the loss incurred by a PGD attack which constrains the distortion of every point to be the same \cite{madry2017towards,kumar2022certifying}. 
This highlights that Cert-DG considers a more general worst-case distribution than the adversarial distribution with a fixed perturbation budget.
%generate the worst-case att by distorting every point differently it can 
%We observe that close to the source, the worst-case loss is even higher than the loss incurred by adversarial distributions, which is expected by definition. 
%Another explanation for this is, that the Wasserstein attack \JH{ You don't wanna call it that?} distorts each point in different amounts to maximize the loss, 
%whereas other point-wise attack like the PGD attack uses the same distortion constraint $\epsilon$ for all points \cite{madry2017towards,kumar2022certifying}. \JH{Cite the paper}.
As shown in Sec.~\ref{sec:certification}, our certification framework is compatible with various loss functions. Here we demonstrate the certification performance of DG methods with the cross-entropy and modified hinge loss. 
%We demonstrate the certification using two different loss functions namely the cross-entropy and the modified hinge loss.
%e also demonstrate the generality of our certification framework to work with different type of loss functions in Fig.~\ref{fig:certification_vanilla_wm_rotatedmnist}. 
As seen from Fig.~\ref{fig:certification_vanilla_wm_rotatedmnist}, the certification produces similar results when using different loss functions although we found that the modified hinge loss makes the inner maximization better behaved (see Appendix~\ref{app:other_losses}).% as mentioned in Sec.~\ref{sec:certification}. 

\subsection{Certifying the worst-case loss of models trained with DR-DG}
\label{sec:certify_dr_dg}
\begin{wrapfigure}[17]{r}{0.65\textwidth}
%\begin{figure}[tb]
\vspace{-0.4cm}
  \centering
  %\subfigure[WM with R-MNIST]{\includegraphics[width=0.43\columnwidth]{Images/certification_ce_vanilla_wm_rotatedmnist.pdf}}
  %\hfill
  \includegraphics[width=0.48\linewidth]{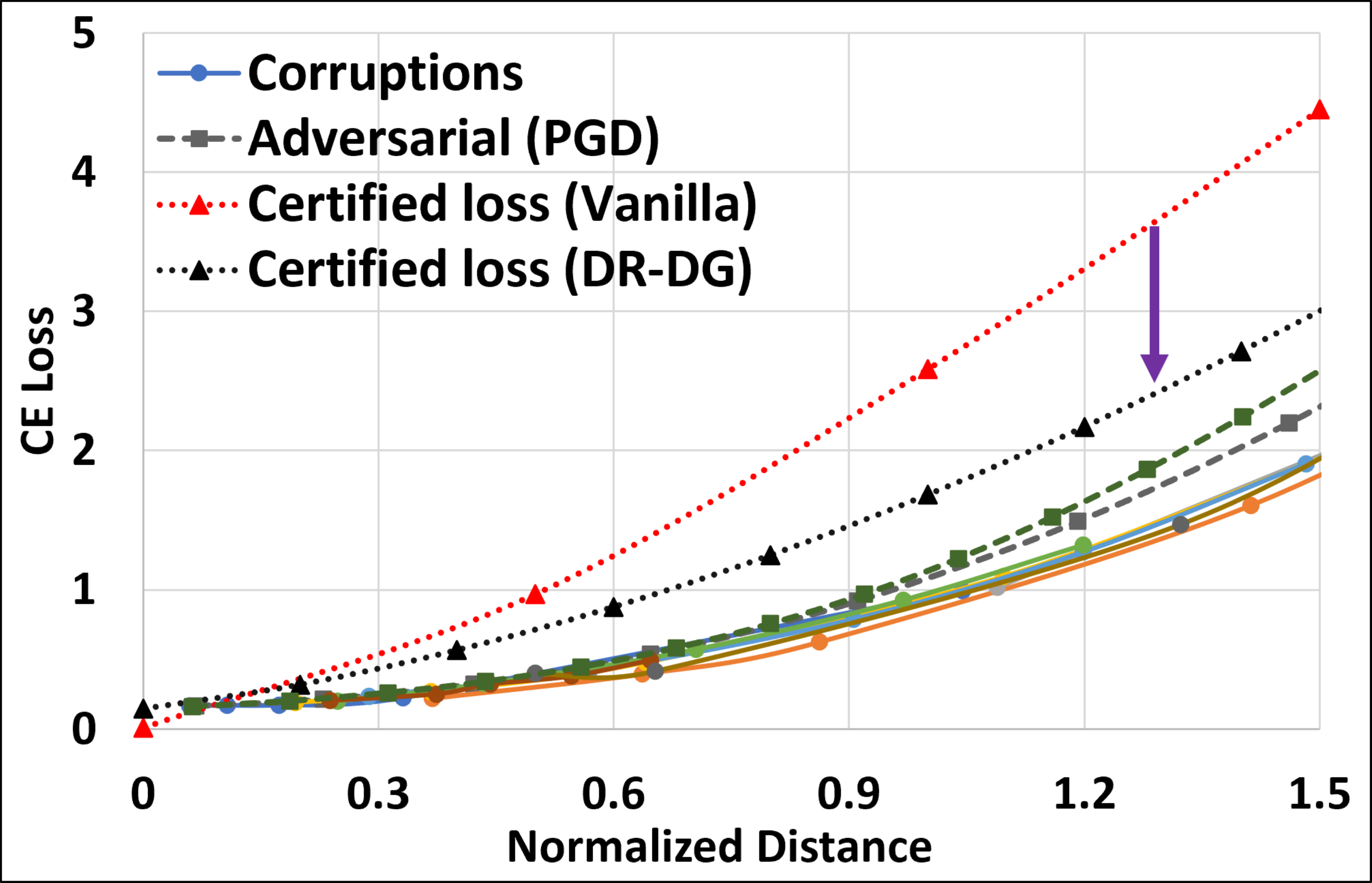}
  %\hfill
  %\subfigure[WM with R-MNIST]{\includegraphics[width=0.44\columnwidth]{Images/certification_hinge_vanilla_wm_rotatedmnist.pdf}}
  %\hfill
  \includegraphics[width=0.48\linewidth]{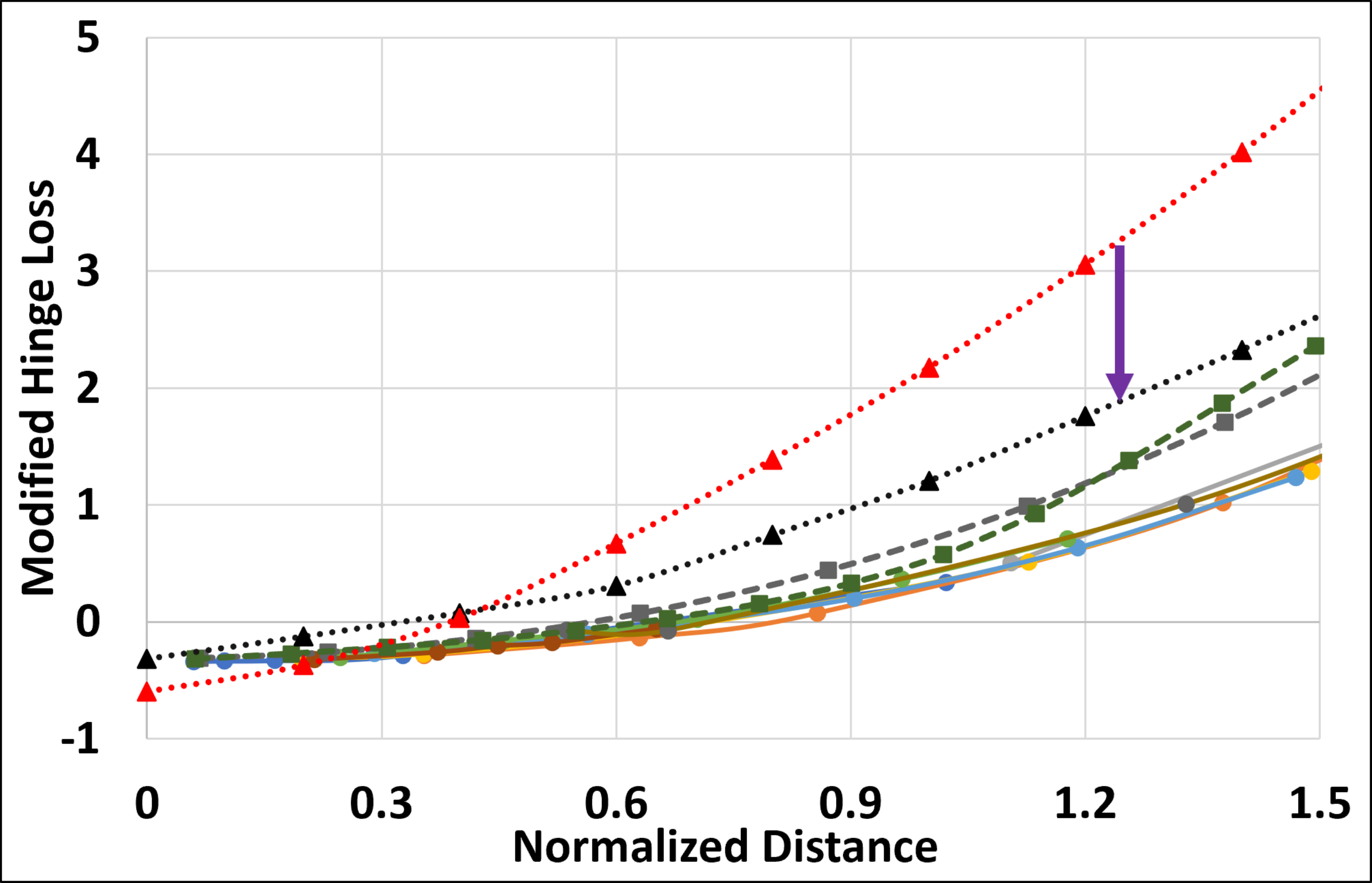}
 %\vspace{-0.3cm}
  \caption{(Best viewed in color.) Improvement in the  worst-case loss (certified loss) with models trained with DR-DG and cross-entropy loss with a $F$ = 0.5 using WM on R-MNIST. Compared to results in Fig.~\ref{fig:certification_vanilla_wm_rotatedmnist}, DR-DG has significantly reduced the variability in the loss of different unseen distributions as well as reduced the worst-case loss. The red dotted line denotes the certified worst-case loss of the model trained with WM without DR-DG (as shown in Fig.~\ref{fig:certification_vanilla_wm_rotatedmnist} and has been shown for reference.
}
  \label{fig:certification_before_after_dro}
%\end{figure}
\end{wrapfigure}
\begin{figure}[tb]
  \centering
  \subfigure[WM with R-MNIST]{\includegraphics[width=0.24\columnwidth]{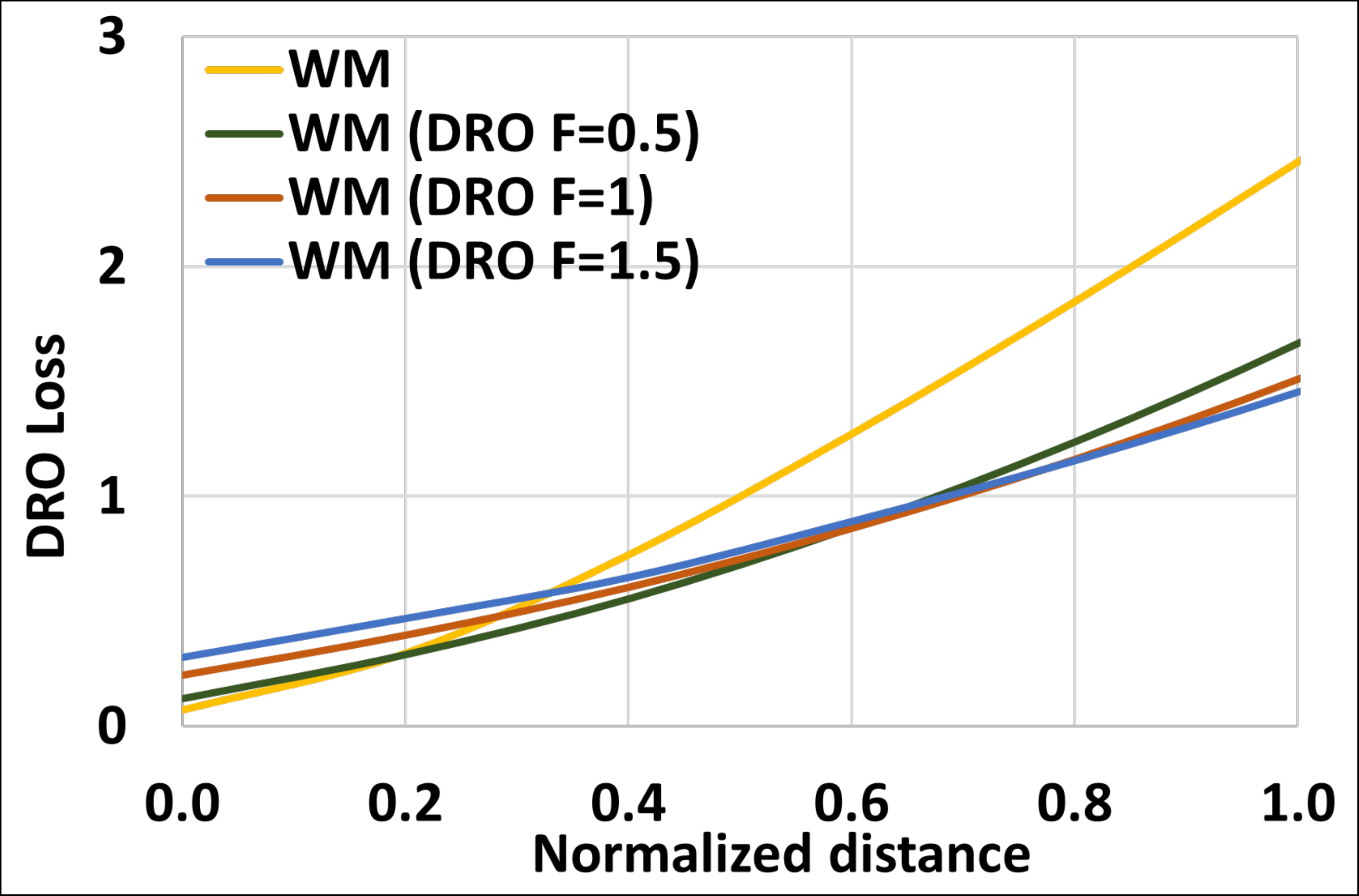}}
  %\hfill
  \subfigure[G2DM with R-MNIST]{\includegraphics[width=0.24\columnwidth]{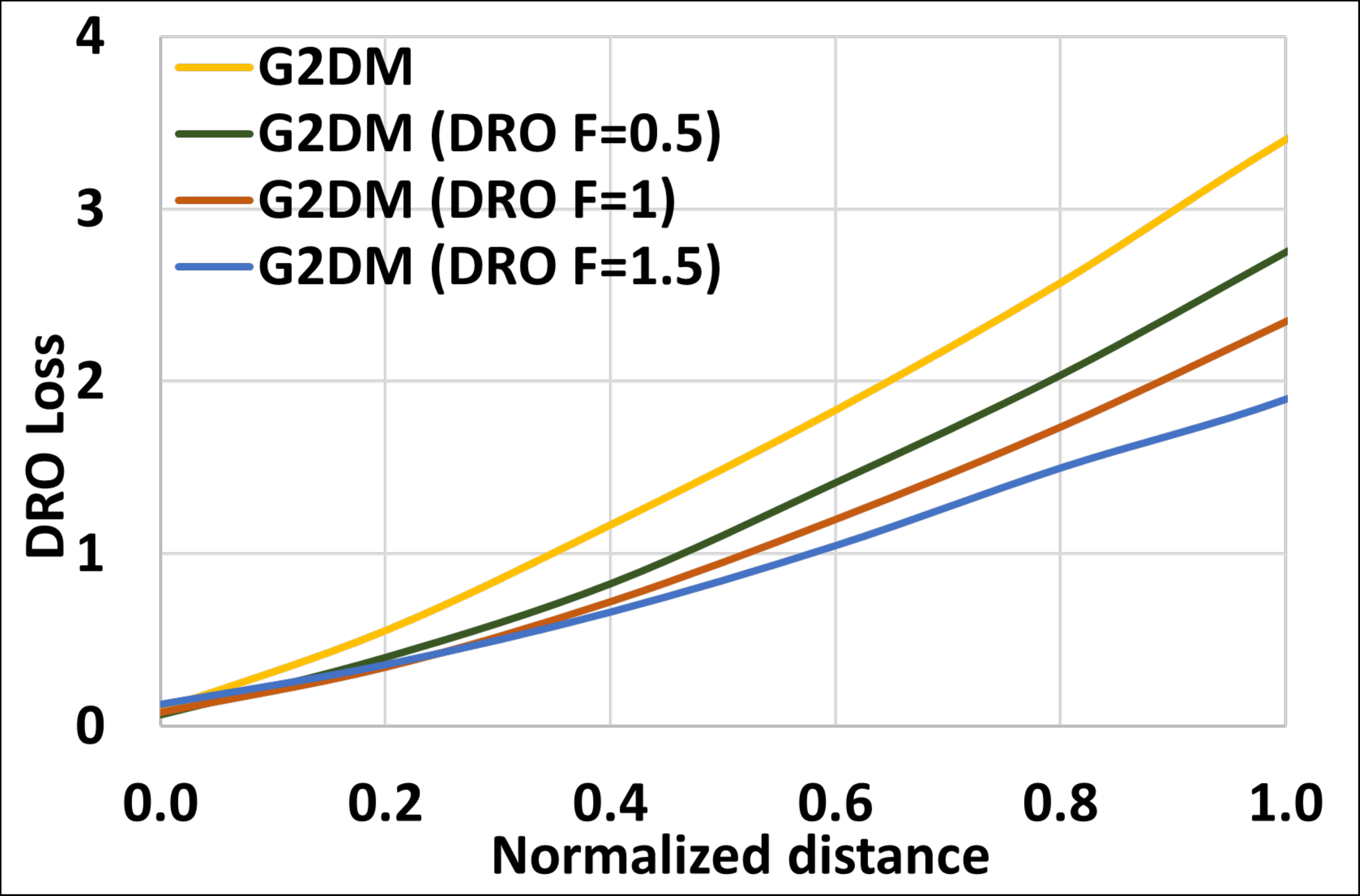}}
  %\hfill
  \subfigure[WM with PACS]{\includegraphics[width=0.24\columnwidth]{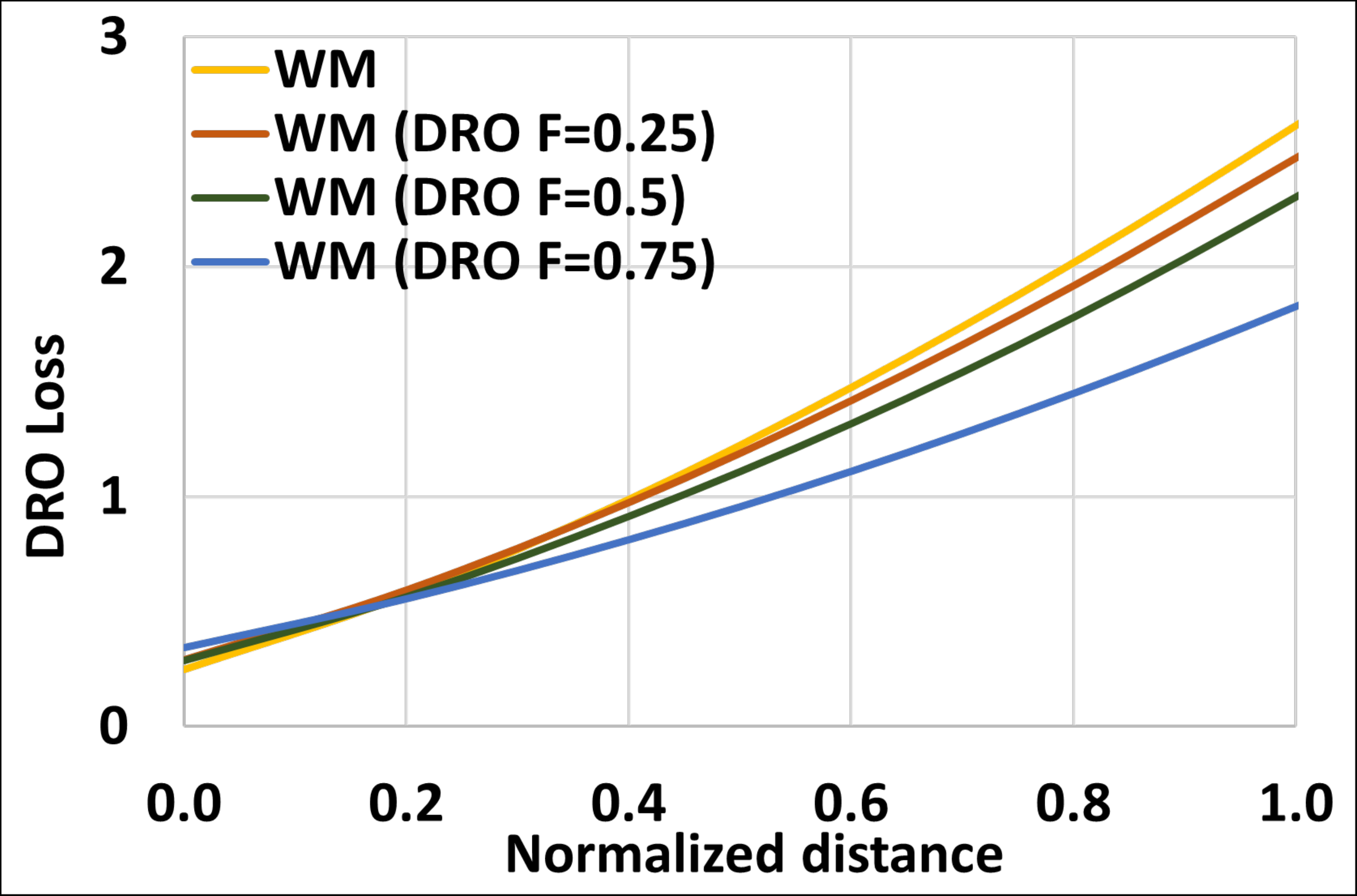}}
  %\hfill
  \subfigure[G2DM with PACS]{\includegraphics[width=0.24\columnwidth]{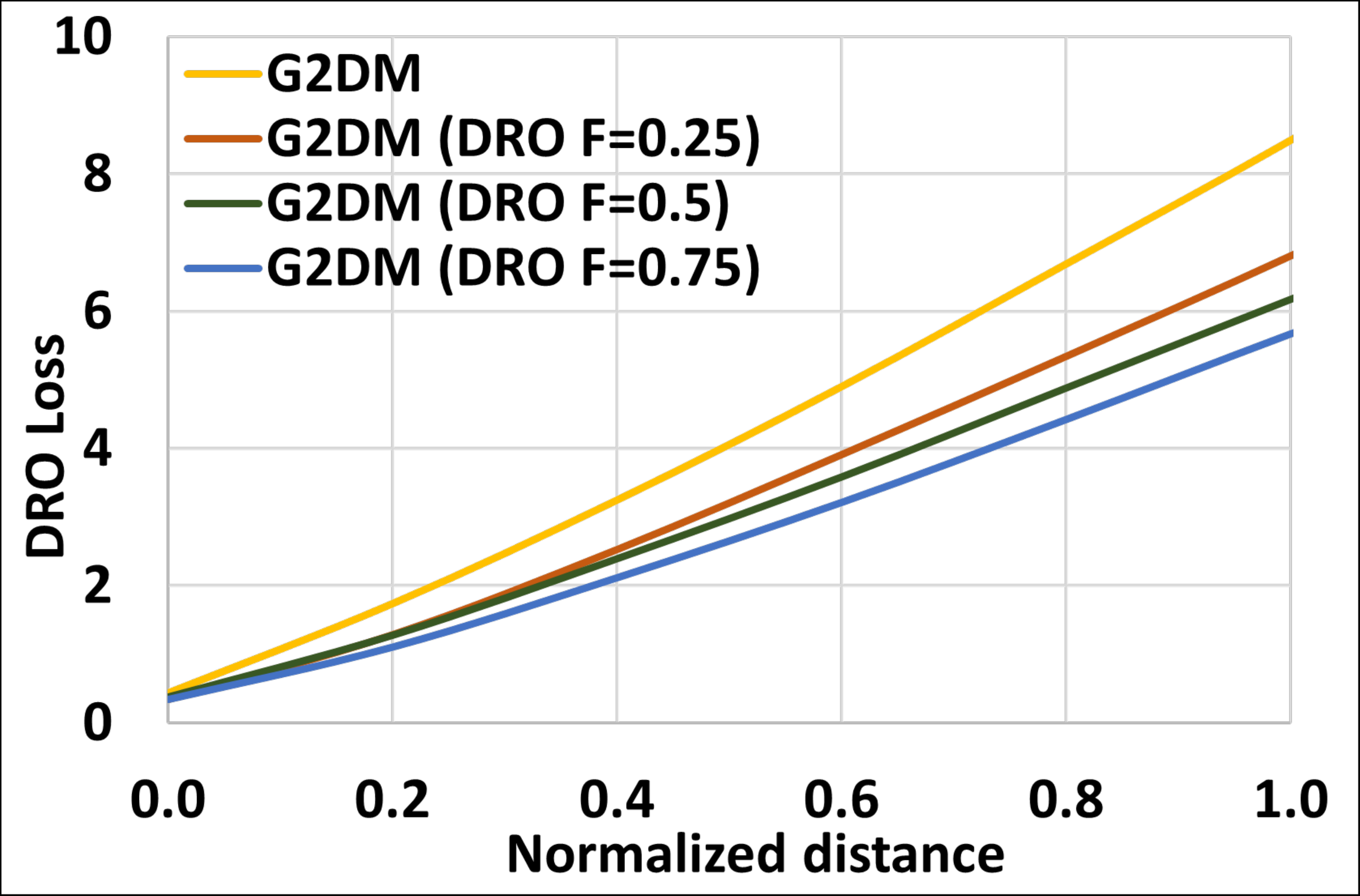}}
 %\vspace{-0.3cm}
  \caption{(Best viewed in color.) DR-DG proposed in Alg.~\ref{alg:dro_training} significantly reduces the worst-case loss of the models trained with DG methods on different datasets. The reduced worst-case loss provably improves the generalization of the models to unseen distributions in comparison to models trained without DR-DG.
   %\AM{Change font}
   }
  \label{fig:dro_trained_models}
\end{figure}

The large gap between the certified worst-case and empirical loss of the models trained with DG methods indicates that their performance can have high variability even on distributions lying at similar distances in the representation space.
DR-DG Alg.~\ref{alg:dro_training}, is proposed to improve the worst-case performance by training on examples drawn from the worst-case distributions lying at different distances in the representation space.
DR-DG requires a $F$  which is used to determine the radius $\rho$ of the Wasserstein ball around the source distributions.
DRO training aims to improve the loss on the worst-case distributions lying in this ball of radius $\rho$. 
The larger the $F$, the larger will be the radius $\rho$ of the ball that will be considered for finding the worst-case distributions.
Similar to point-wise adversarial robustness procedures such as adversarial training \cite{madry2017towards} or randomized smoothing \cite{cohen2019certified}, 
DR-DG with larger values of $\rho$ can improve the worst-case loss but can lead to a reduction in the performance of the model near the source distributions. 
Thus, there exists a trade-off between lowering the worst-case loss at a larger distance versus maintaining the low loss near the source(s).
To demonstrate this trade-off we use three different values of $F$.
The results in Fig.~\ref{fig:dro_trained_models} demonstrate that models trained with DR-DG achieve significantly better worst-case loss in comparison to the models trained with vanilla DG methods with only a minor increase in the loss on the source distribution(s). 
Moreover, compared to the results in Fig.~\ref{fig:certification_vanilla_wm_rotatedmnist}, we observe that in Fig.~\ref{fig:certification_before_after_dro} DR-DG has 1) significantly reduced the variability in the performance of the models across various unseen distributions and 2) lowered the loss of the worst-case distribution at most distances.
This demonstrates that DR-DG has improved the generalization of DG methods to unseen domains.
We observe similar results on other benchmark datasets and algorithms and present those in Fig.~\ref{fig:loss_before_after_wm_g2dm},~\ref{fig:loss_before_after_cdan_vrex} in the Appendix. 

%\vspace{-0.25cm}
\subsection{Performance of models trained with DR-DG on benchmark datasets}
\label{sec:eval_on_benchmark_data}
In this section, we evaluate the performance of models trained with DR-DG, Alg.~\ref{alg:dro_training}, on various target distributions used for comparing the performance of DG methods in previous works. 
As demonstrated in Table~\ref{Table:dro_on_benchmark_datasets}, our algorithm improves the certified loss of the models for all benchmark distributions. 
This reduces the gap between the empirical loss of the model on a particular unseen distribution and the worst-case distribution at that distance (from the sources).
The reduced gap not only leads to a provably reduced variability in the performance of the model on target distributions lying at that distance but also makes evaluation on benchmark datasets more representative of the generalization of the models to unseen distributions. 
However, as discussed in the previous section, gaining robustness to the worst-case distributions may lead to a decrease in the performance of the model on sources as well as on some unseen domains.
This is similar to the accuracy vs robustness trade-off observed in point-wise adversarial robustness literature.
Recent research in that area has shown that it is possible to maintain the accuracy of the model and gain robustness by using different loss functions \cite{zhang2019theoretically} and modifying the training procedure. 
While this is an interesting direction, we leave the use of different techniques for keeping the performance of DG methods on sources and benchmark targets for future work.

\begin{table}
\vspace{-0.42cm}
  \caption{\small{DR-DG trained models on two source domains from R-MNIST ($F_1 = 0.5, F_2 = 1, F_3 = 1.5$) and PACS ($F_1 = 0.25, F_2 = 0.5, F_3 = 0.75$), incur small worst-case (certified) loss. DR-DG also helps reduce the gap between the empirical loss of the models on unseen domains in these datasets and the worst-case loss. This makes the performance of DG models on benchmark datasets more representative of their performance on unseen domains lying at a similar distance.}}
  \label{Table:dro_on_benchmark_datasets}
  \centering
  \small
  \resizebox{\columnwidth}{!}{
    \begin{tabular}{c|cccc|cccc|cccc|cccc}
    \toprule
    & \multicolumn{4}{c|}{$30^\circ$} & \multicolumn{4}{c|}{$45^\circ$} & \multicolumn{4}{c|}{Art} & \multicolumn{4}{c}{Photo}\\
    \cmidrule{2-17}
    Method & Accuracy & Loss & \makecell{Certified \\ Loss} & Gap & Accuracy & Loss & \makecell{Certified \\ Loss} & Gap & Accuracy & Loss & \makecell{Certified \\ Loss} & Gap & Accuracy & Loss & \makecell{Certified \\ Loss} & Gap \\
    \midrule
    WM & 92.11 & 0.26 & 1.87 & 1.61 & 65.59 & 1.21 & 3.76 & 2.55 & 73.24 & 0.82 & 2.63 & 1.82 & 91.55 & 0.35 & 2.14 & 1.79\\
    DR-DG$_{F_1}$ & 89.61 & 0.38 & 0.85 & {\bf0.47} & 61.12 & 1.24 & 2.02 & {\bf0.82} & 71.48 & 0.92 & 2.21 & {\bf1.29} & 85.93 & 0.49 & 1.87 & {\bf1.39}\\
    DR-DG$_{F_2}$ & 88.25 & 0.52 & 0.80 & {\bf0.28} & 54.26 & 1.36 & 1.78 & {\bf0.44} & 69.58 & 1.01 & 2.12 & {\bf1.11} & 87.12 & 0.52 & 1.60 & {\bf1.08}\\
    DR-DG$_{F_3}$ & 86.90 & 0.61 & 0.87 & {\bf0.26} & 54.46 & 1.41 & 1.77 & {\bf0.36} & 74.23 & 0.88 & 1.53 & {\bf0.64} & 89.41 & 0.48 & 1.27 & {\bf0.79}\\
    \midrule
    G2DM & 93.86 & 0.24 & 4.54 & 4.29 & 70.58 & 1.25 & 7.21 & 5.97 & 75.68 & 1.16 & 11.57 & 10.41 & 89.64 & 0.45 & 11.23 & 10.78\\
    DR-DG$_{F_1}$ & 93.24 & 0.23 & 3.27 & {\bf2.99} & 69.23 & 1.14 & 5.34 & {\bf4.20} & 69.14 & 1.19 & 9.61 & {\bf8.42} & 84.43 & 0.58 & 9.51 & {\bf8.93}\\
    DR-DG$_{F_2}$ & 90.43 & 0.31 & 2.24 & {\bf1.93} & 63.72 & 1.24 & 4.06 & {\bf2.82} & 67.23 & 1.32 & 9.59 & {\bf8.23} & 89.40 & 0.41 & 9.45 & {\bf9.05}\\
    DR-DG$_{F_3}$ & 87.94 & 0.39 & 1.48 & {\bf1.08} & 60.08 & 1.25 & 2.84 & {\bf1.58} & 70.65 & 1.05 & 8.26 & {\bf7.22} & 85.74 & 0.50 & 8.29 & {\bf7.78}\\
    \if0
    \midrule
    \midrule
    & \multicolumn{4}{c|}{Art} & \multicolumn{4}{c}{Photo}\\
    \cmidrule{1-9}
    WM & 73.24 & 0.82 & 2.63 & 1.82 & 91.55 & 0.35 & 2.14 & 1.79\\
    ~+DRO ($F=0.25$) & 71.48 & 0.92 & 2.21 & {\bf1.29} & 85.93 & 0.49 & 1.87 & {\bf1.39} \\
    ~+DRO ($F=0.50$) & 69.58 & 1.01 & 2.12 & {\bf1.11} & 87.12 & 0.52 & 1.60 & {\bf1.08} \\
    ~+DRO ($F=0.75$) & 74.23 & 0.88 & 1.53 & {\bf0.64} & 89.41 & 0.48 & 1.27 & {\bf0.79} \\
    \midrule
    G2DM & 75.68 & 1.16 & 11.57 & 10.41 & 89.64 & 0.45 & 11.23 & 10.78\\
    ~+DRO ($F=0.25$) & 69.14 & 1.19 & 9.61 & {\bf8.42} & 84.43 & 0.58 & 9.51 & {\bf8.93} \\
    ~+DRO ($F=0.50$) & 67.23 & 1.32 & 9.59 & {\bf8.23} & 89.40 & 0.41 & 9.45 & {\bf9.05} \\
    ~+DRO ($F=0.75$) & 70.65 & 1.05 & 8.26 & {\bf7.22} & 85.74 & 0.50 & 8.29 & {\bf7.78} \\
    \fi
    \bottomrule
    \end{tabular}
    }
\end{table}

\section{Conclusion}
%%%%%%%%%%%%%%%%%%%%%%%%%%%%%%%%%%%%%%%%%%%%%%%%%%%%%%%%%%%%%%%%%%%%%%%%%%%%%%%%%%%%%%%%%%%%%%%%%%%%%%%%%%%%%%%%%%%%%%%%%%%%%%%%%%%%%%%%%%%%%%%%%%%%%%%%%%%%%%%%%%%%
%\vspace{-0.2cm}
Domain generalization aims to produce models whose performance remains high even on data from domains unseen during training.
However, the current evaluation of these methods using a few benchmark datasets is a poor indicator of their generalization to data from unseen domains.
To make the evaluation of DG reliable, we proposed a novel target-independent certification framework that allows us to evaluate the worst-case performance of any DG method based on the distance in the representation space.
To reduce the large gap between the certified loss and the empirical loss incurred by DG methods on different benchmark distributions we proposed a novel training algorithm based on DRO that can provably improve the worst-case loss of any DG method.
Based on our results, we believe that evaluation of DG methods via a target-independent metric such as the worst-case loss at a given distance could provide additional insights beyond empirical evaluation on benchmark datasets.
Thus, allowing us to objectively judge the progress made towards learning models whose performance does not degrade when faced with data from domains unseen during training.

\section*{Acknowledgement}
This work was supported by the NSF EPSCoR-Louisiana Materials Design Alliance (LAMDA) program \#OIA-1946231 and by the LLNL-LDRD Program under Project No. 20-ERD-014.
This work was performed under the auspices of the U.S. Department of Energy by Lawrence Livermore National Laboratory under Contract DE-AC52-07NA27344.

%\clearpage
\bibliographystyle{plain}
%\typeout{} 
{\small \bibliography{neurips_2022}}

\medskip

\clearpage
\appendix
%%%%%%%%%%%%%%%%%%%%%%%%%%%%%%%%%%%%%%%%%%%%%%%%%%%%%%%%%%%%%%%%%%%%%%%%%%%%%%%%%%%%%%%%%
\begin{center}
{\LARGE \bf Appendix}
\end{center}
%%%%%%%%%%%%%%%%%%%%%%%%%%%%%%%%%%%%%%%%%%%%%%%%%%%%%%%%%%%%%%%%%%%%%%%%%%%%%%
We present additional evaluation results of Cert-DG and DR-DG in Appendix~\ref{app:additional_experiments}, followed by details of our experiments and model architectures in Appendix~\ref{app:experimental_details}. We also provide a brief review of various DG algorithms used in our paper along with existing analyses of the DG problem in Appendix~\ref{app:dg_algorithms}. Lastly, we discuss certification using different loss functions and the difference between input-space and representation-space certification in Appendix~\ref{app:discussion}.

%%%%%%%%%%%%%%%%%%%%%%%%%%%%%%%%%%%%%%%%%%%%%%%%%%%%%%%%%%%%%%%%%%%%%%%%
\section{Additional experimental results}
\label{app:additional_experiments}
%%%%%%%%%%%%%%%%%%%%%%%%%%%%%%%%%%%%%%%%%%%%%%%%%%%%%%%%%%%%%%%%%%%%
\subsection{Variability in the performance of models trained with various DG methods}
As discussed in the introduction and Sec.~\ref{sec:corruptions}, models trained with different DG methods can have high variability in their performance. 
Surprisingly, this variability happens even on distributions at the same distance from the sources in the representation space.
In addition to Fig.~\ref{fig:high_variability_of_dg_a}, Fig.~\ref{fig:high_variability_of_dg_b} demonstrates this variability on models trained with WM, G2DM, CDAN \cite{long2018conditional}, and VREX \cite{krueger2021out} on the R-MNIST dataset. The variability in the performance of models on unseen distributions lying at the same distance from the sources in the representation space, makes it hard to evaluate the generalization performance of models trained with DG methods using only a few benchmark datasets.

\subsection{Additional results of Cert-DG and DR-DG on CDAN and VREX}
Similar to Fig.~\ref{fig:dro_trained_models}, we demonstrate the effect of using different $F$ values for DR-DG training on other popular DG methods, namely, CDAN \cite{long2018conditional} and VREX \cite{krueger2021out} on R-MNIST, PACS and VLCS in Fig.~\ref{fig:dro_trained_models_all}. 
As noted in Sec.~\ref{sec:certify_dr_dg}, we observe that DR-DG training using CDAN and VREX with different $F$ values improves the worst-case loss of models over models trained with their vanilla counterparts.
Larger values of $F$ lead to a higher decrease in the worst-case loss but can increase the loss on the source distributions and other distributions that lie close to the sources in the representation space.
This again suggests a trade-off between improving the worst-case loss of the models at a larger distance from the sources at the cost of potentially increasing the loss of models on distributions lying close to the sources. 

\subsection{Additional results of Cert-DG and DR-DG on VLCS dataset}
\begin{figure}[tb]
  \centering
  \subfigure[WM with R-MNIST]{\includegraphics[width=0.32\columnwidth]{Images/dro_wm_rotatedmnist.pdf}}
  \subfigure[WM with PACS]{\includegraphics[width=0.32\columnwidth]{Images/dro_wm_pacs.pdf}}
  \subfigure[WM with VLCS]{\includegraphics[width=0.32\columnwidth]{Images/dro_wm_vlcs.pdf}}
  
  \subfigure[G2DM with R-MNIST]{\includegraphics[width=0.32\columnwidth]{Images/dro_g2dm_rotatedmnist.pdf}}
  \subfigure[G2DM with PACS]{\includegraphics[width=0.32\columnwidth]{Images/dro_g2dm_pacs.pdf}}
  \subfigure[G2DM with VLCS]{\includegraphics[width=0.32\columnwidth]{Images/dro_g2dm_vlcs.pdf}}
  
  \subfigure[CDAN with R-MNIST]{\includegraphics[width=0.32\columnwidth]{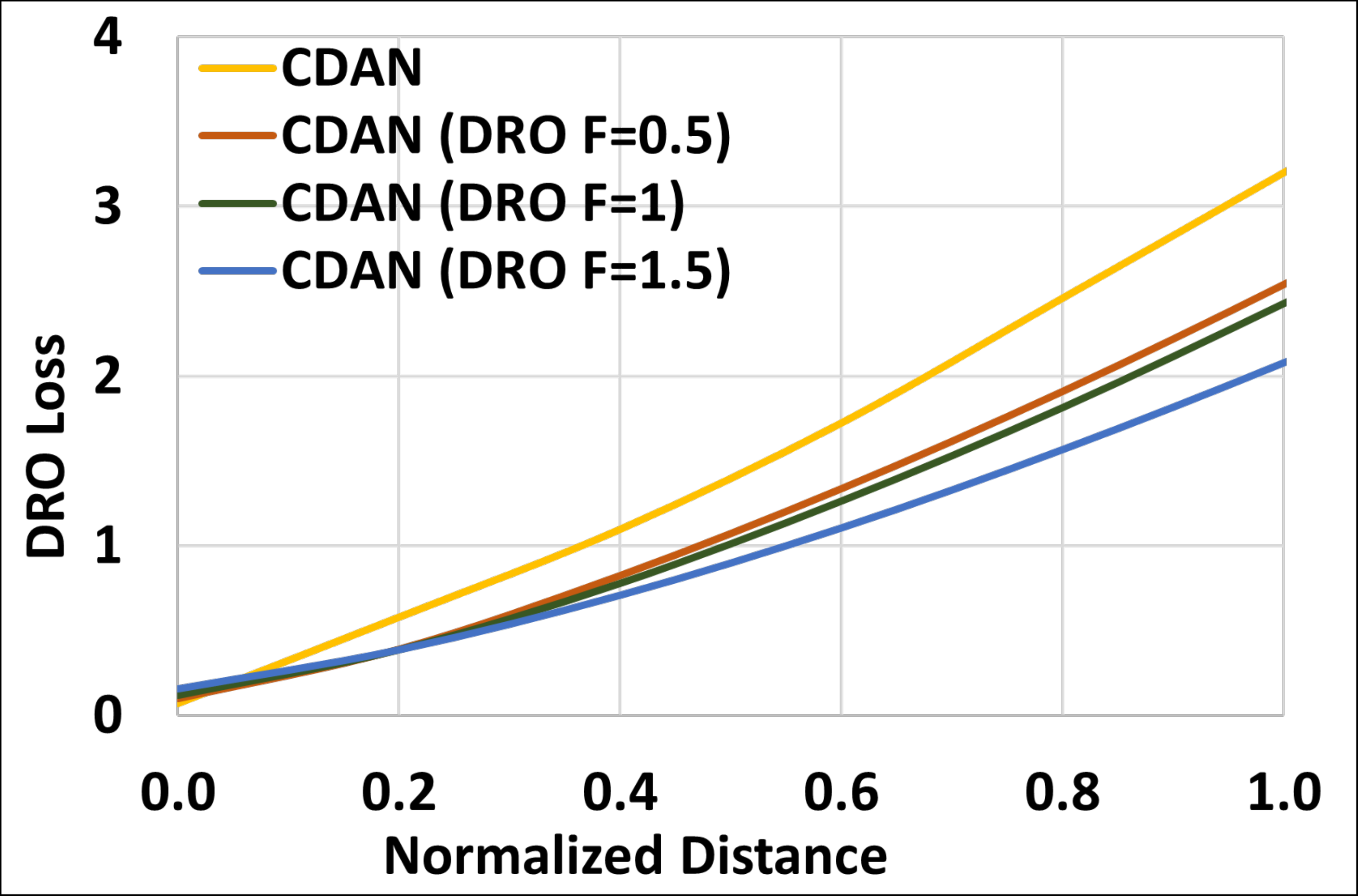}}
  \subfigure[CDAN with PACS]{\includegraphics[width=0.32\columnwidth]{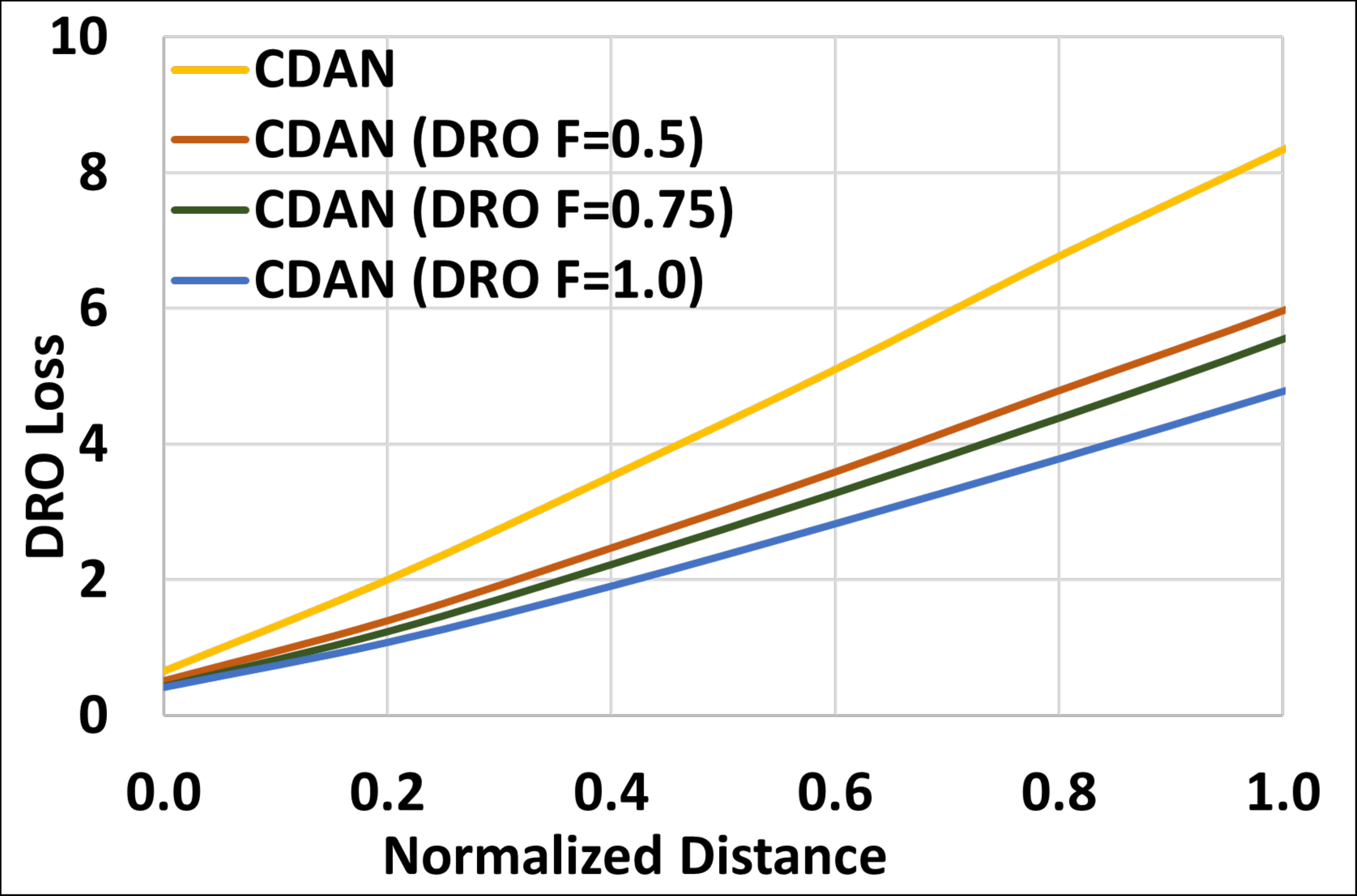}}
  \subfigure[CDAN with VLCS]{\includegraphics[width=0.32\columnwidth]{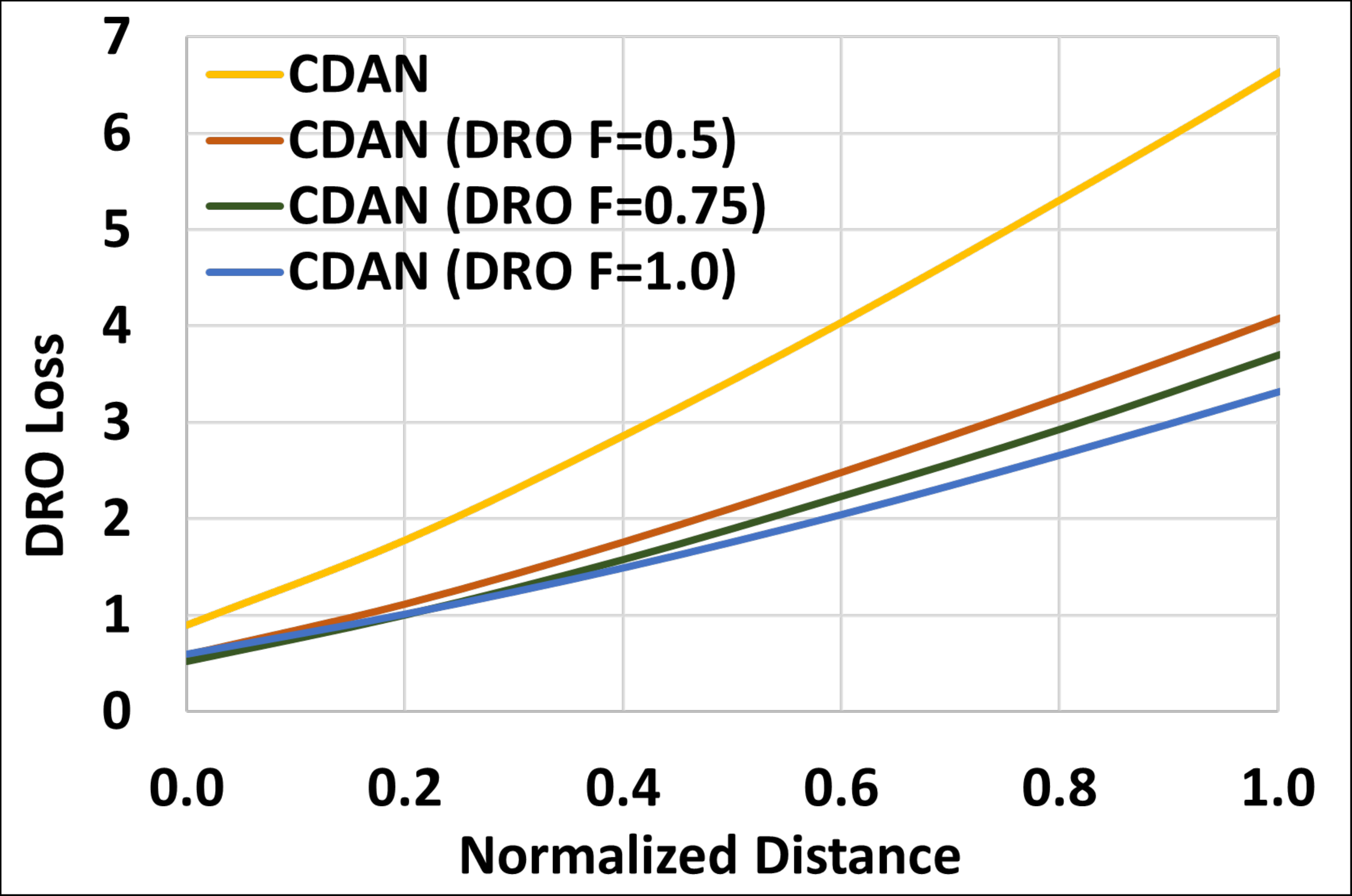}}
  
  \subfigure[VREX with R-MNIST]{\includegraphics[width=0.32\columnwidth]{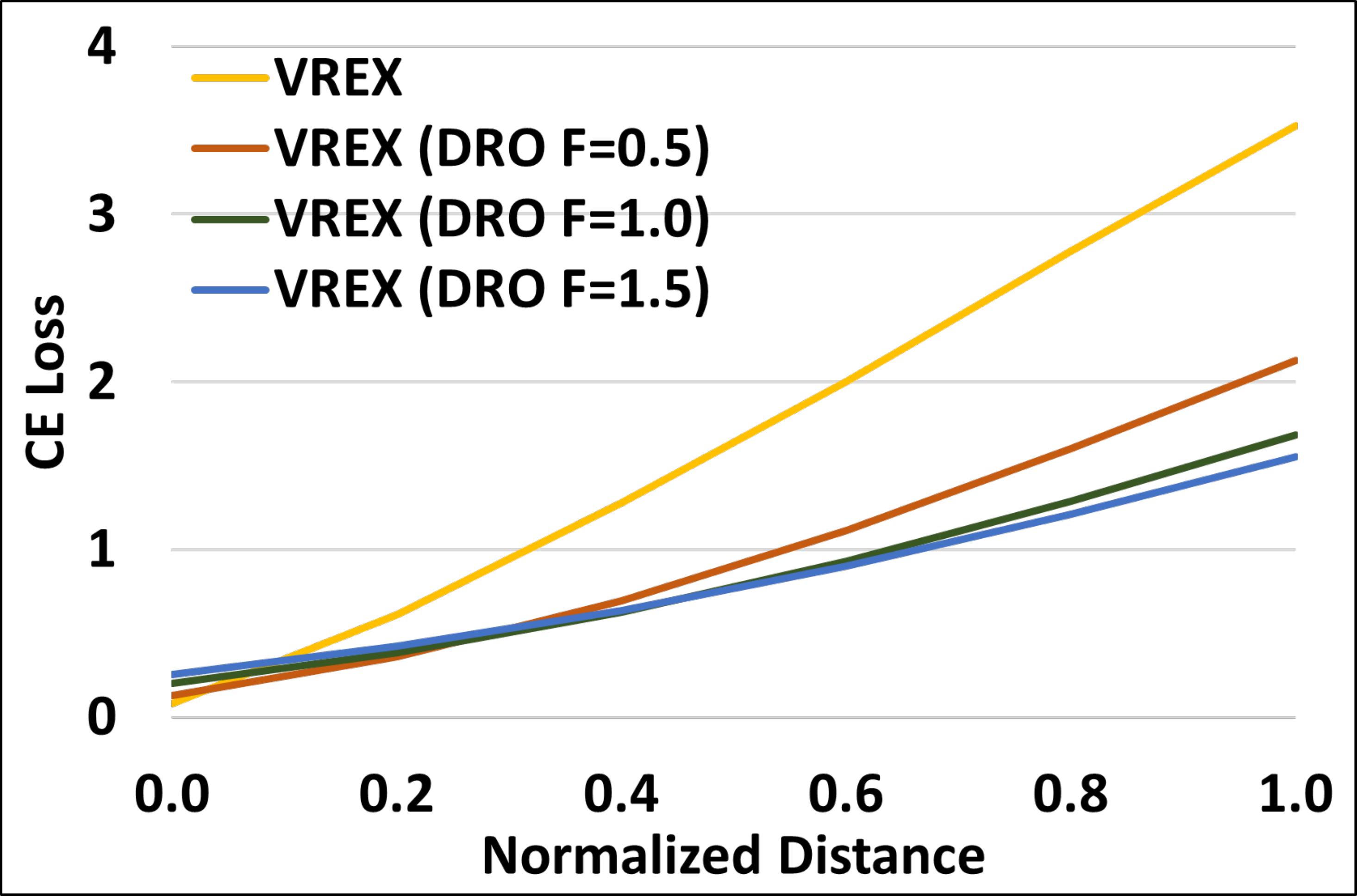}}
  \subfigure[VREX with PACS]{\includegraphics[width=0.32\columnwidth]{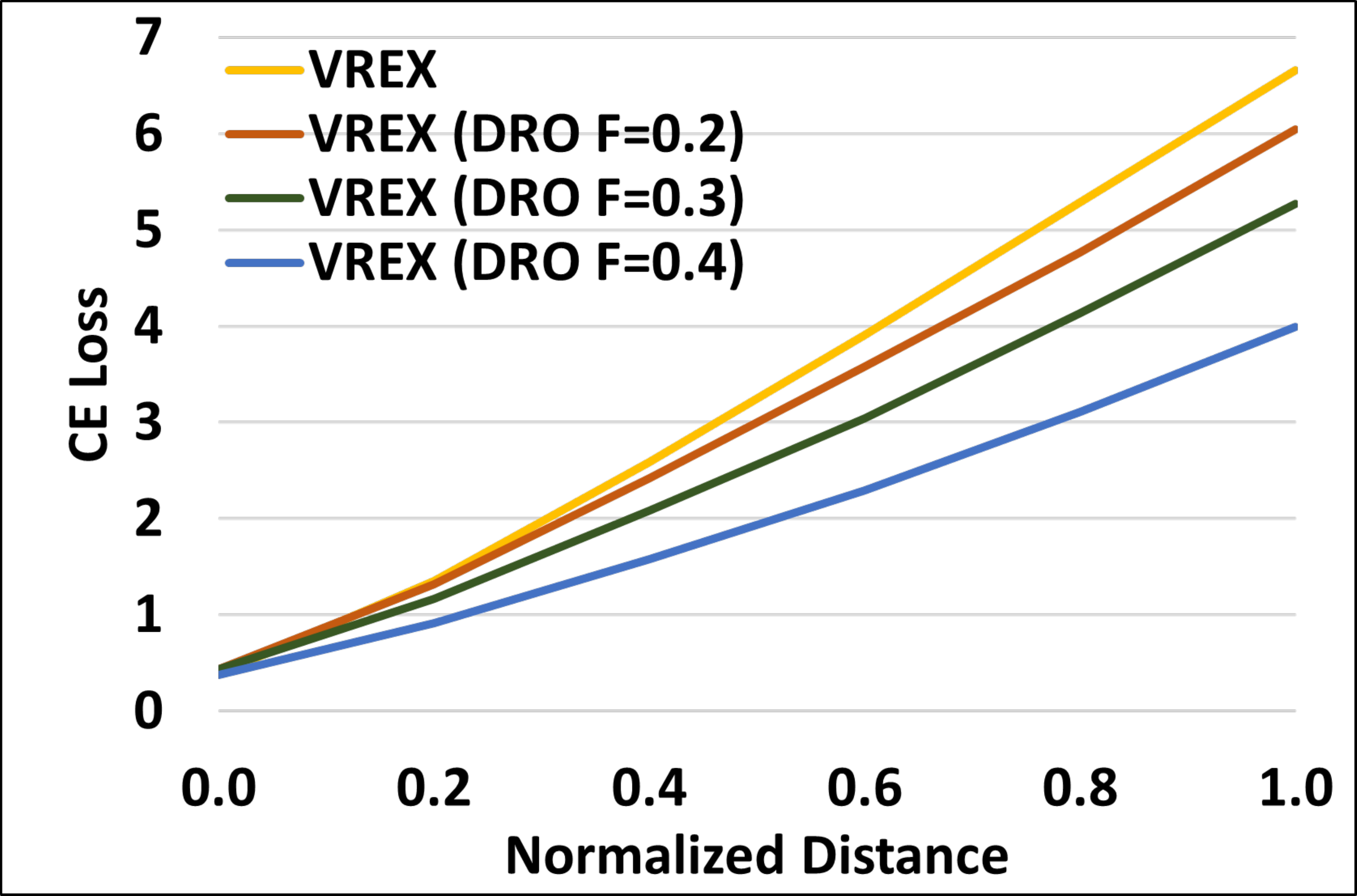}}
  \subfigure[VREX with VLCS]{\includegraphics[width=0.32\columnwidth]{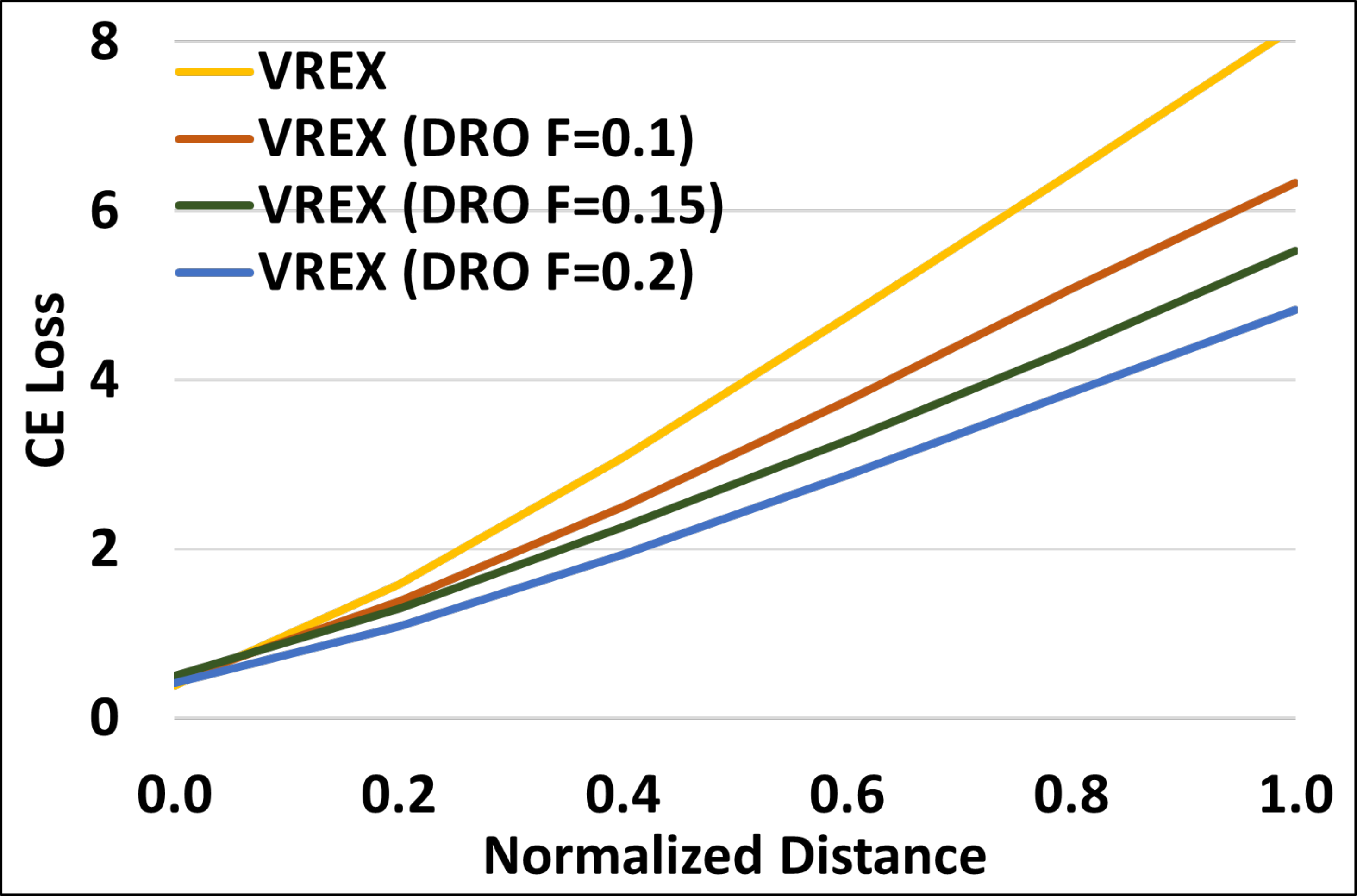}}
  \caption{(Best viewed in color.) Improved worst-case loss of models trained with CDAN \cite{long2018conditional} and VREX \cite{krueger2021out} using DR-DG proposed in Alg.~\ref{alg:dro_training} on R-MNIST, PACS and VLCS datasets. 
   }
  \label{fig:dro_trained_models_all}
\end{figure}

Here, we present the results of using our certification and training method on the VLCS \cite{fang2013unbiased} dataset. 
Last column in Fig.~\ref{fig:dro_trained_models_all} demonstrates that models trained on this dataset with different algorithms also achieve improved worst-case loss when trained with DR-DG, similar to the results presented on R-MNIST and PACS datasets in Fig.~\ref{fig:dro_trained_models}.
%Additionally, we observe similar performance improvements on this dataset with CDAN and VREX (Figs.~\ref{fig:dro_trained_models_all}) as well.
For training models on this dataset, we used Caltech101 and SUN09 as our source data. 
We observe that models trained with this setting using different (vanilla) DG methods (WM, G2DM, CDAN, and VREX) achieve comparable performance on the source test sets as well as other unseen domains in the dataset (LabelMe and VOC2007).

\subsection{Detailed certification results on models trained with DG methods}
\begin{figure}[tb]
  \centering
  \subfigure[WM on R-MNIST]{\includegraphics[width=0.32\columnwidth]{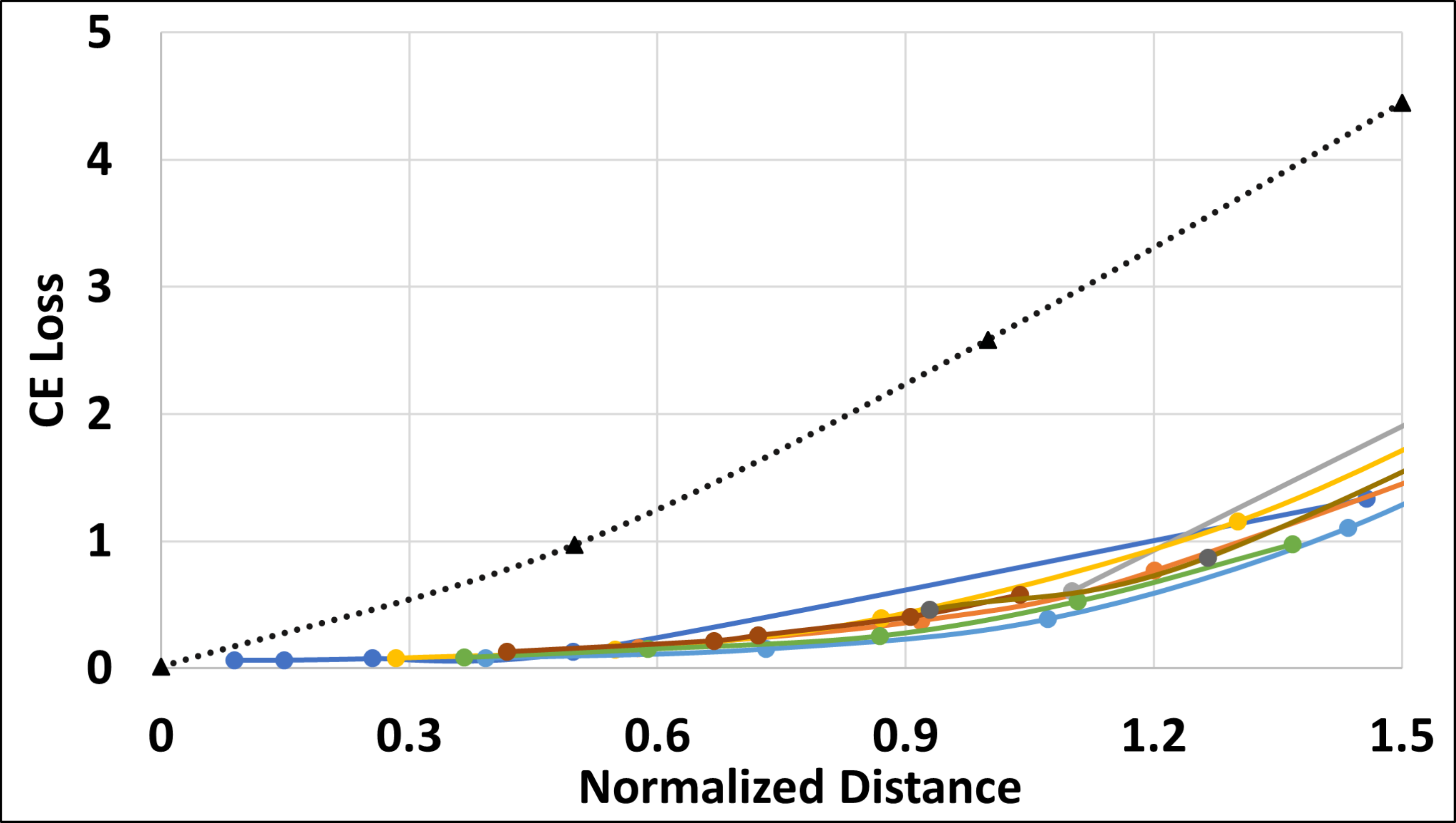}}
  \subfigure[WM on PACS]{\includegraphics[width=0.32\columnwidth]{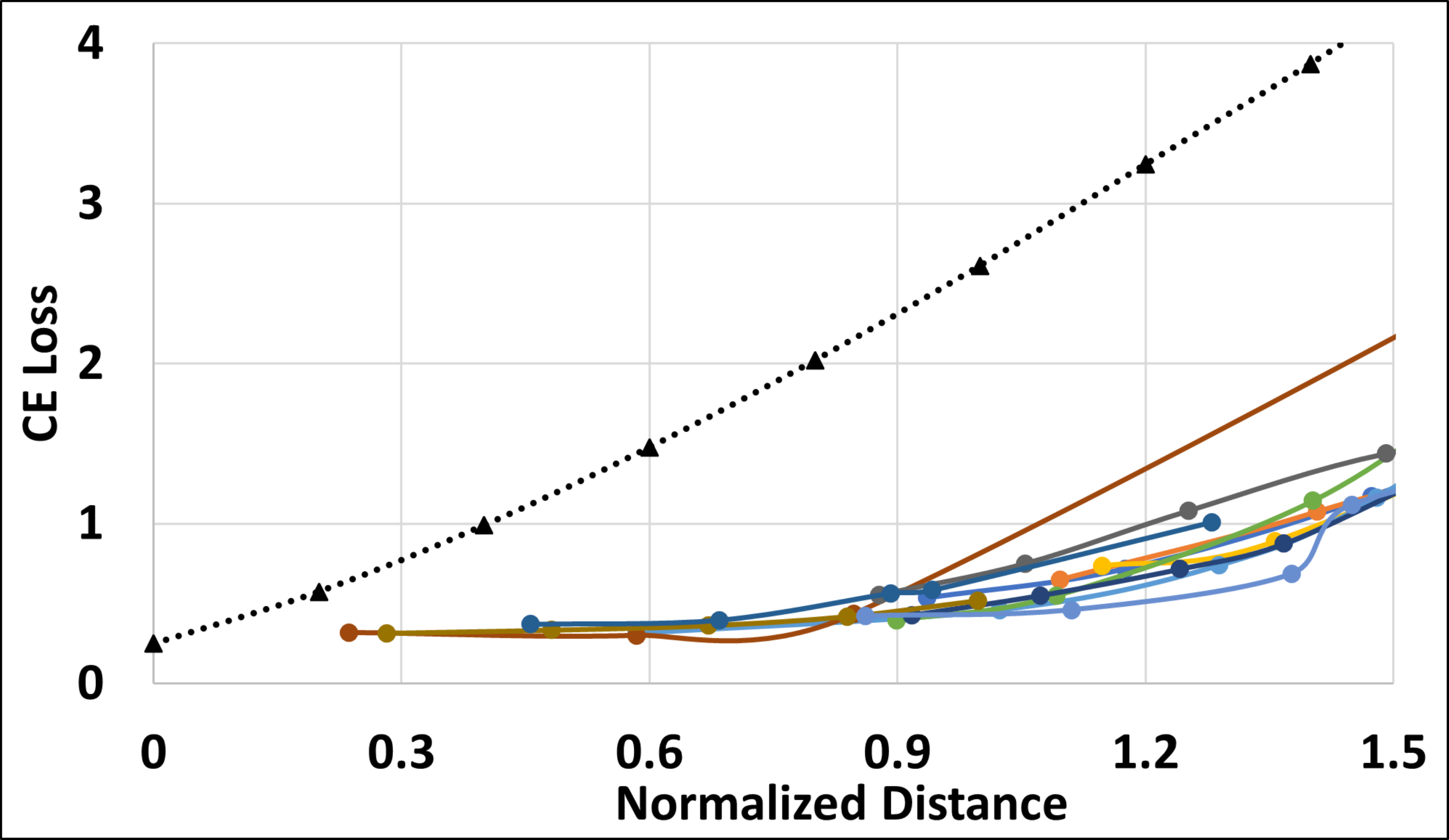}}
  \subfigure[WM on VLCS]{\includegraphics[width=0.32\columnwidth]{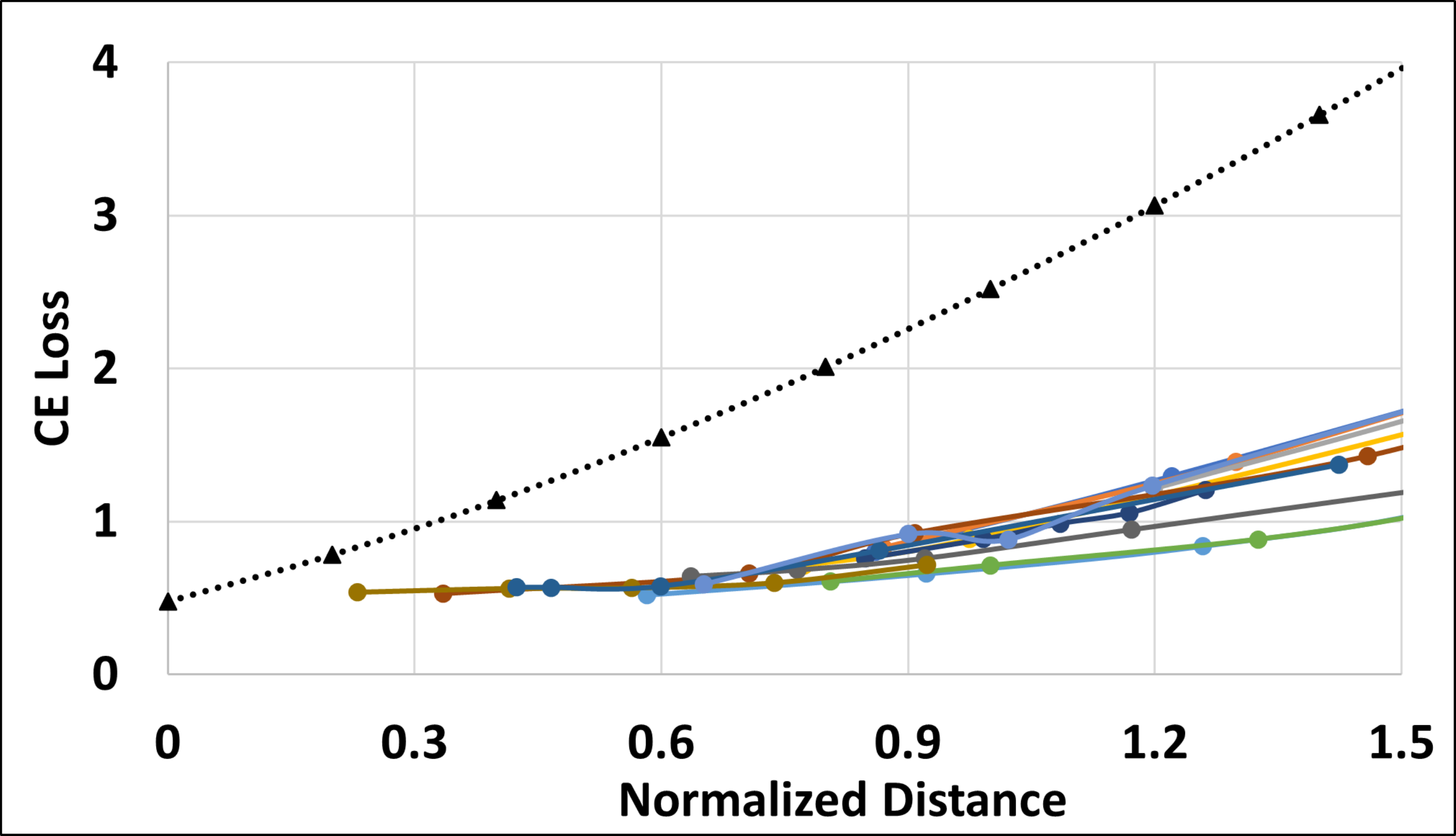}}
  
  \subfigure[DR-DG ($F$ = 0.5) with WM on R-MNIST]{\includegraphics[width=0.32\columnwidth]{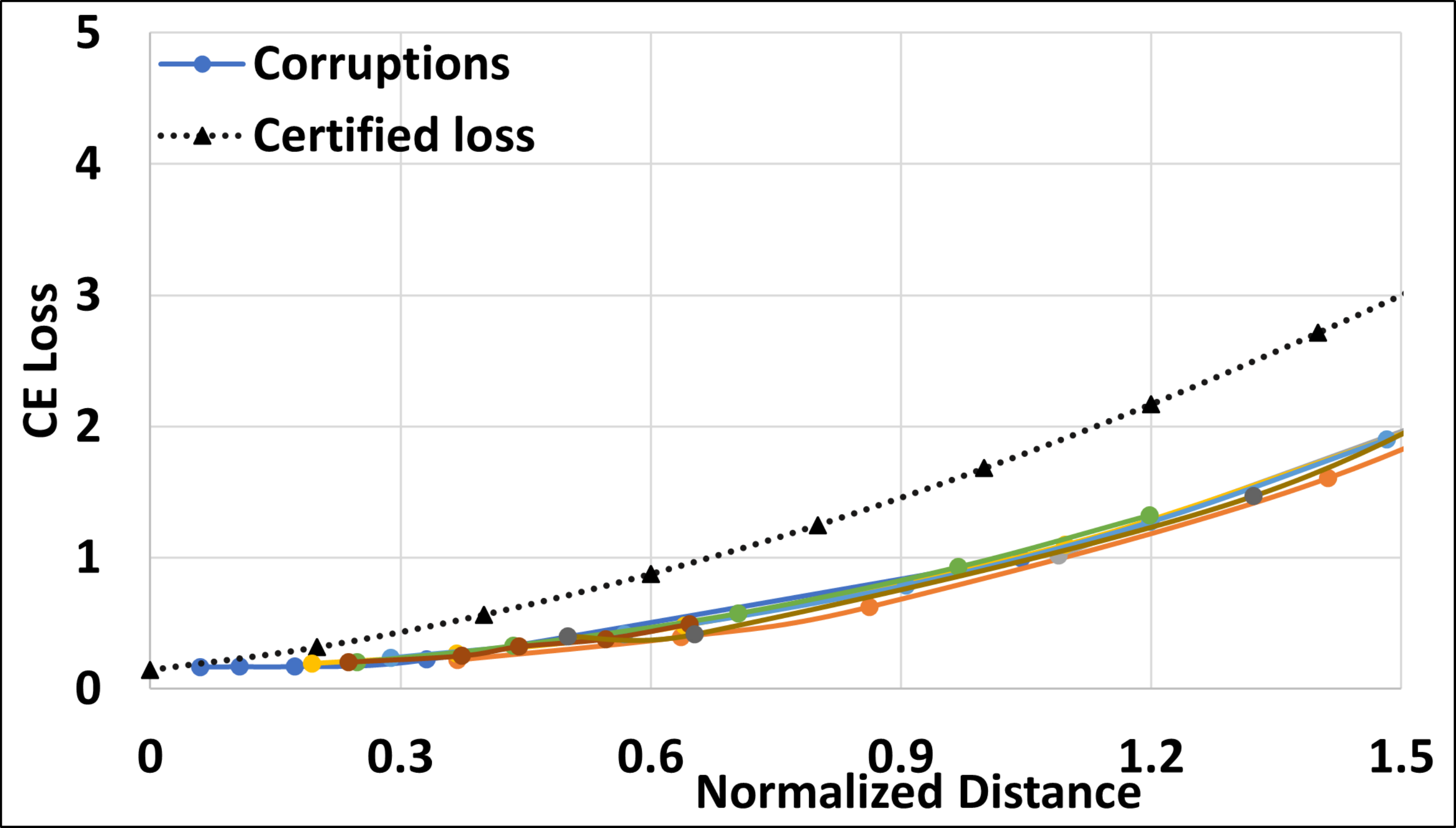}}
  \subfigure[DR-DG ($F$ = 0.75) with WM on PACS]{\includegraphics[width=0.32\columnwidth]{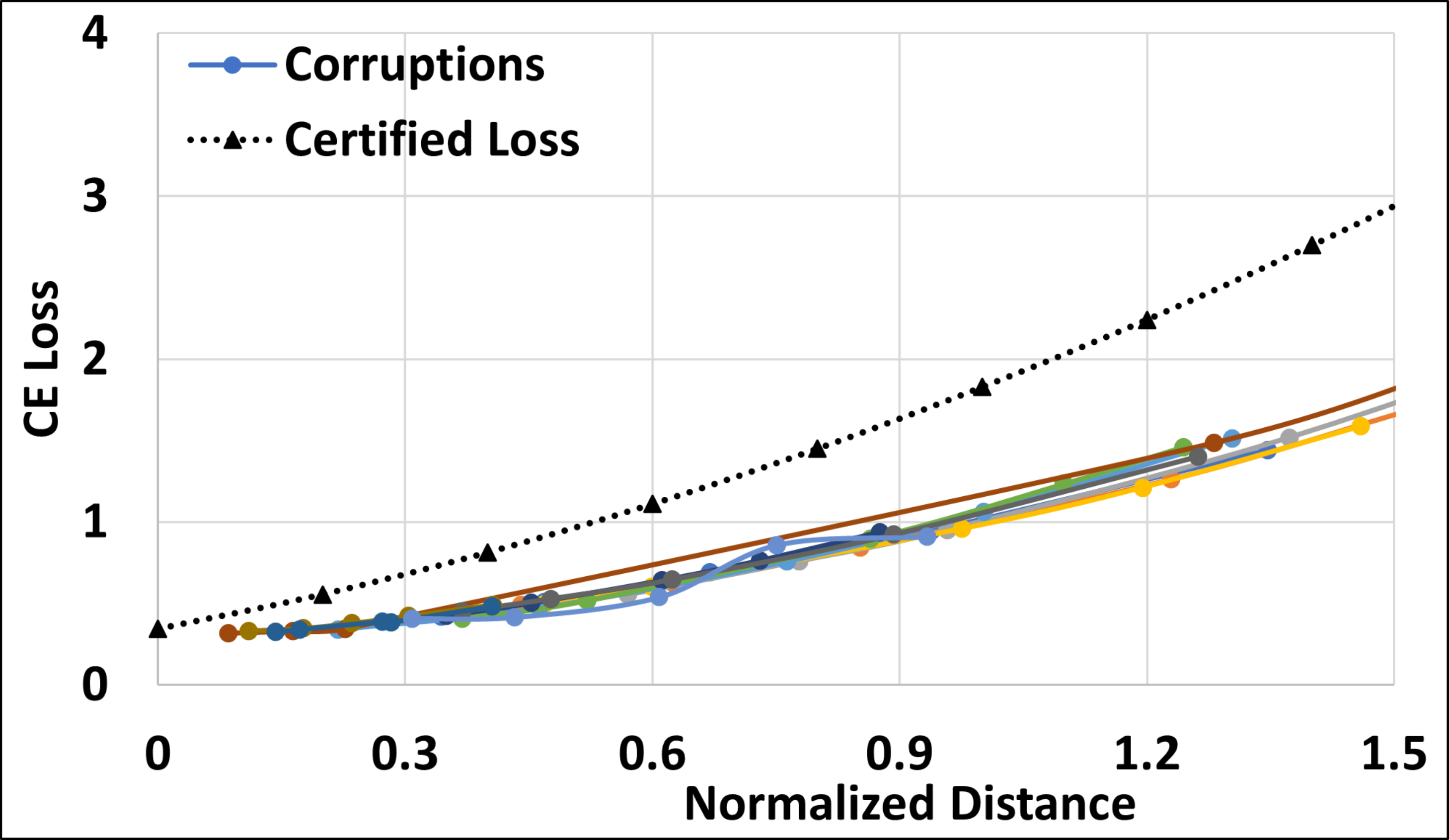}}
  \subfigure[DR-DG ($F$ = 0.75) with WM on VLCS]{\includegraphics[width=0.32\columnwidth]{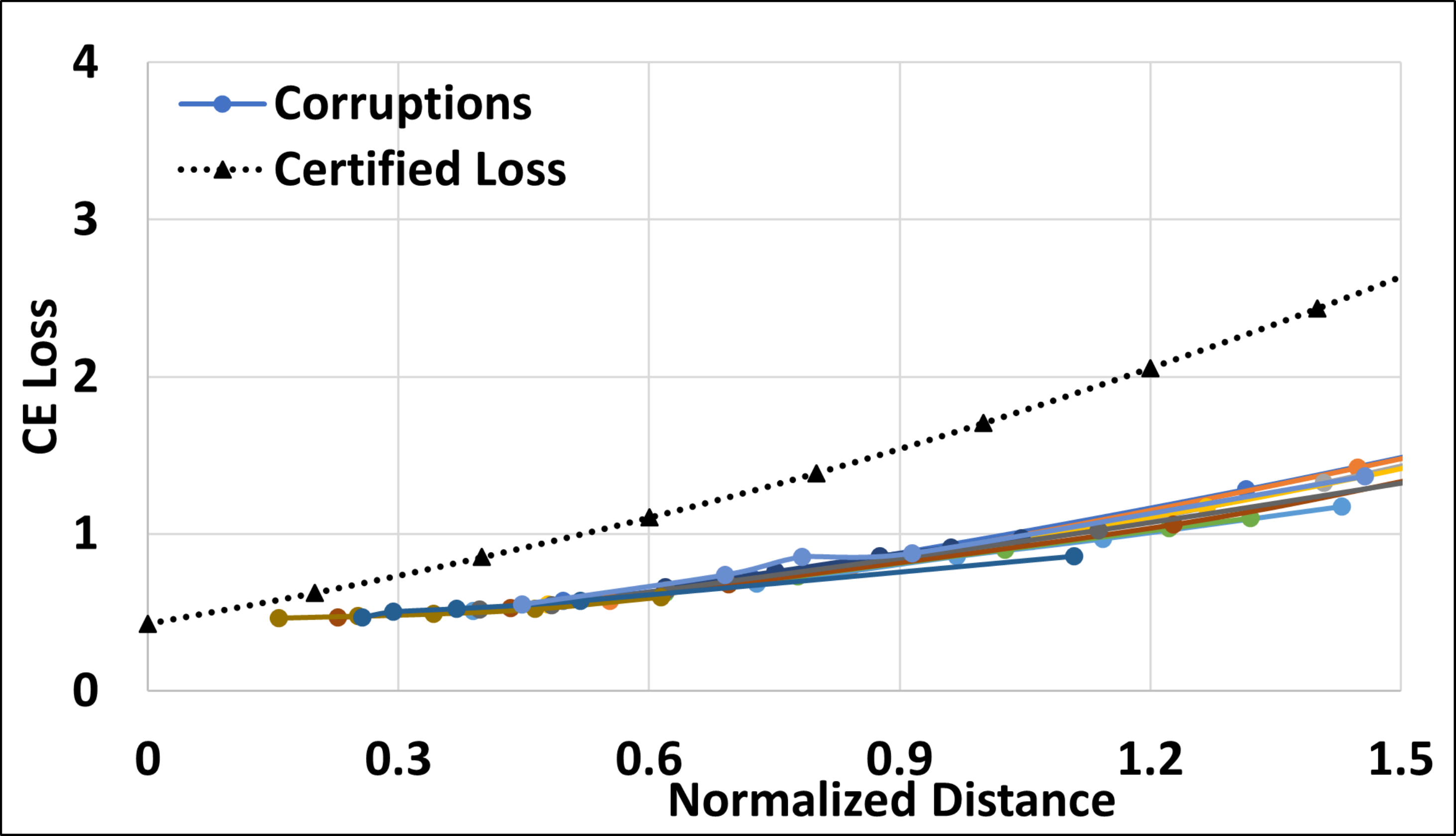}}
  
  \subfigure[G2DM on R-MNIST]{\includegraphics[width=0.32\columnwidth]{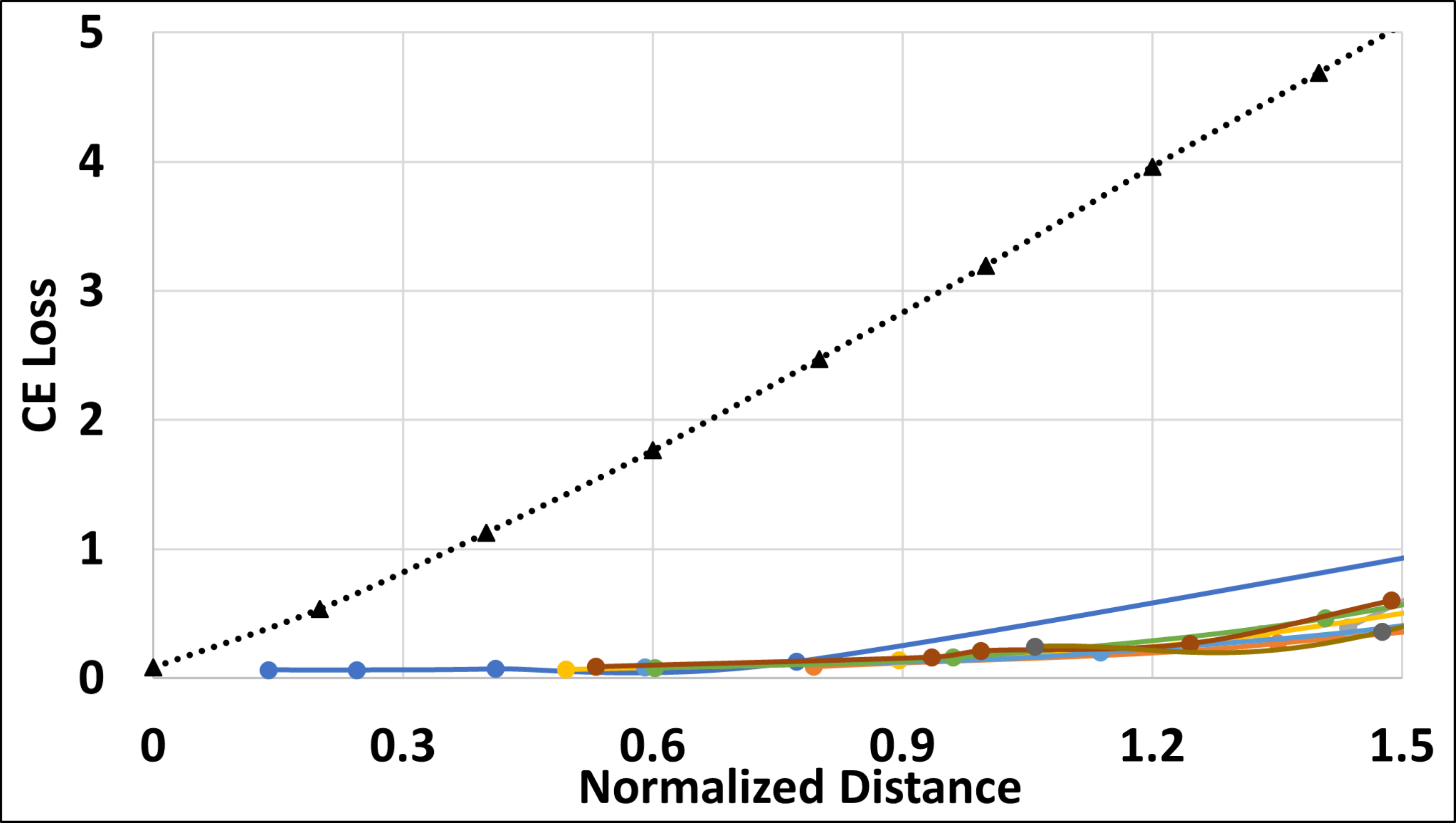}}
  \subfigure[G2DM on PACS]{\includegraphics[width=0.32\columnwidth]{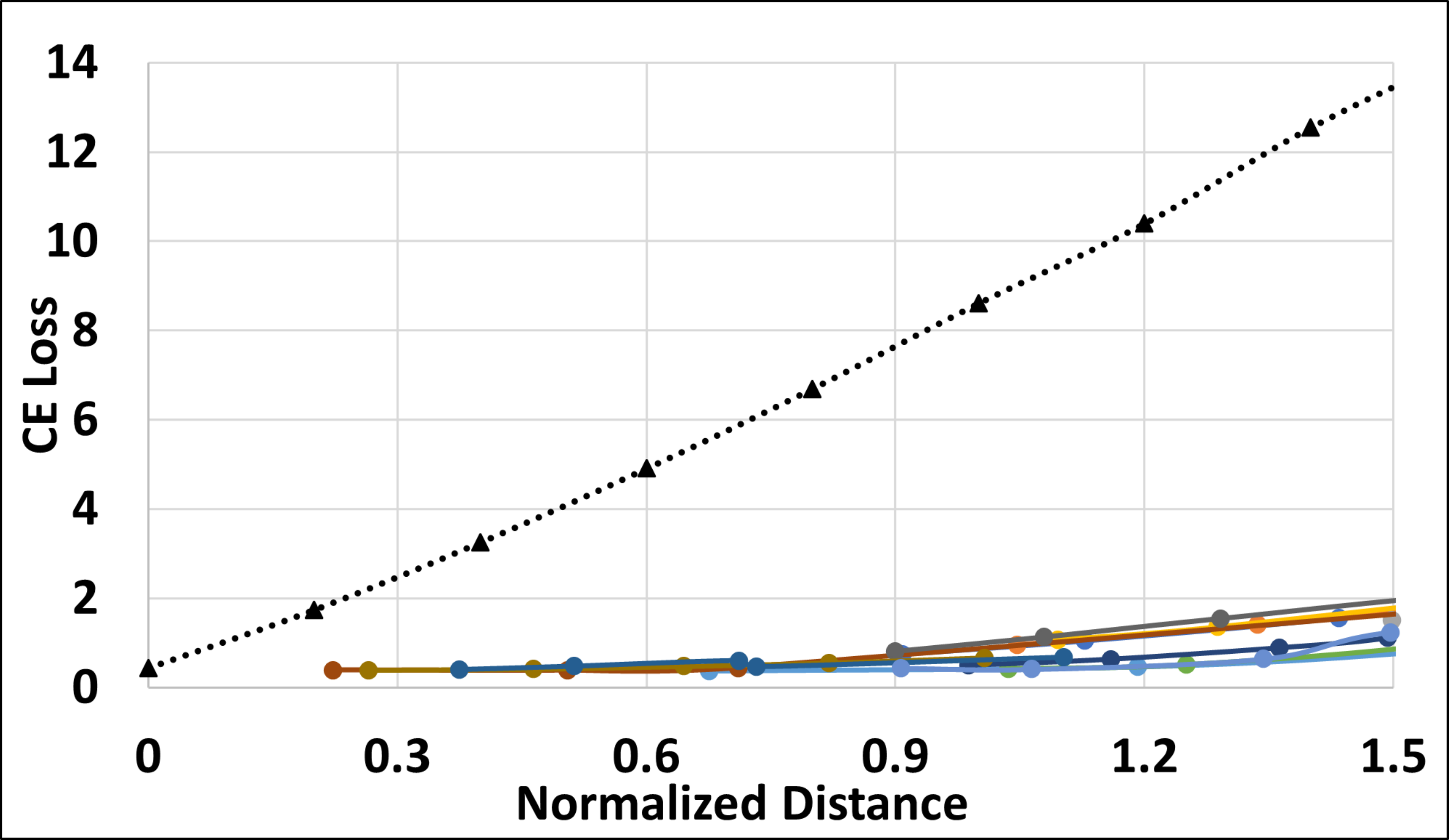}}
  \subfigure[G2DM on VLCS]{\includegraphics[width=0.32\columnwidth]{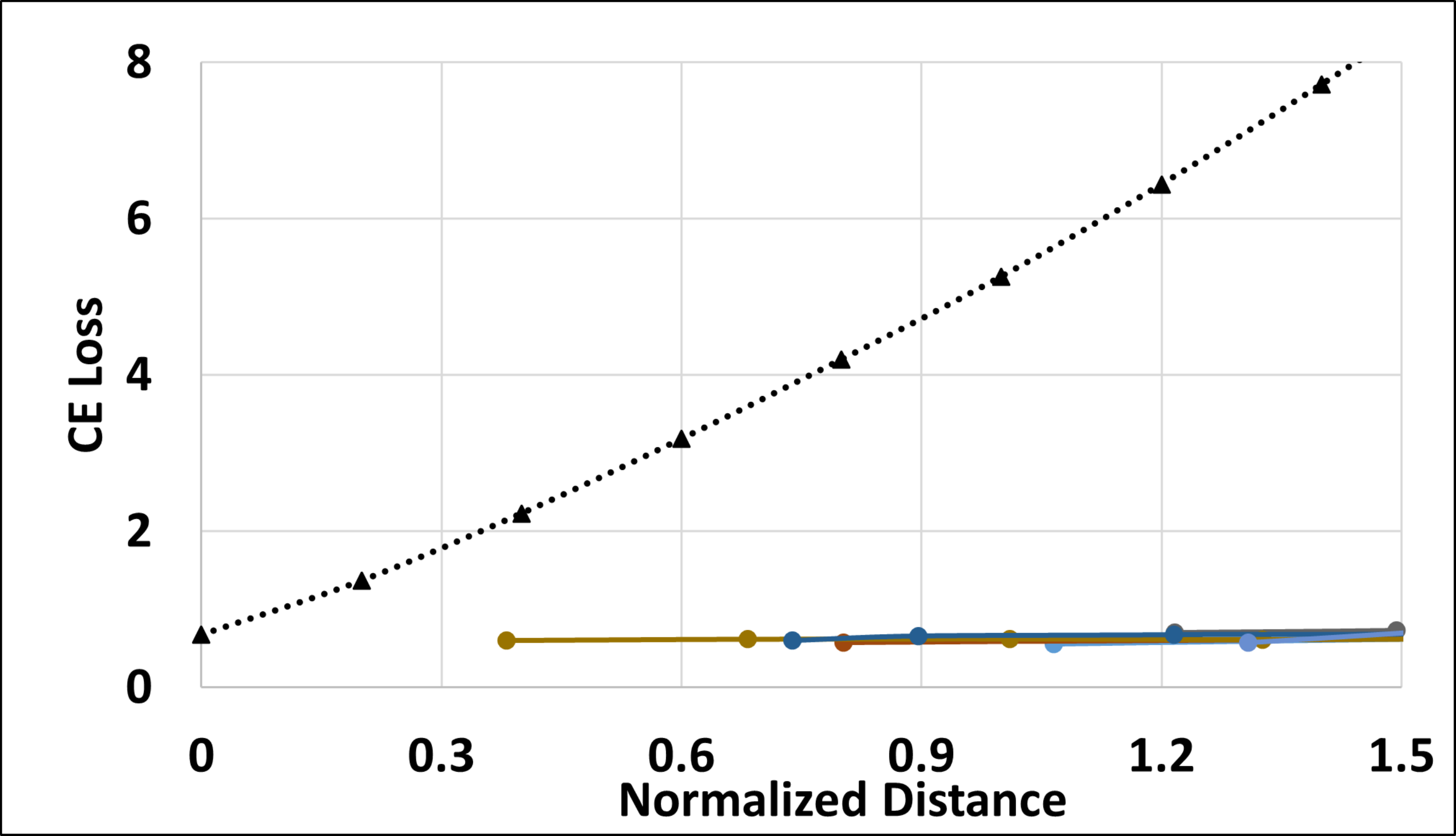}}
  
  \subfigure[DR-DG ($F$ = 1.5) with G2DM on R-MNIST]{\includegraphics[width=0.32\columnwidth]{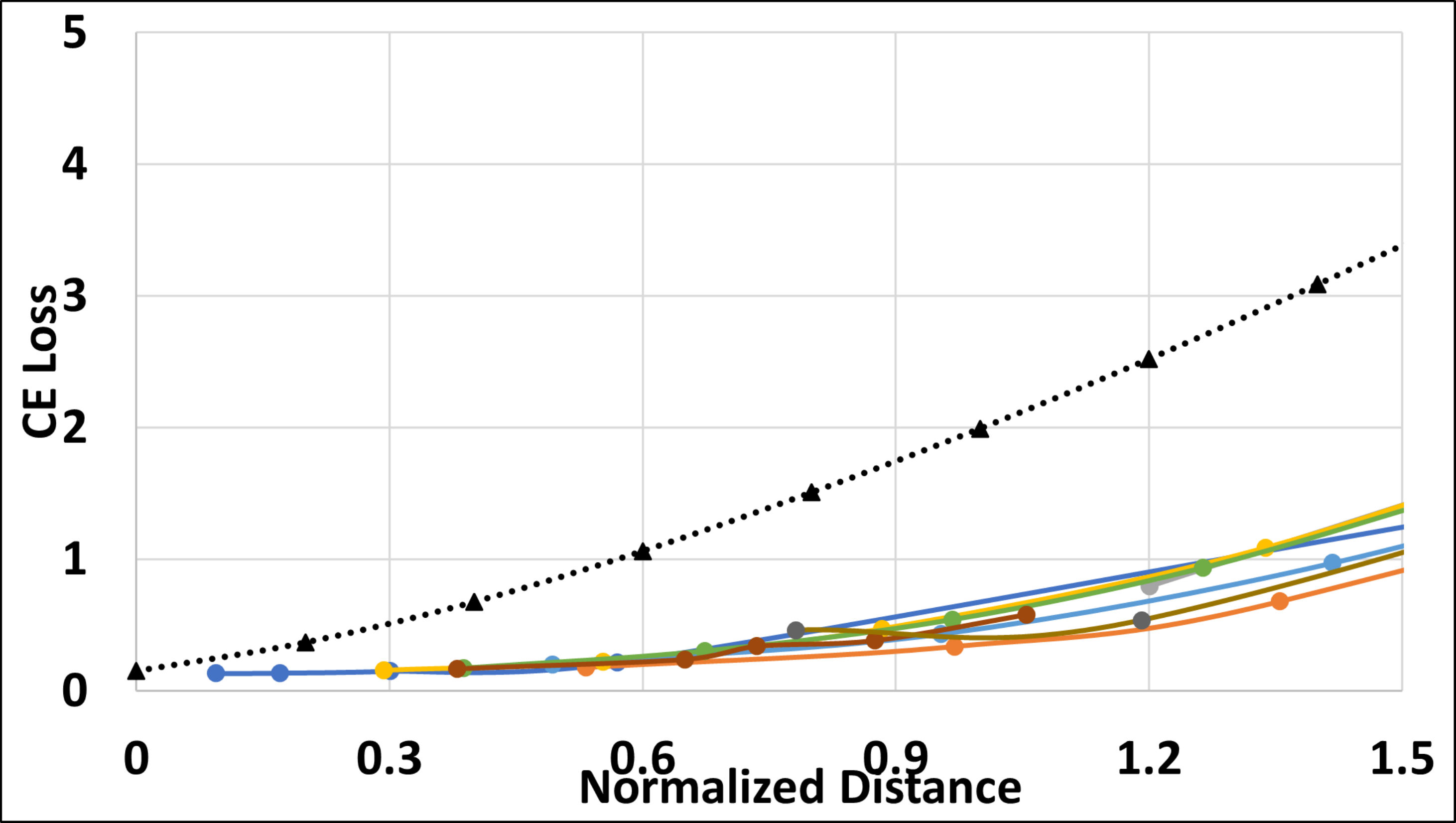}}
  \subfigure[DR-DG ($F$ = 0.75) with G2DM on PACS]{\includegraphics[width=0.32\columnwidth]{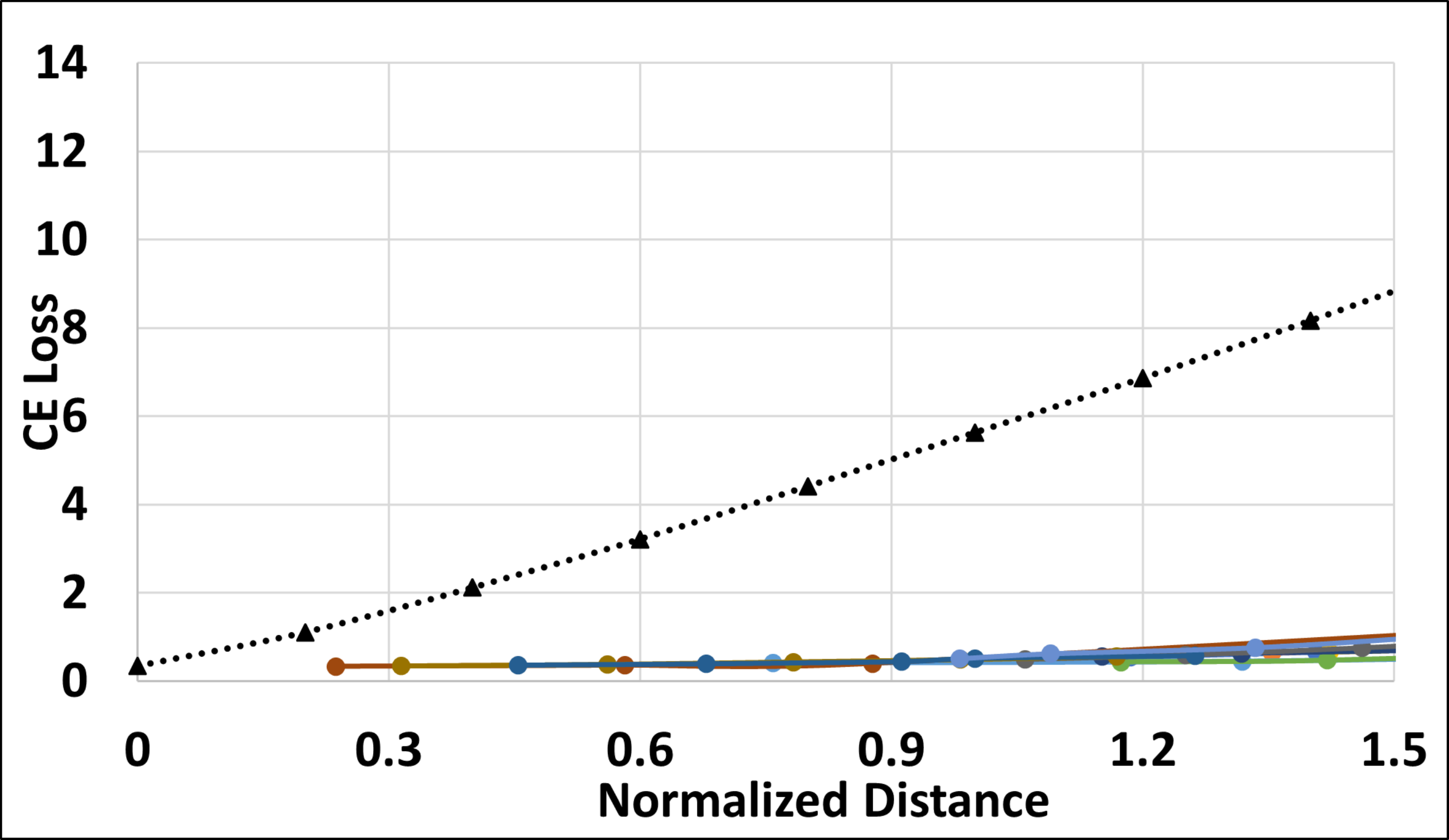}}
  \subfigure[DR-DG ($F$ = 0.4) with G2DM on VLCS]{\includegraphics[width=0.32\columnwidth]{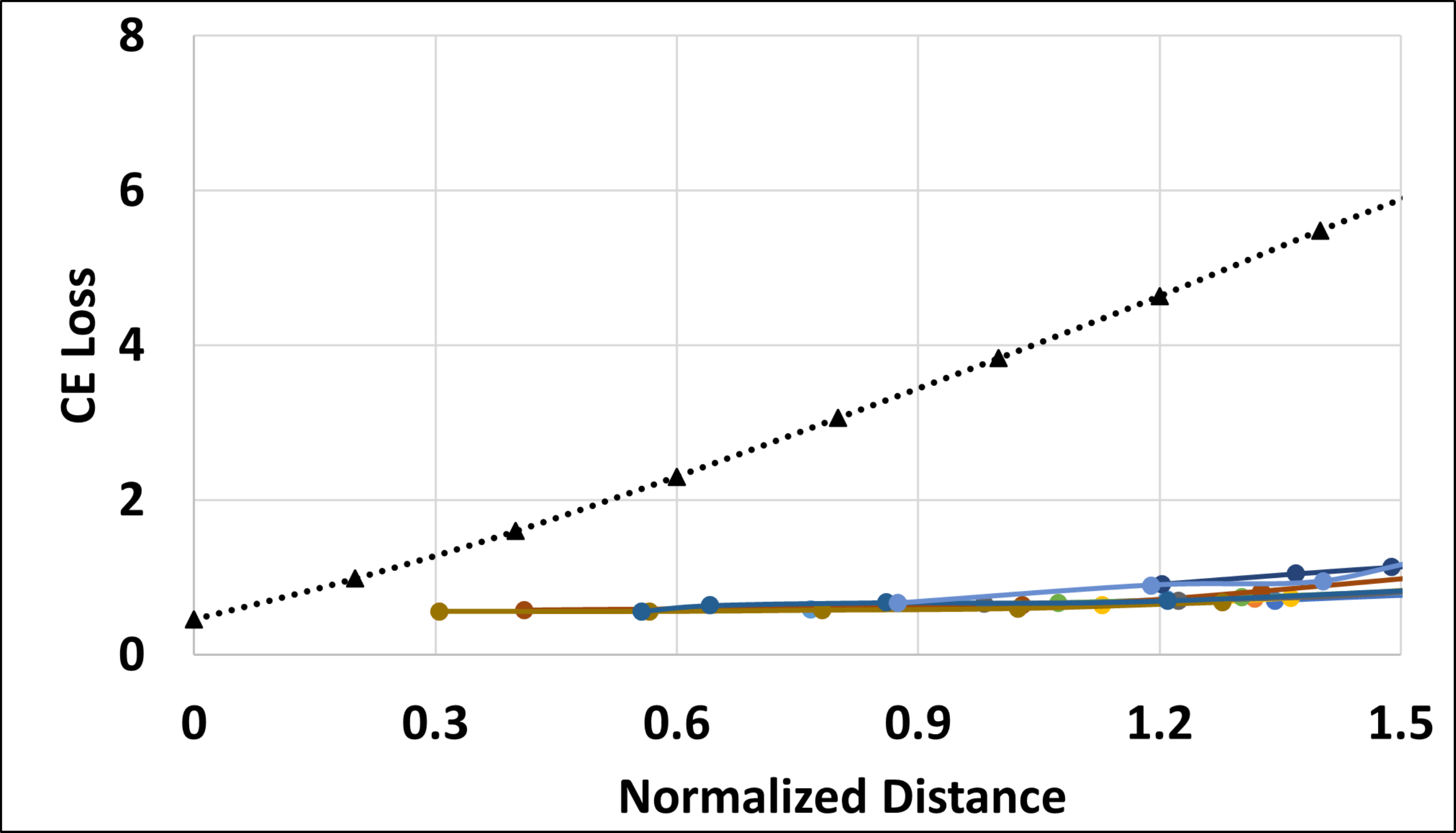}}

  \caption{(Best viewed in color.) Comparison of the certified (worst-case) loss of the models trained with WM and G2DM on R-MNIST, PACS, and VLCS dataset. Rows 1 and 3 show models trained with Vanilla DG methods and rows 2 and 4 show models trained with DR-DG using additional losses from WM and G2DM, respectively. The models trained with DR-DG incur smaller certified loss compared to their vanilla counterparts and only slightly higher loss on unseen distributions created through common corruptions.}
  \label{fig:loss_before_after_wm_g2dm}
\end{figure}

\begin{figure}[tb]
  \centering
  \subfigure[R-MNIST]{\includegraphics[width=0.32\columnwidth]{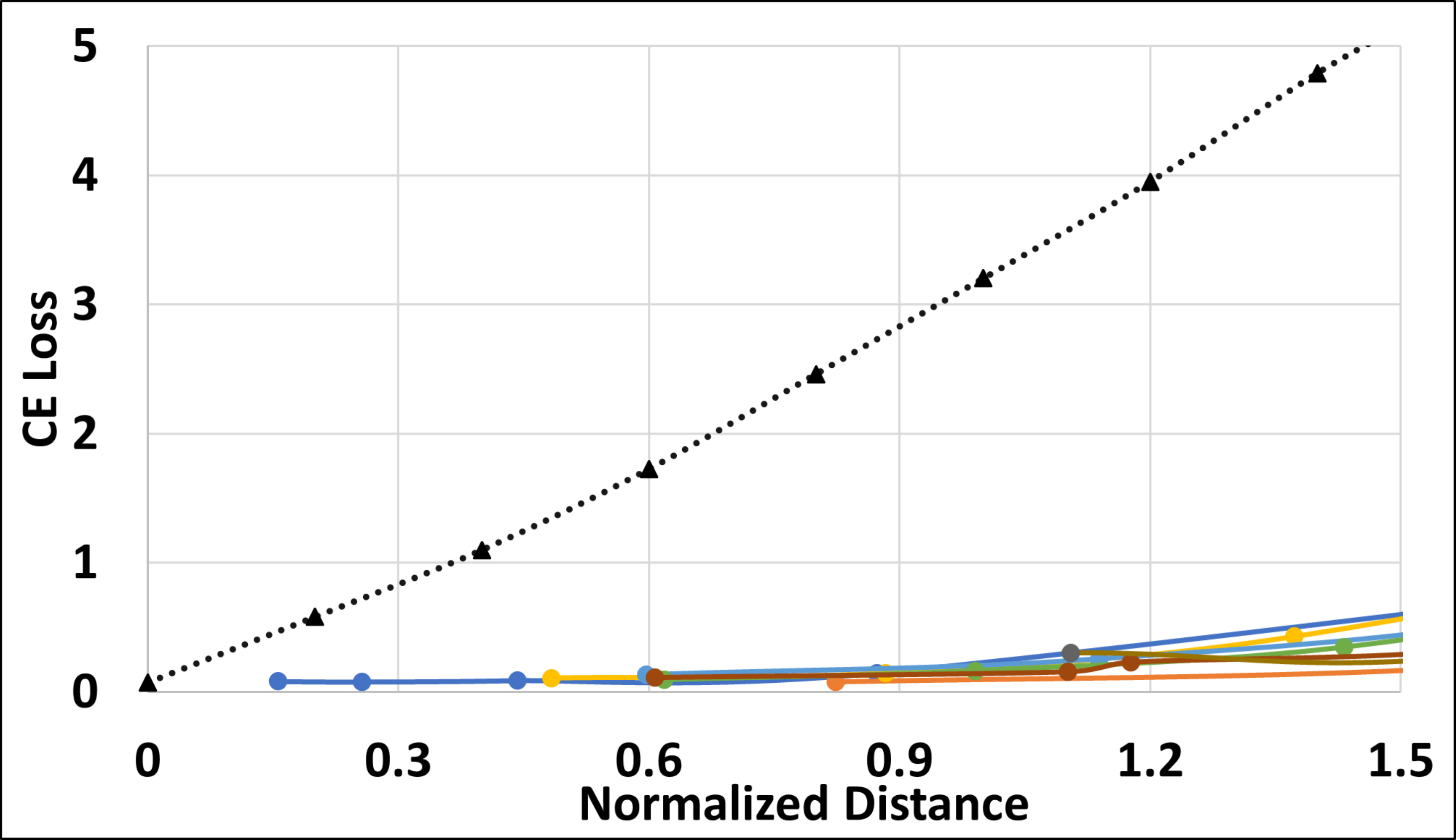}}
  \subfigure[PACS]{\includegraphics[width=0.32\columnwidth]{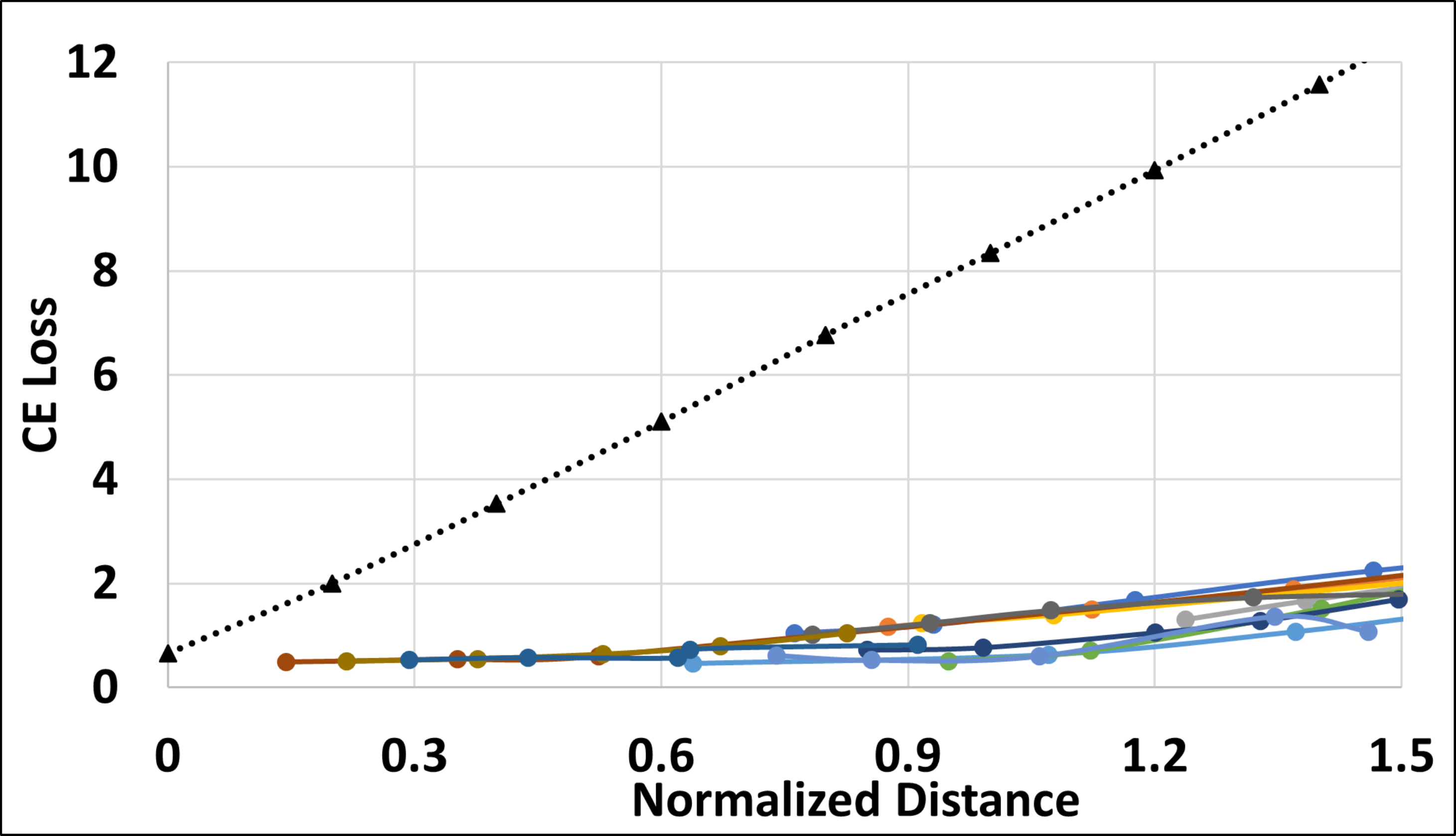}}
  \subfigure[VLCS]{\includegraphics[width=0.32\columnwidth]{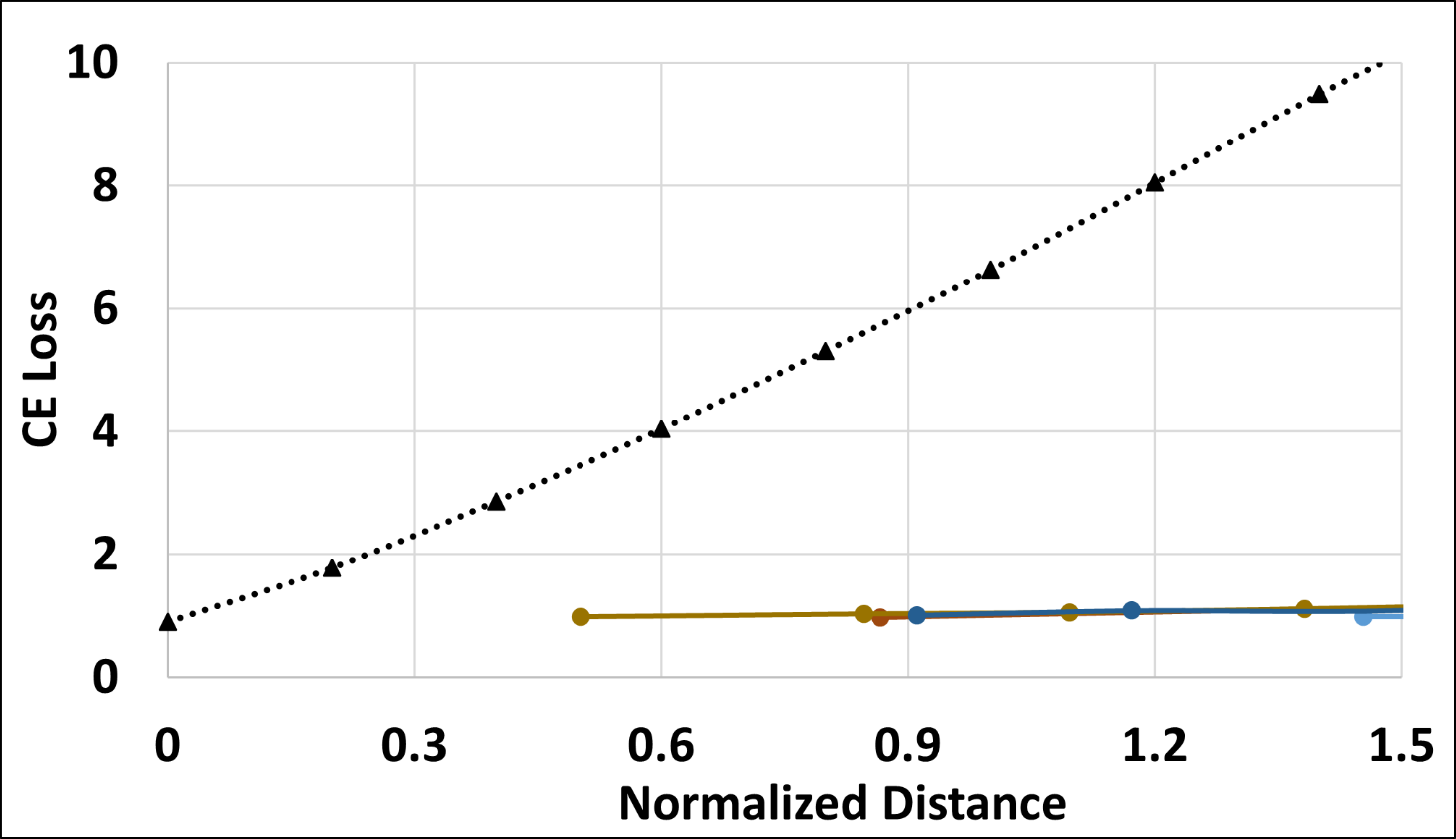}}
  
  \subfigure[DR-DG ($F$ = 1.5) on R-MNIST ]{\includegraphics[width=0.32\columnwidth]{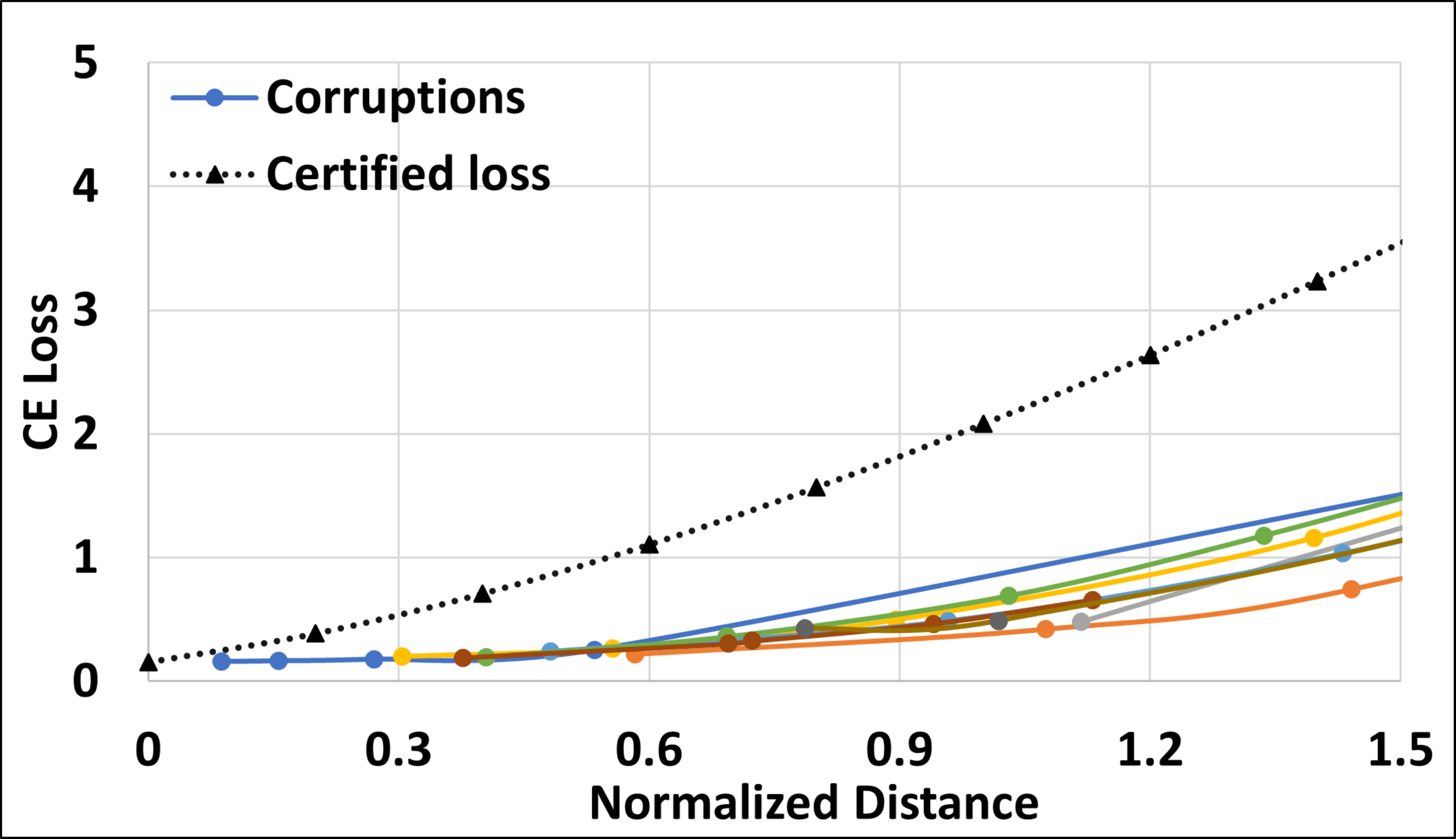}}
  \subfigure[DR-DG ($F=0.5$) on PACS]{\includegraphics[width=0.32\columnwidth]{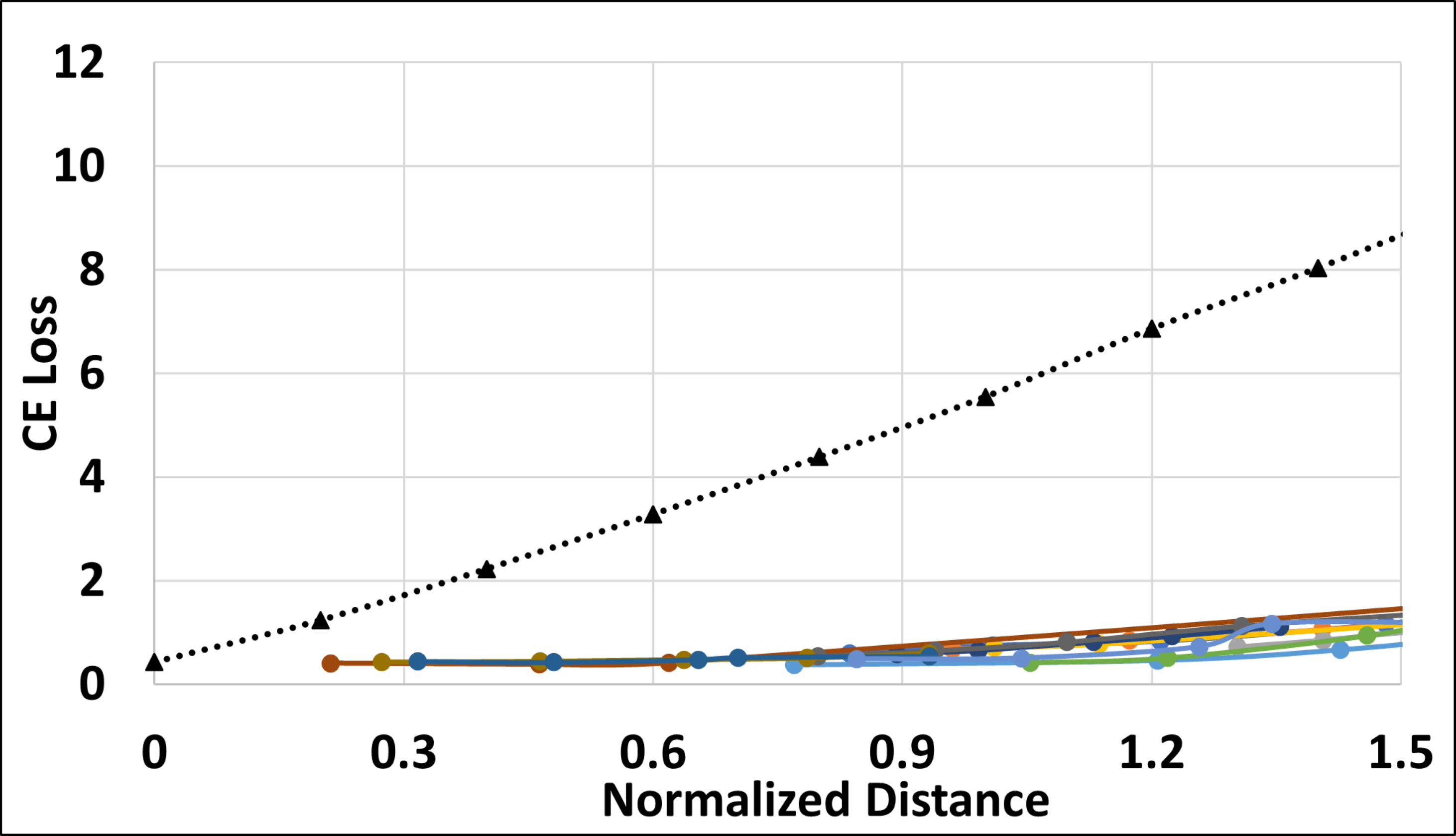}}
  \subfigure[DR-DG ($F=1.0$) on VLCS]{\includegraphics[width=0.32\columnwidth]{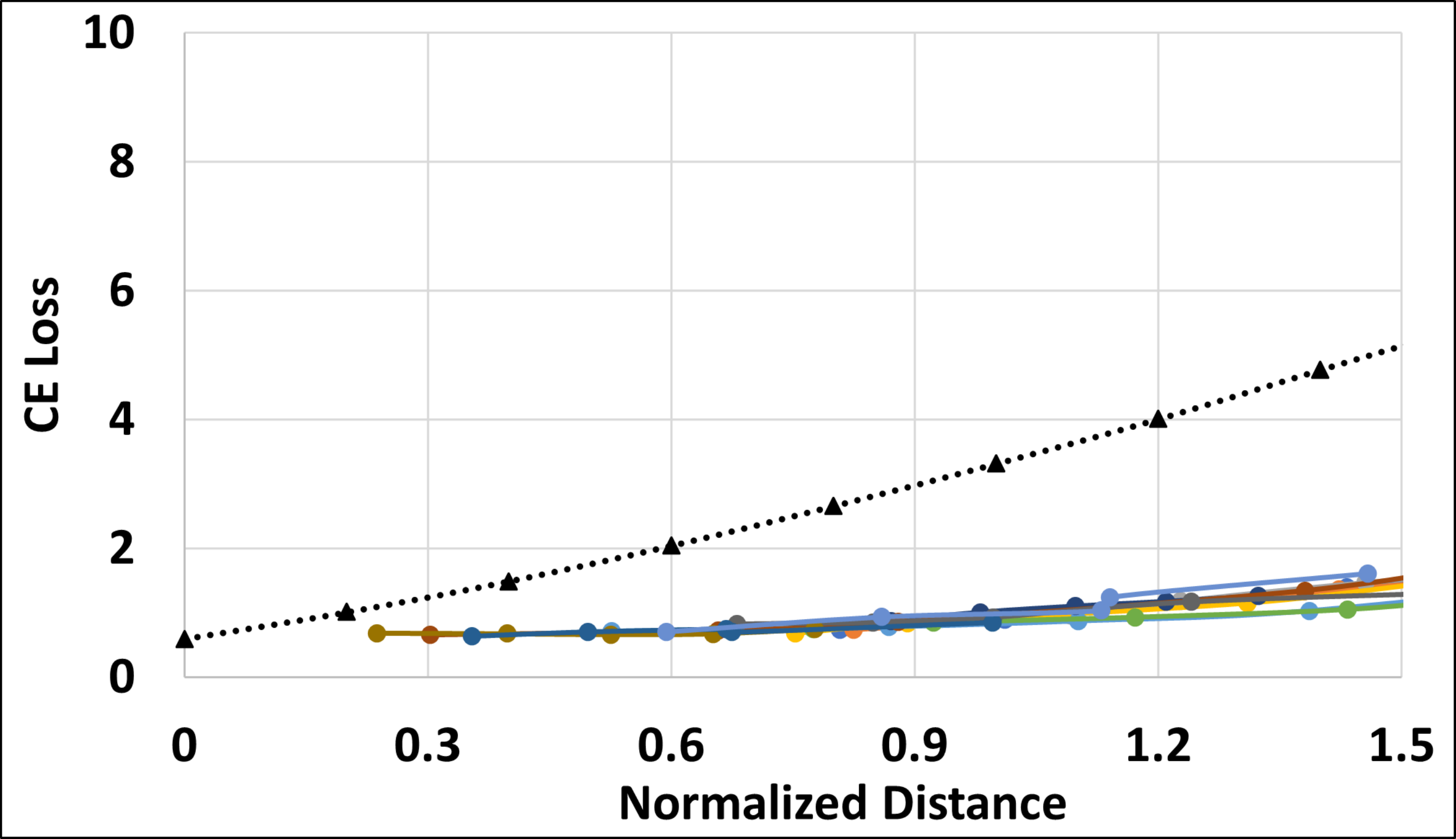}}
  
  \subfigure[R-MNIST]{\includegraphics[width=0.32\columnwidth]{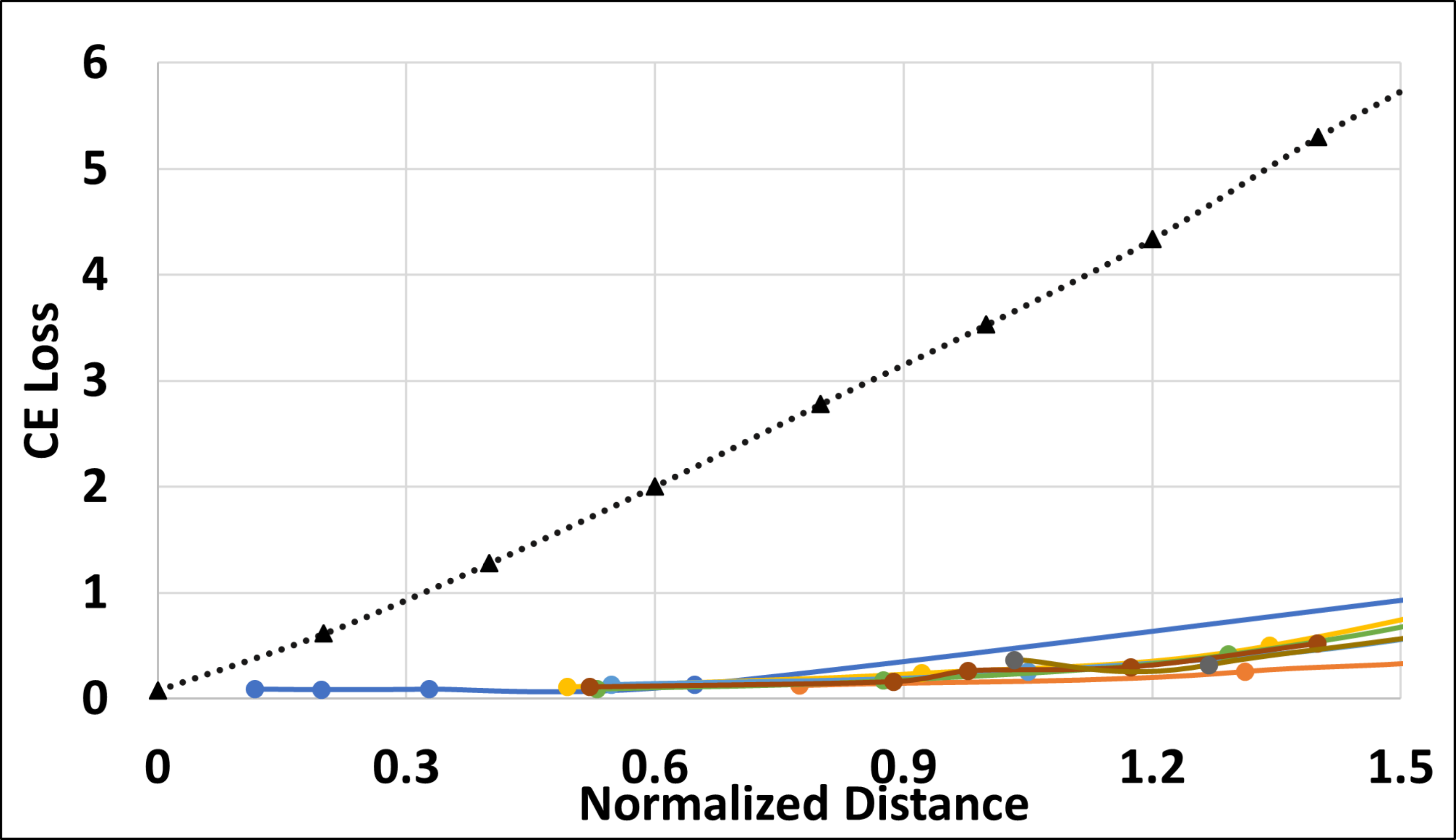}}
  \subfigure[PACS]{\includegraphics[width=0.32\columnwidth]{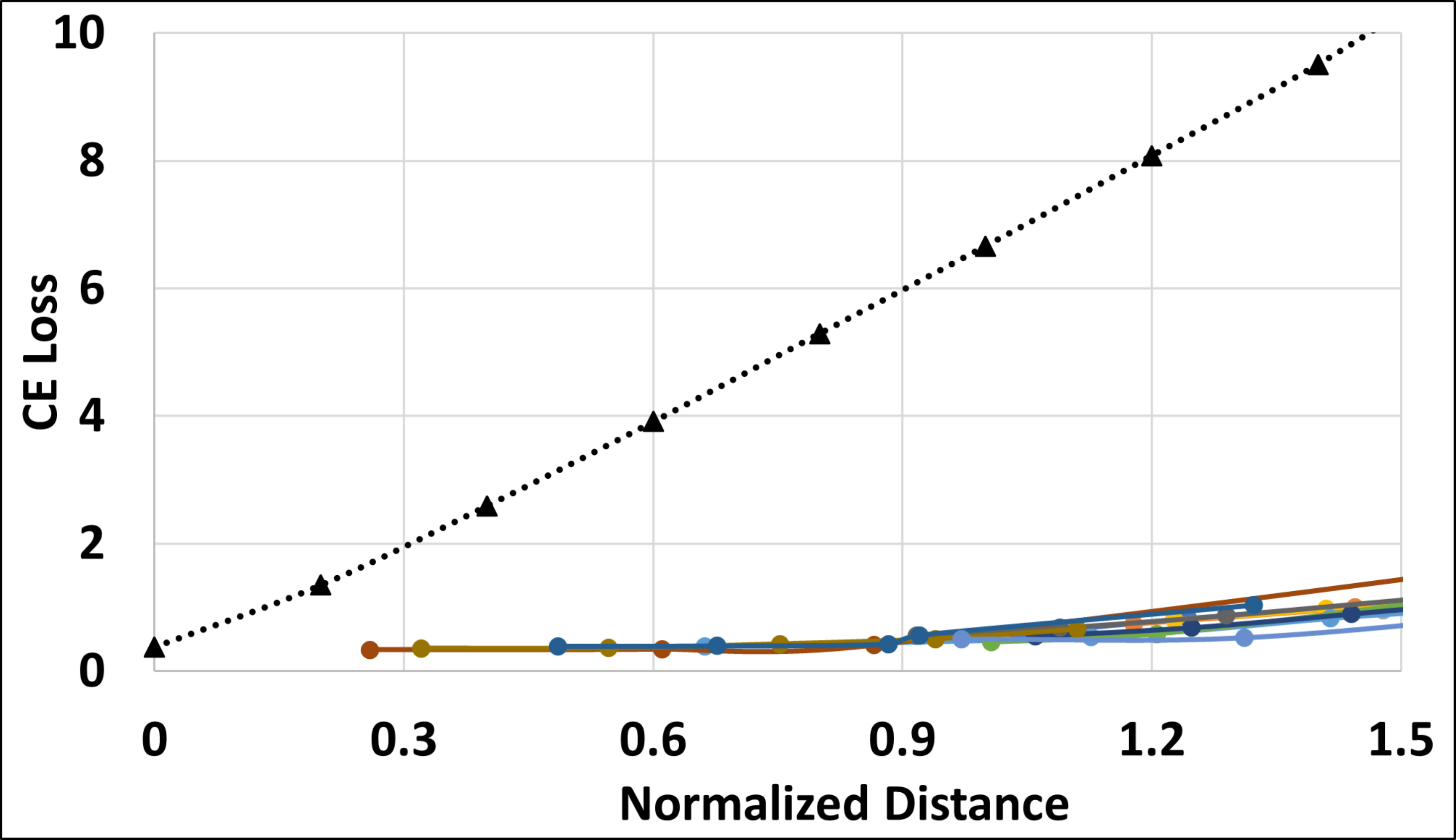}}
  \subfigure[VLCS]{\includegraphics[width=0.32\columnwidth]{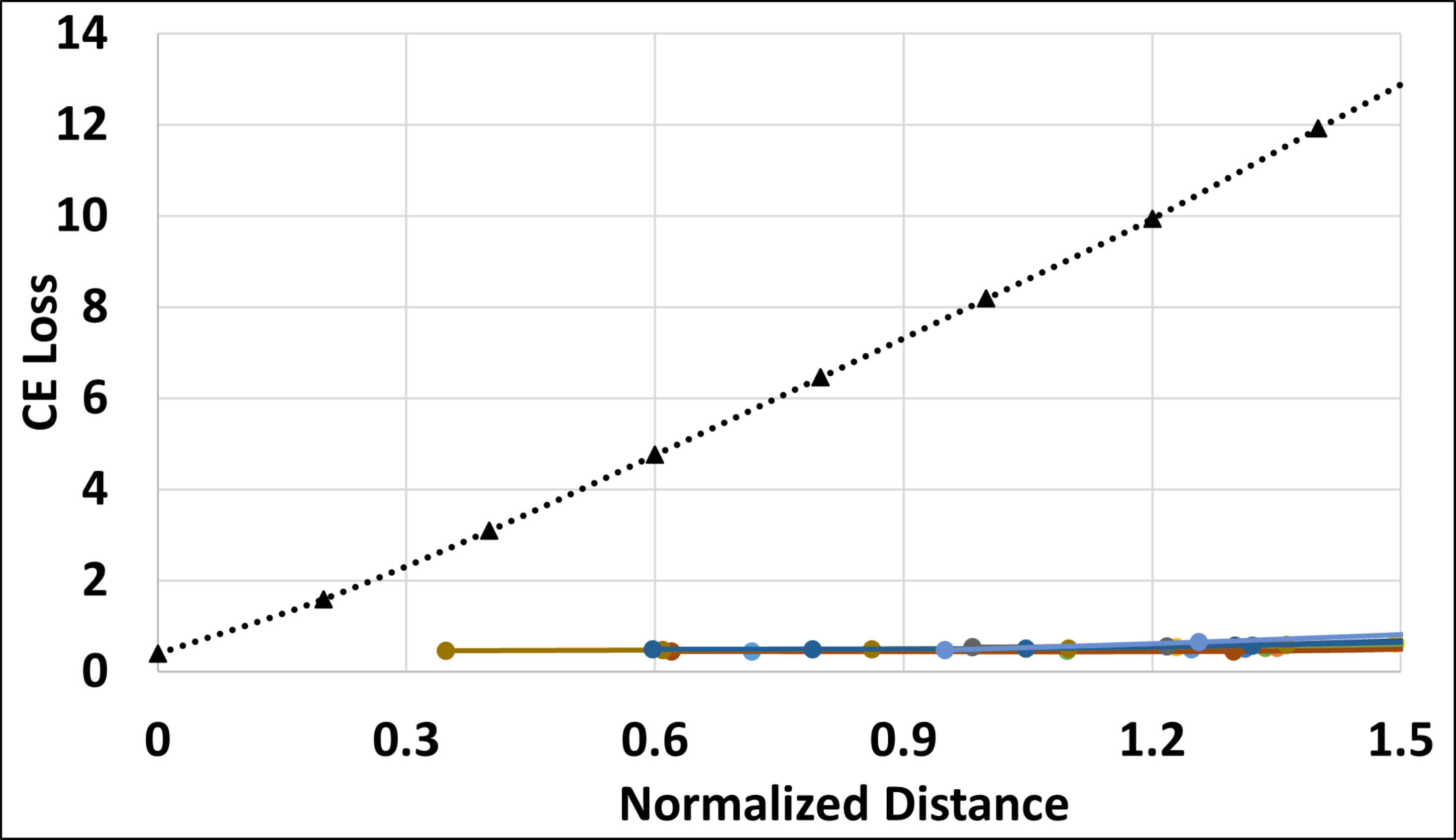}}
  
  \subfigure[DR-DG ($F$ = 1.5) on R-MNIST]{\includegraphics[width=0.32\columnwidth]{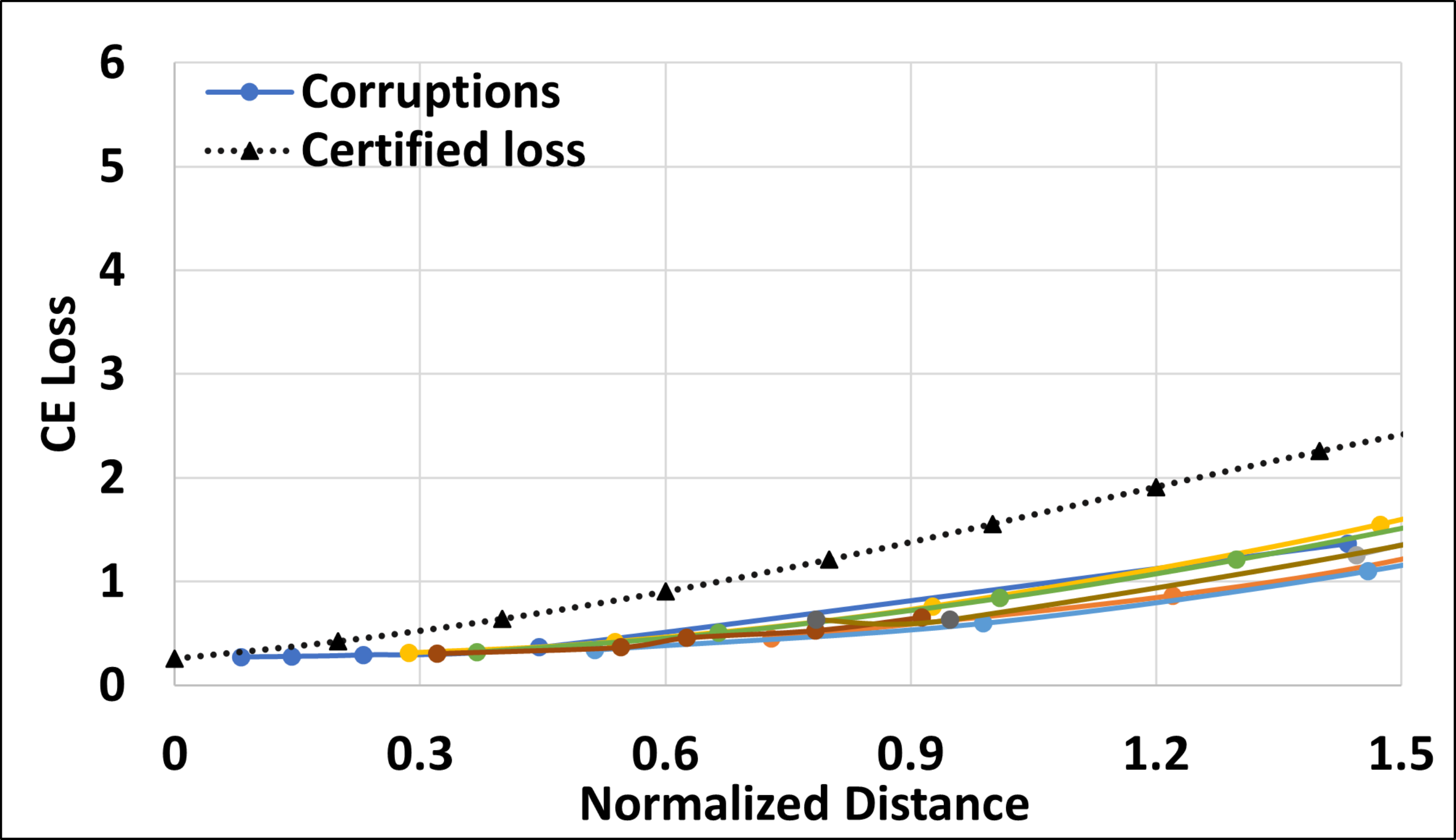}}
  \subfigure[DR-DG ($F=0.3$) on PACS]{\includegraphics[width=0.32\columnwidth]{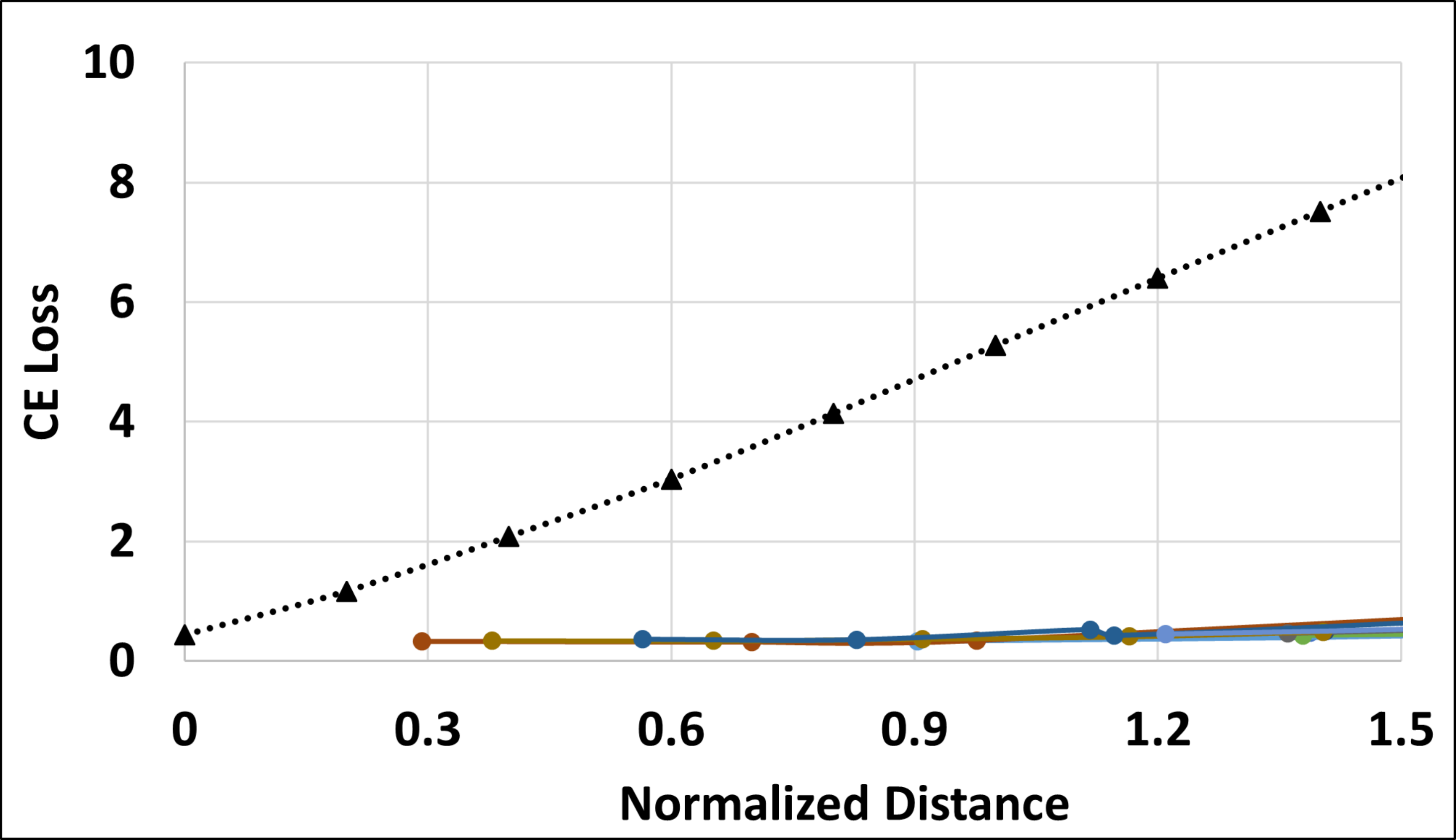}}
  \subfigure[DR-DG ($F=0.2$) on VLCS]{\includegraphics[width=0.32\columnwidth]{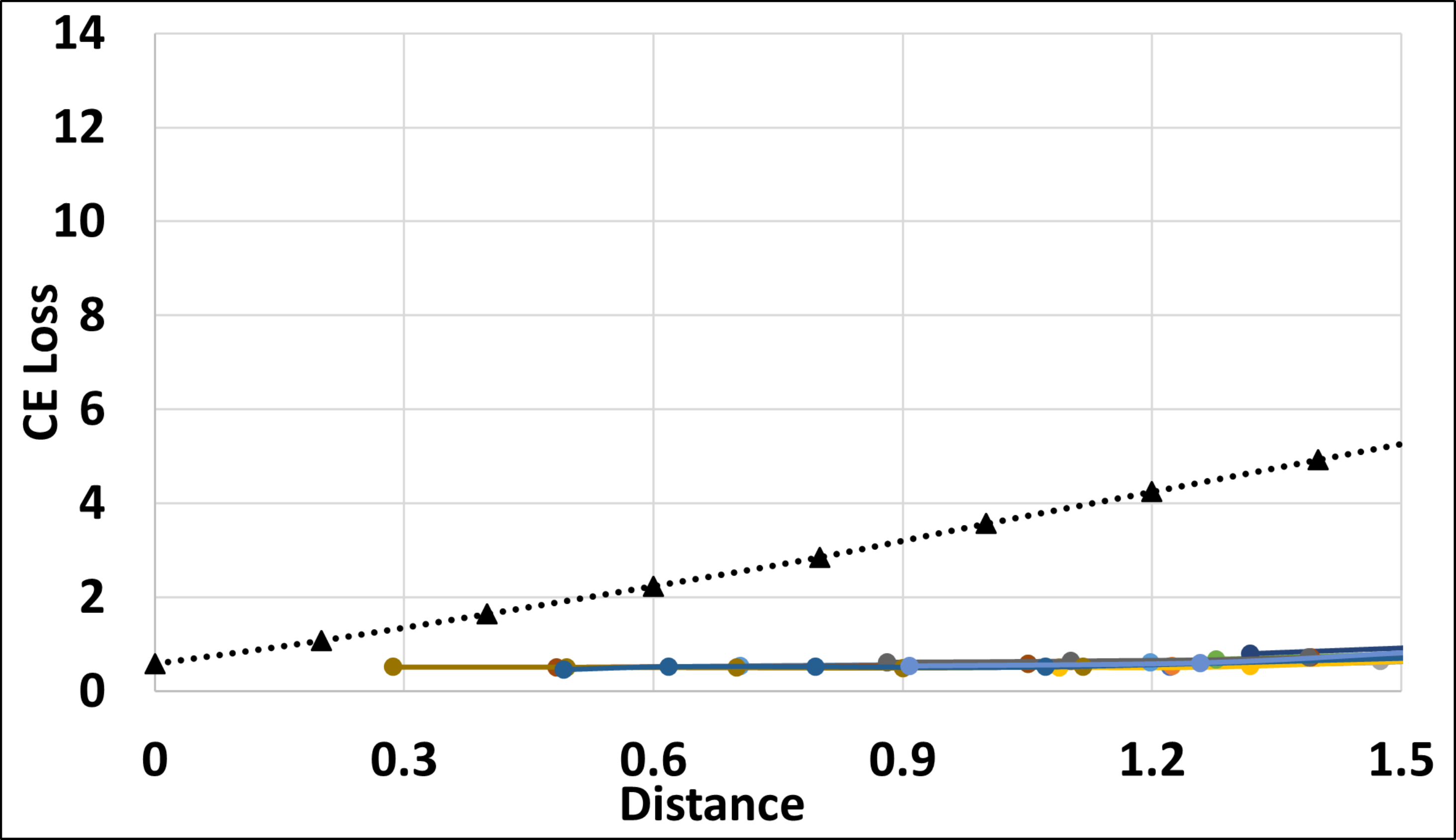}}
  
  \caption{(Best viewed in color.) Comparison of the certified (worst-case) loss of the models trained with CDAN and VREX methods on R-MNIST, PACS, VLCS. Rows 1 and 3 show models trained with Vanilla DG methods and rows 2 and 4 show models trained with DR-DG using additional losses from CDAN and VREX. The models trained with DR-DG incur smaller certified loss compared to their vanilla counterparts and only slightly higher loss on unseen distributions created through common corruptions.}
  \label{fig:loss_before_after_cdan_vrex}
\end{figure}

Here we present detailed results (Rows 1 and 3 of Figs.~\ref{fig:loss_before_after_wm_g2dm} and~\ref{fig:loss_before_after_cdan_vrex}) of using our certification method Cert-DG on models trained using WM, G2DM \cite{albuquerque2019generalizing} and CDAN \cite{long2018conditional} on R-MNIST, PACS, and VLCS datasets as presented in Sec.~\ref{sec:cert_vanilla}. 
Similar to the results in Fig~\ref{fig:certification_vanilla_wm_rotatedmnist}, our results in the rows 1 and 3 of Figs.~\ref{fig:loss_before_after_wm_g2dm} and~\ref{fig:loss_before_after_cdan_vrex} show that models trained with vanilla DG methods incur high certified loss even very close to the source. 
The high certified loss implies that the performance of DG methods can vary substantially on different distributions at the same distance and thus good performance on certain benchmark datasets is not enough to guarantee good generalization performance of these methods on unseen domains.
Comparing different algorithms on same datasets we observe that models trained with WM achieve comparatively smaller worst-case loss than the models trained with G2DM, CDAN and VREX.
This is due to the fact that WM, which explicitly minimizes the Wasserstein distance between the distributions in the representation space, is closely related to the certification procedure which also uses the Wasserstein distance. 
On the other hand, G2DM and CDAN rely on discriminators to align the two domains which minimize the divergence between distributions but as seen in our results their performance is less certifiable. Moreover, VREX does not explicitly consider reducing the distance of the sources in the representation space and focuses on making the hypothesis yield similar risk/loss on the source domains. Although the objective of VREX is different from those of the other algorithms considered here but we include it in our evaluation to demonstrate that our certification procedure can be used with any DG method. 
A detailed empirical/theoretical comparison of which DG algorithm works the best across the board still remains open although we can now use our target independent certification procedure to gain better insights into this problem rather than only relying on the performance of these methods on a small number of unseen domains available in benchmark datasets.
%compae are related and which method is theoretically more suitable for achieving better-certified loss is interesting and is a part of future work.

Lastly, in addition to the certified loss, we also plot loss on different corruption of the data in Figs.~\ref{fig:loss_before_after_wm_g2dm} and~\ref{fig:loss_before_after_cdan_vrex}. Similar to Fig~\ref{fig:certification_vanilla_wm_rotatedmnist}, we observe that certified loss is much higher than the loss on the corrupted versions of the source test sets which indicates the existence of other unseen distributions that can lower the performance of the DG models. % beyond those generated from common corruptions. 

\begin{figure}[tb]
  \centering
  \subfigure[WM]{\includegraphics[width=0.24\columnwidth]{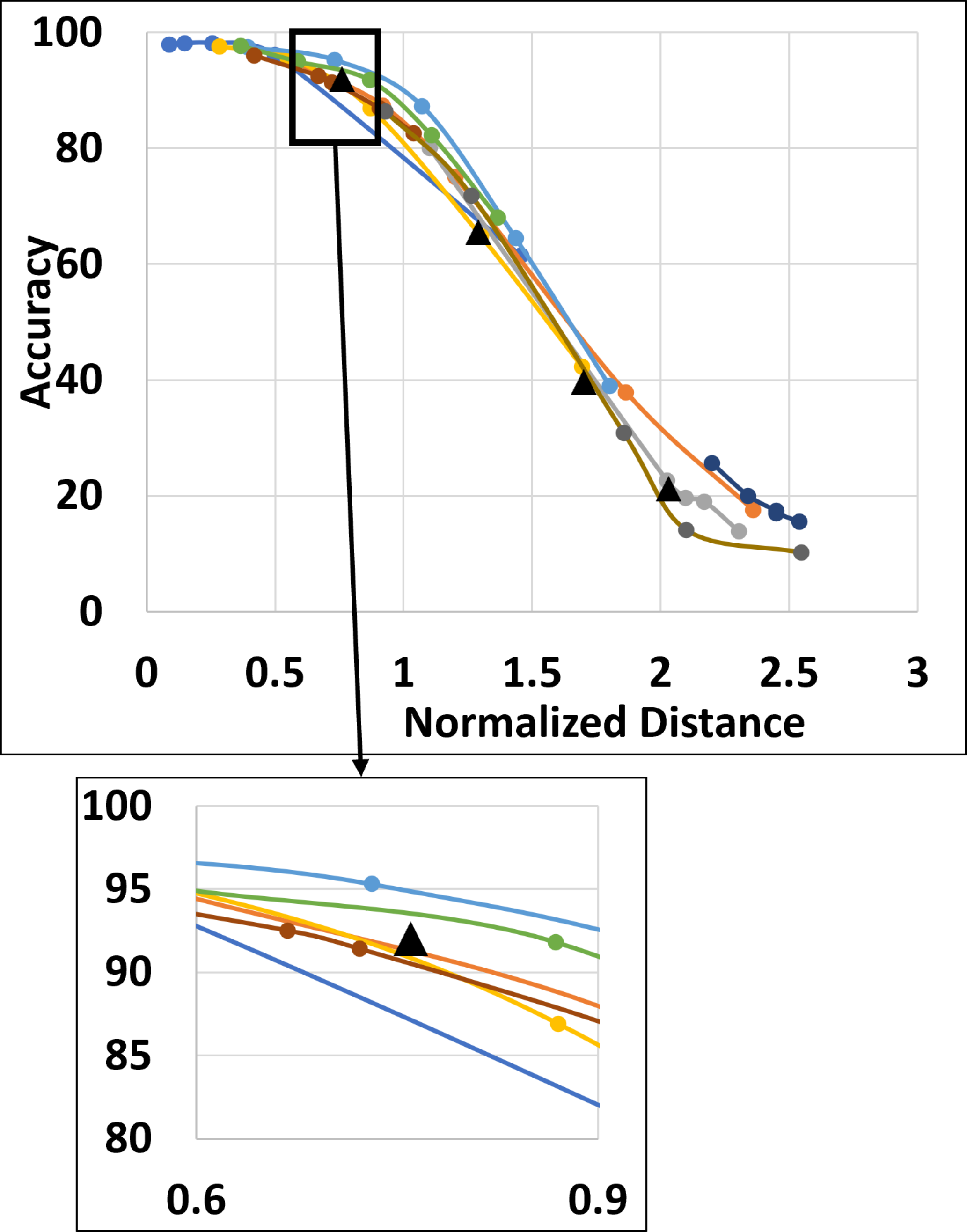}}
  %\hfill
  \subfigure[G2DM]{\includegraphics[width=0.24\columnwidth]{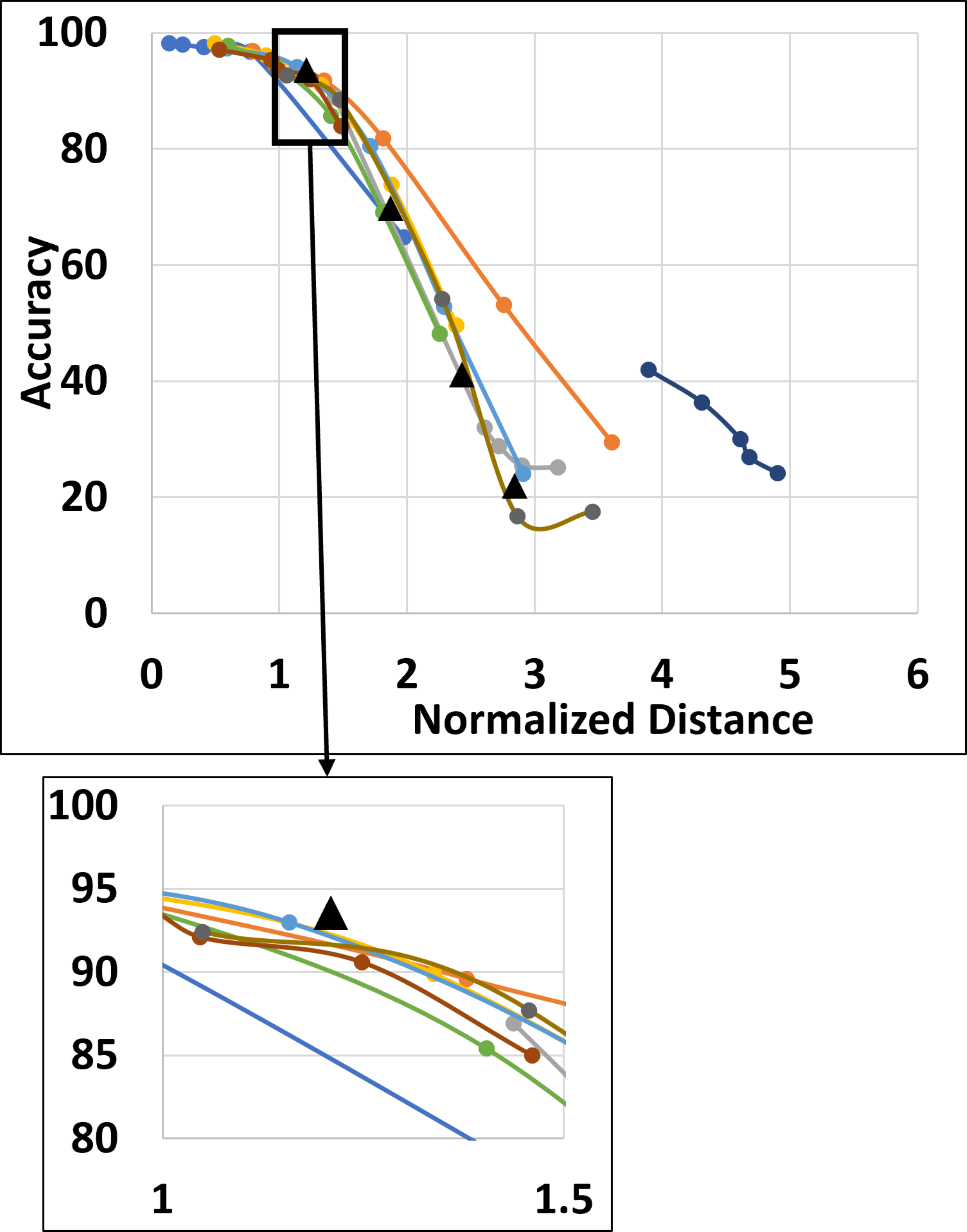}}
  \subfigure[CDAN]{\includegraphics[width=0.24\columnwidth]{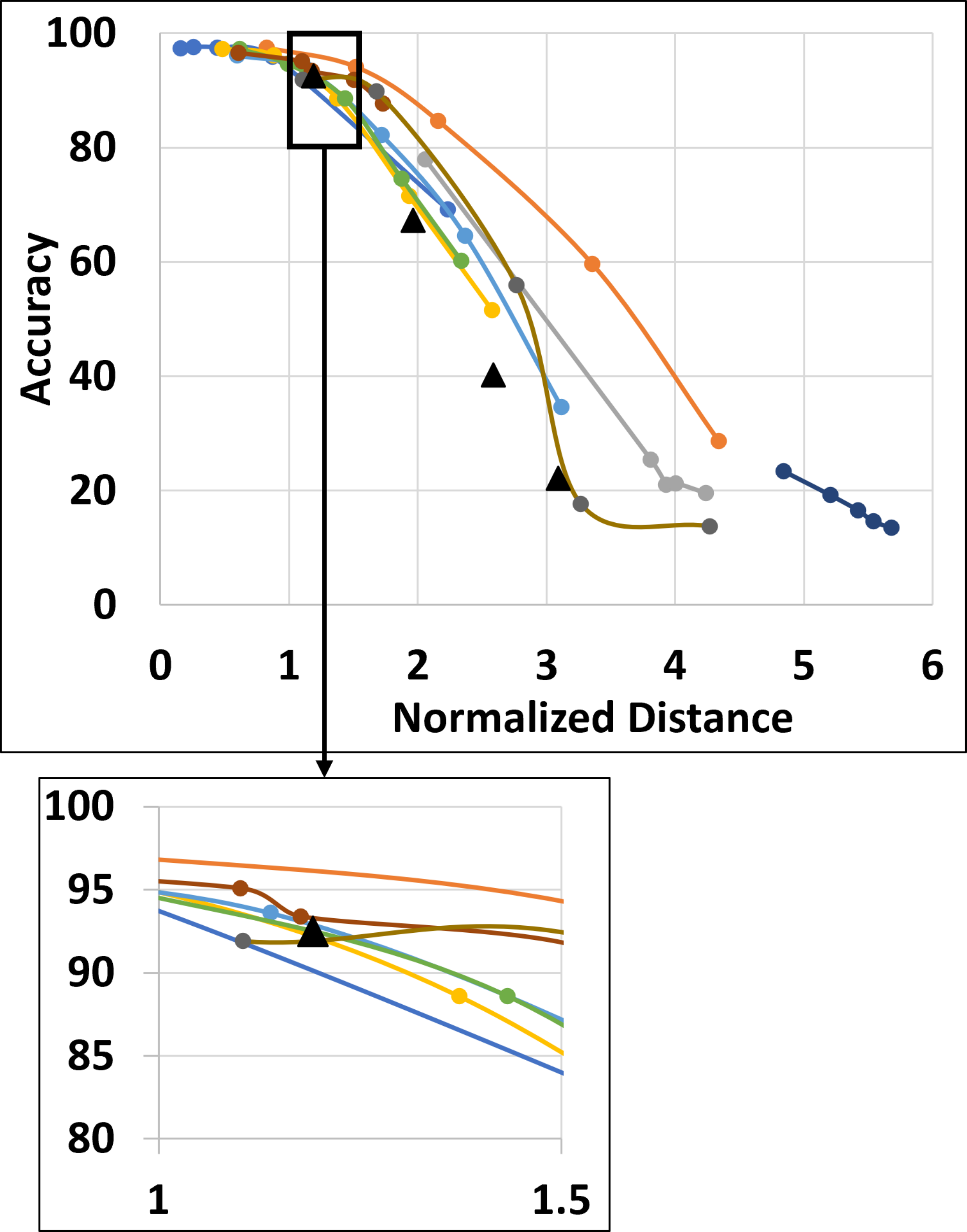}}
  \subfigure[VREX]{\includegraphics[width=0.24\columnwidth]{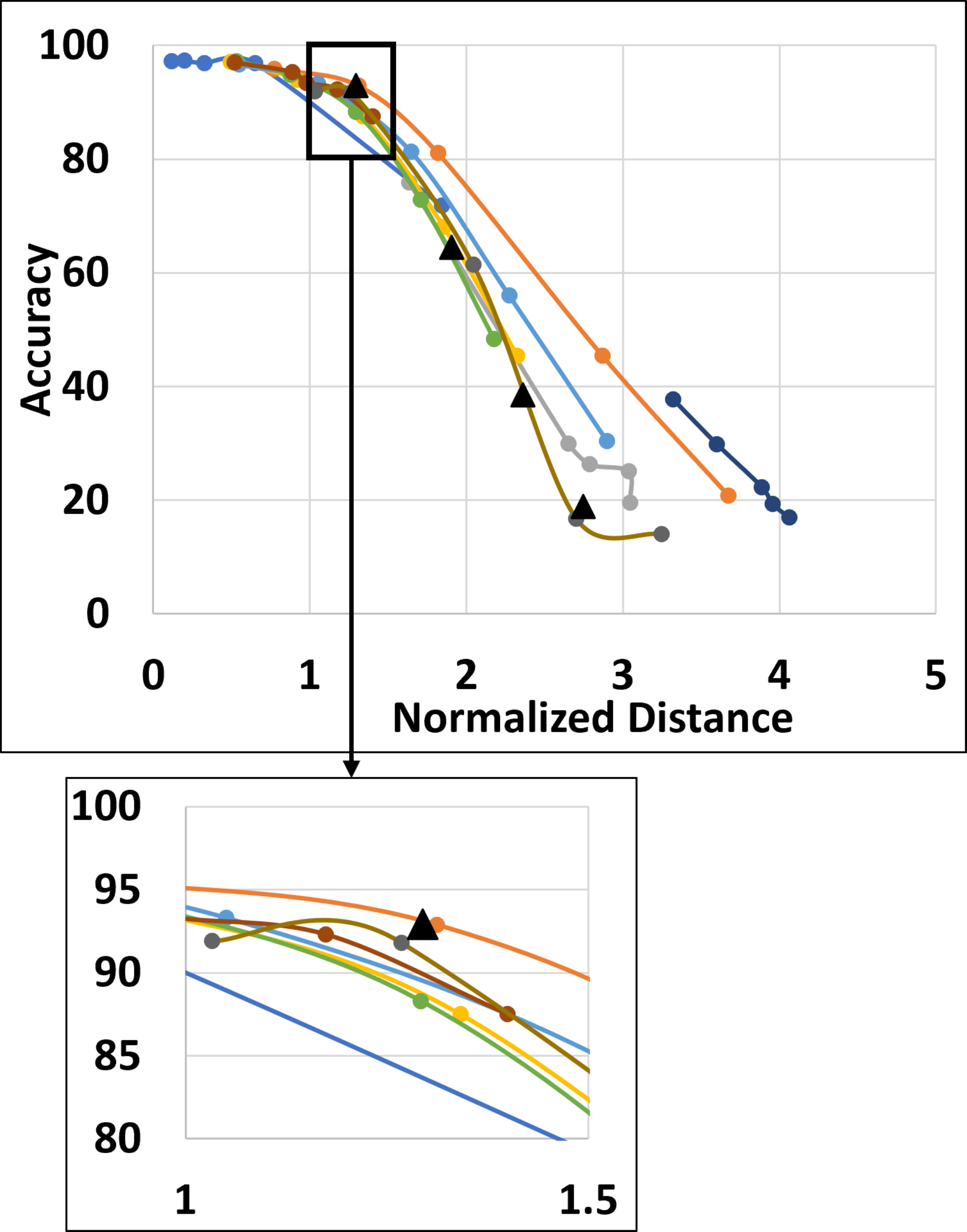}}
  \includegraphics[width=0.75\textwidth]{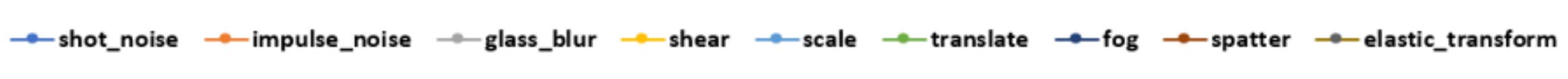}
  \caption{(Best viewed in color.)
    {\bf Difficulty of empirically evaluating the generalization performance of a DG method:} 
   The difference in the performance of DG methods on data from different unseen domains demonstrates that comparing the performance of DG methods using a few benchmark datasets is not sufficient for assessing the generalization performance of DG methods. 
   The triangles denote unseen benchmark distributions ($30^\circ, 45^\circ, 60^\circ, 75^\circ$ for R-MNIST (left to right)) and lines denote distributions under common corruptions.
   } 
  \label{fig:high_variability_of_dg_b}
\end{figure}

\subsection{Certification results of models trained with DR-DG}
Here we present the full certification results (Rows 2 and 4 in Figs.~\ref{fig:loss_before_after_wm_g2dm} and~\ref{fig:loss_before_after_cdan_vrex}) of models trained with DR-DG. In addition to minimizing the loss on the worst-case distribution in the representation space, we use losses of four DG methods for DR-DG training. Specifically, we use WM, G2DM \cite{albuquerque2019generalizing}, CDAN \cite{long2018conditional}, and VREX \cite{krueger2021out} with DR-DG.
Similar to the results presented in the main paper in Fig.~\ref{fig:certification_before_after_dro}, we observe that the DR-DG significantly improves the worst-case loss of the models compared to training with vanilla DG methods. 
Specifically, the dotted lines, depicting the certified loss, in the rows 2 and 4 are much lower than the dotted lines in the rows 1 and 3 of Figs.~\ref{fig:loss_before_after_wm_g2dm} and~\ref{fig:loss_before_after_cdan_vrex}.
This improvement in the worst-case loss suggests that the models trained with DR-DG are more robust to unseen domains than their vanilla counterparts.
%\AM{Add discussion on why different DG algorithms perform similarly} 
We also observe that the gap between the certified and the empirical loss of the models is much smaller for the models trained with WM in comparison to models trained with G2DM, CDAN and VREX. As mentioned in the previous section, this difference is due to the similarity certification and WM both relying on Wasserstein distance.
In comparison to G2DM and CDAN, we observe marginally better results with using a conditional discriminator as suggested by CDAN although we found the CDAN algorithm to be more stable when used in DR-DG. 

Another important observation that makes DR-DG models better than their vanilla counterparts is that after DR-DG training the corrupted distributions lie closer to the source as evident from the higher density of points in regions close to the source data. 
This suggests that the additional objective of minimizing the worst-case loss helps to align the distributions closer to the source in the representation space. 
Thus, helping DG methods reduce the divergence between the source and unseen distributions.
Lastly, we also note that compared to vanilla counterparts, the performance of DR-DG trained models is slightly worse on the source and distribution of corrupted data. 
As noted in Sec.~\ref{sec:eval_on_benchmark_data}, this is similar to the accuracy versus robustness trade-off observed in area of adversarial robustness. 
Efficient methodologies exist for solving this problem specifically by using different loss functions and gradually increasing $\rho$ for DR-DG training which will be explored in future work.

\subsection{Comparison of accuracy of models trained with and without DR-DG}
\begin{figure}[tb]
  \centering
  \subfigure[WM on R-MNIST]{\includegraphics[width=0.31\columnwidth]{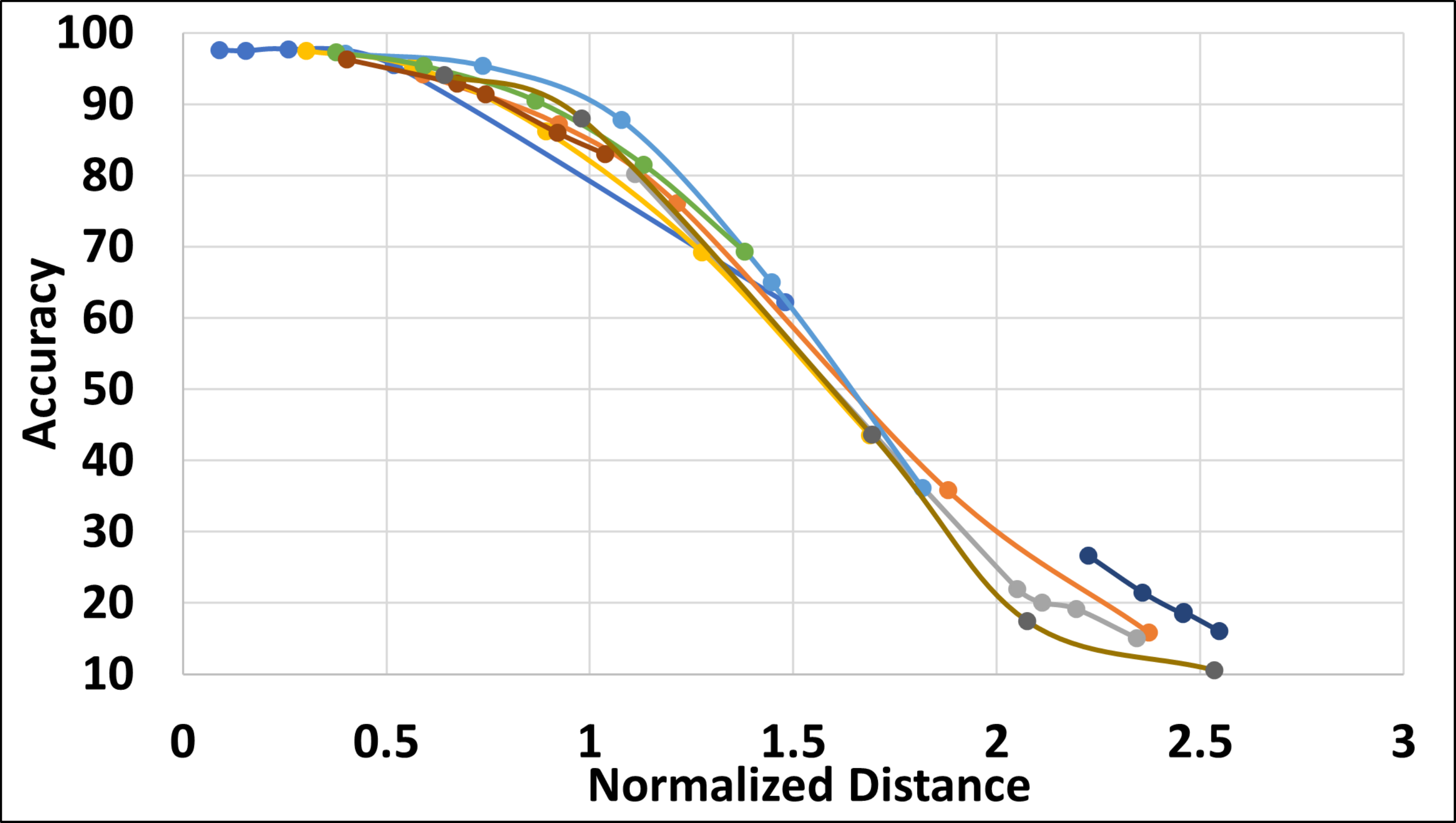}}
   \subfigure[G2DM on R-MNIST]{\includegraphics[width=0.31\columnwidth]{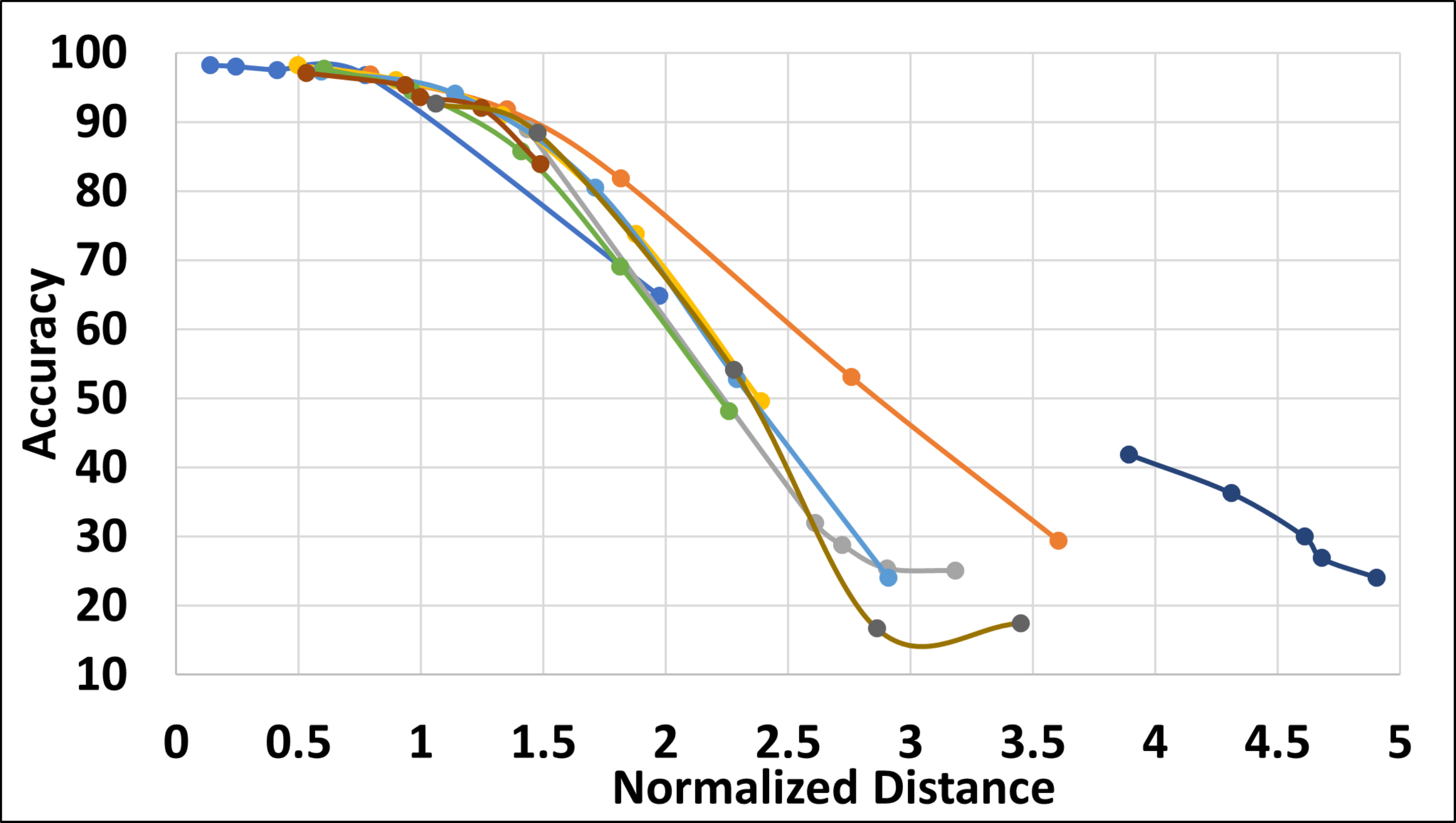}}
   \subfigure[CDAN on R-MNIST]{\includegraphics[width=0.31\columnwidth]{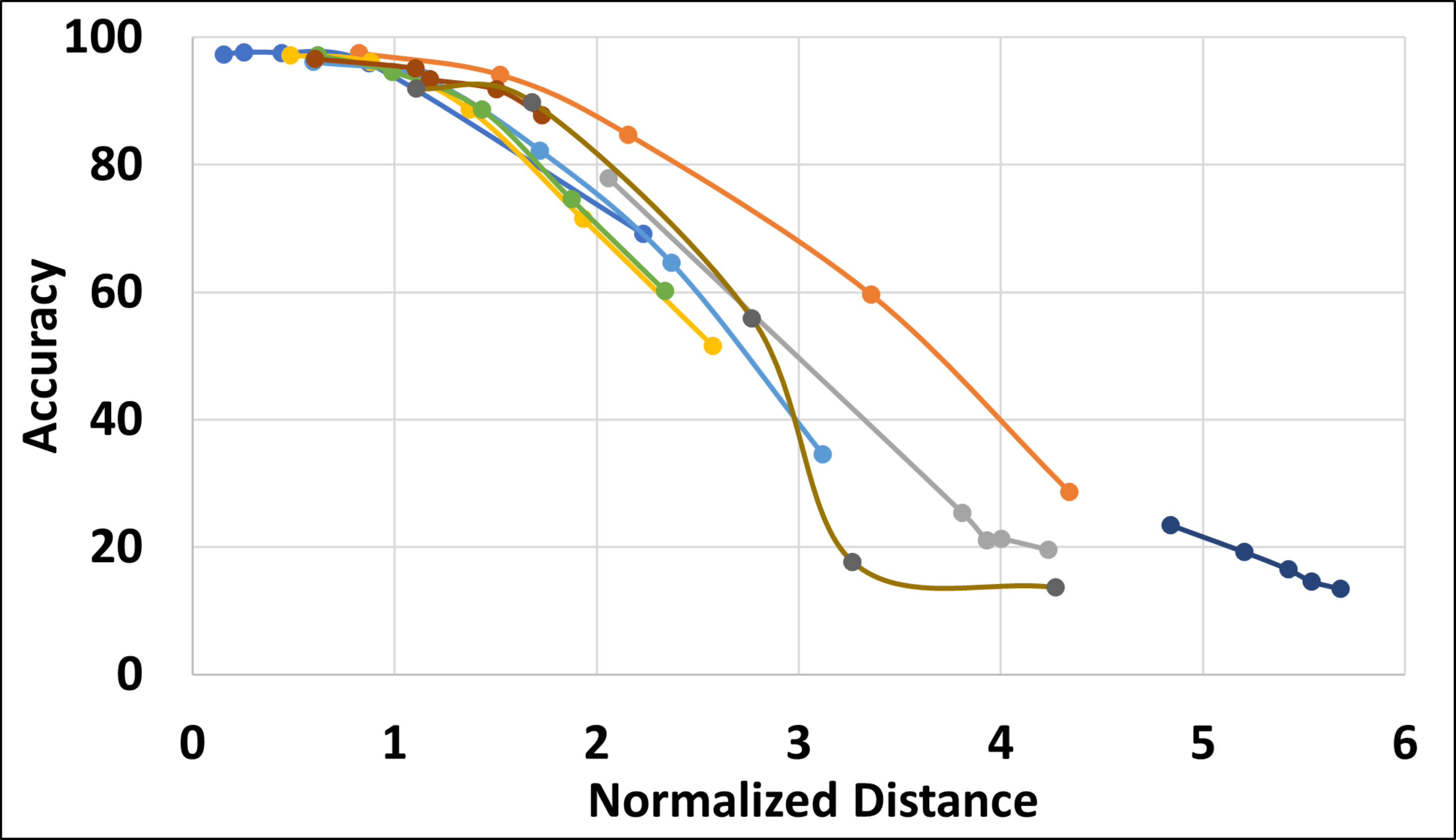}}
  \subfigure[DR-DG ($F$ = 0.5) with WM on R-MNIST]{\includegraphics[width=0.31\columnwidth]{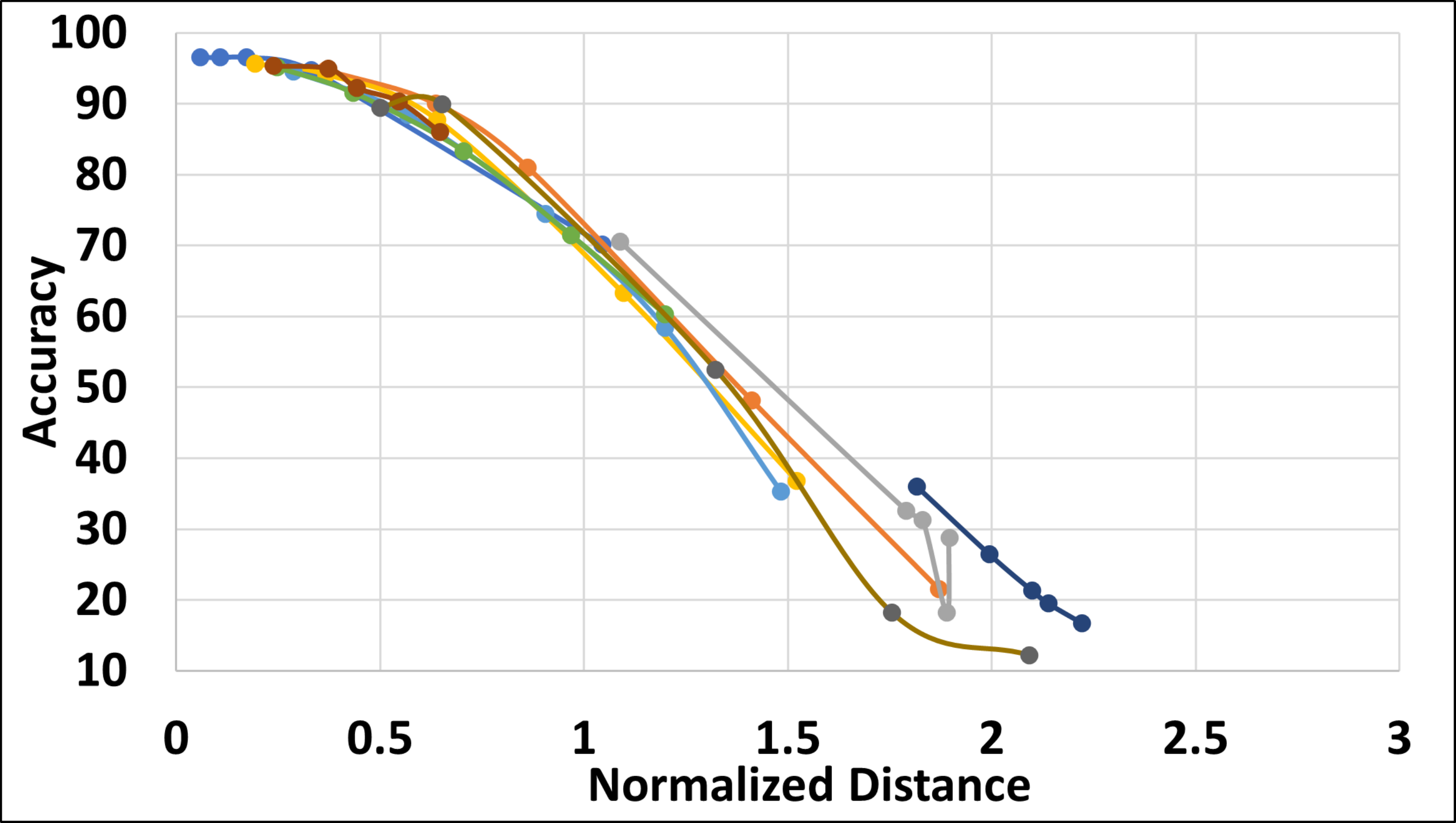}}
  \subfigure[DR-DG ($F$ = 1.5) with G2DM on R-MNIST]{\includegraphics[width=0.31\columnwidth]{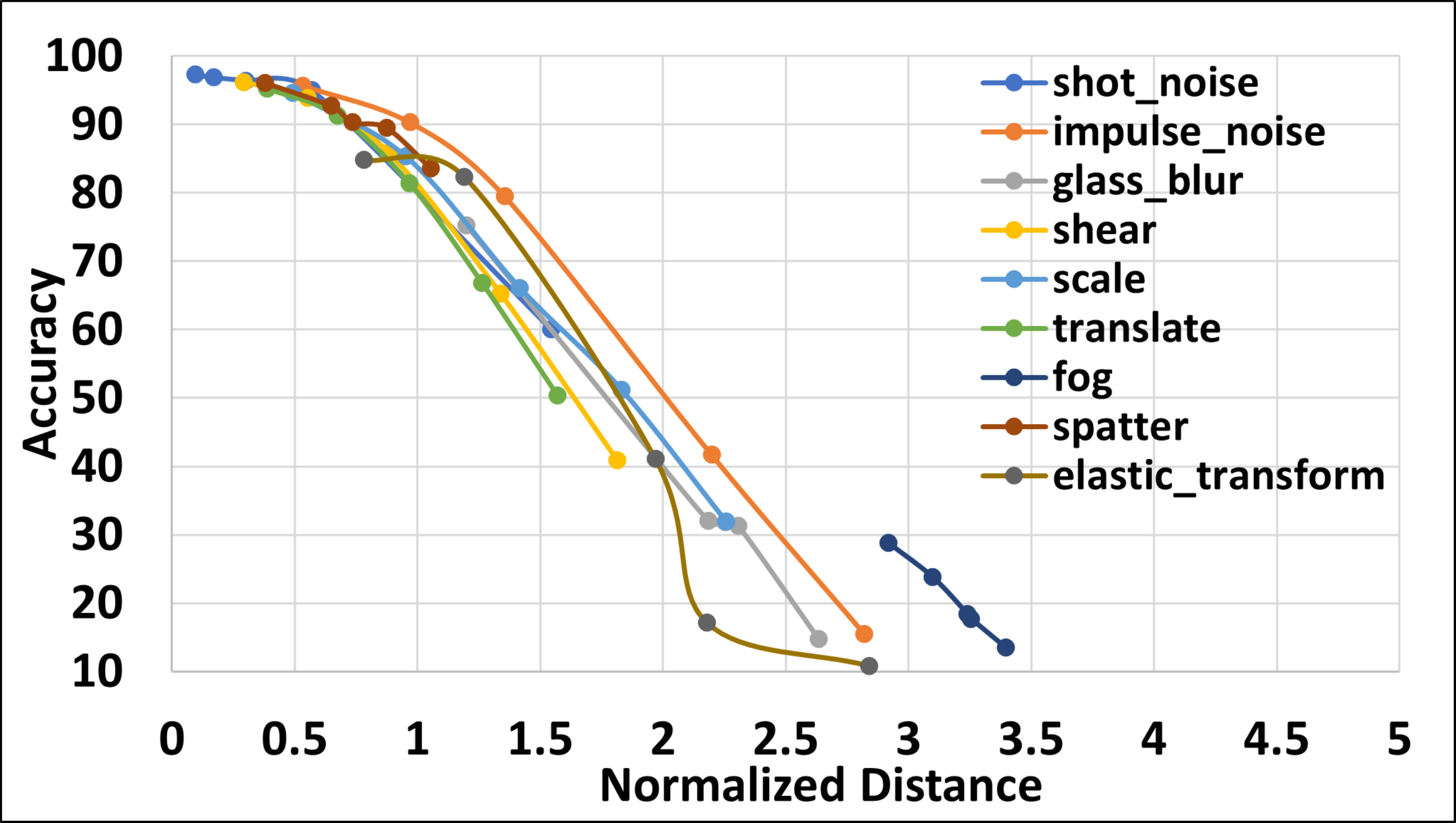}}
  \subfigure[DR-DG ($F$ = 1.5) with CDAN on R-MNIST]{\includegraphics[width=0.31\columnwidth]{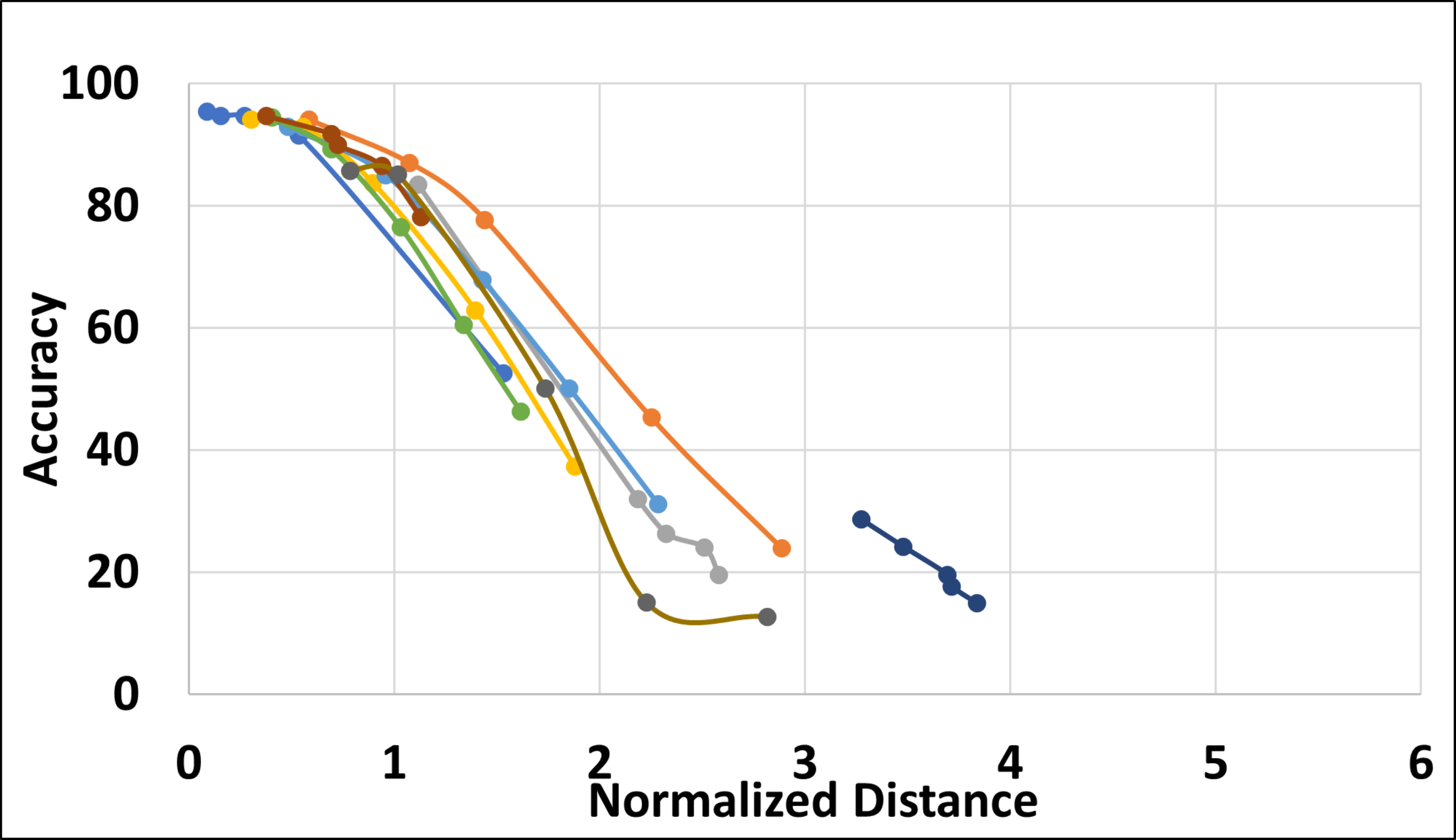}}
  
  \subfigure[WM on PACS]{\includegraphics[width=0.31\columnwidth]{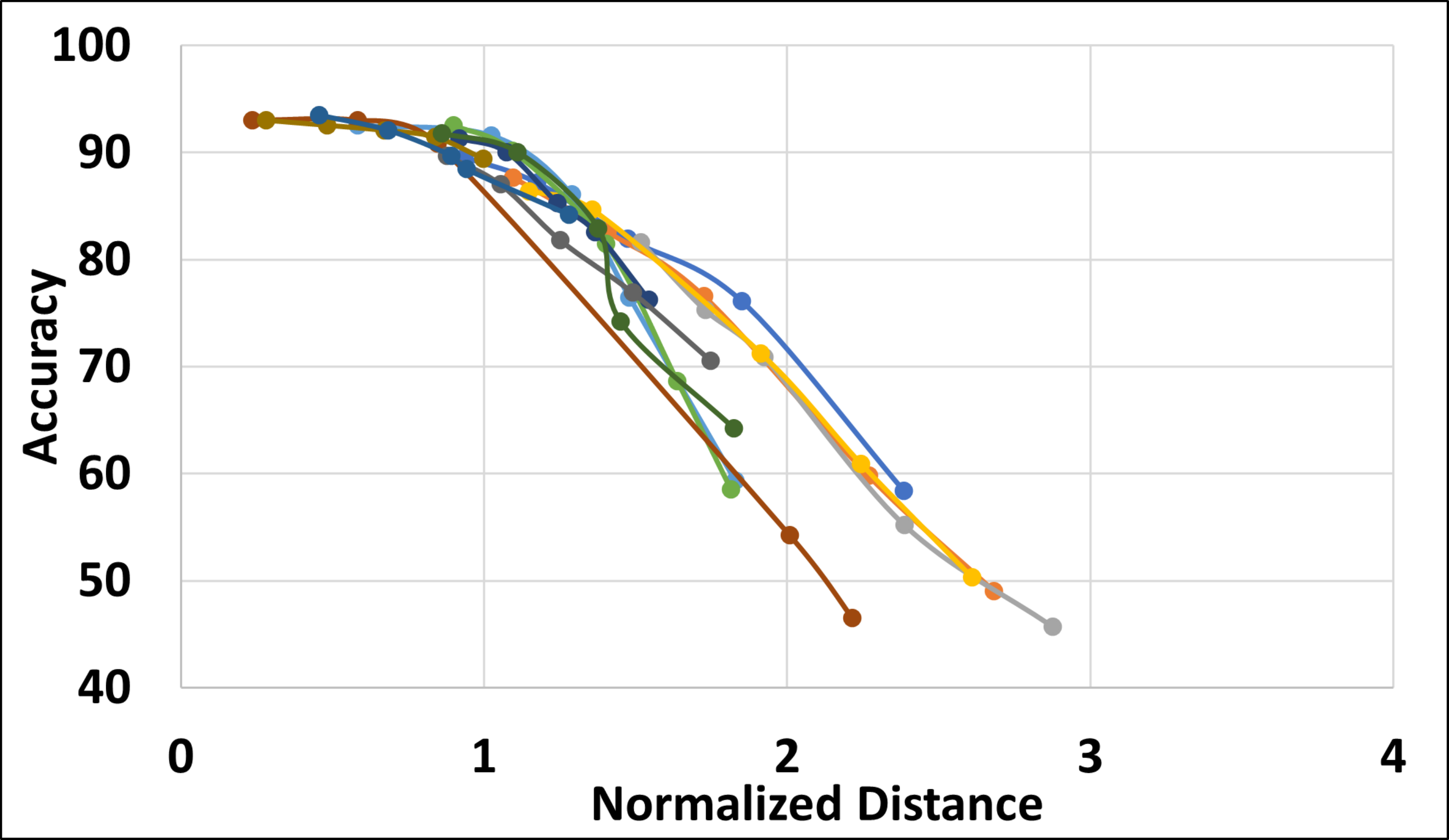}}
  \subfigure[G2DM on PACS ]{\includegraphics[width=0.31\columnwidth]{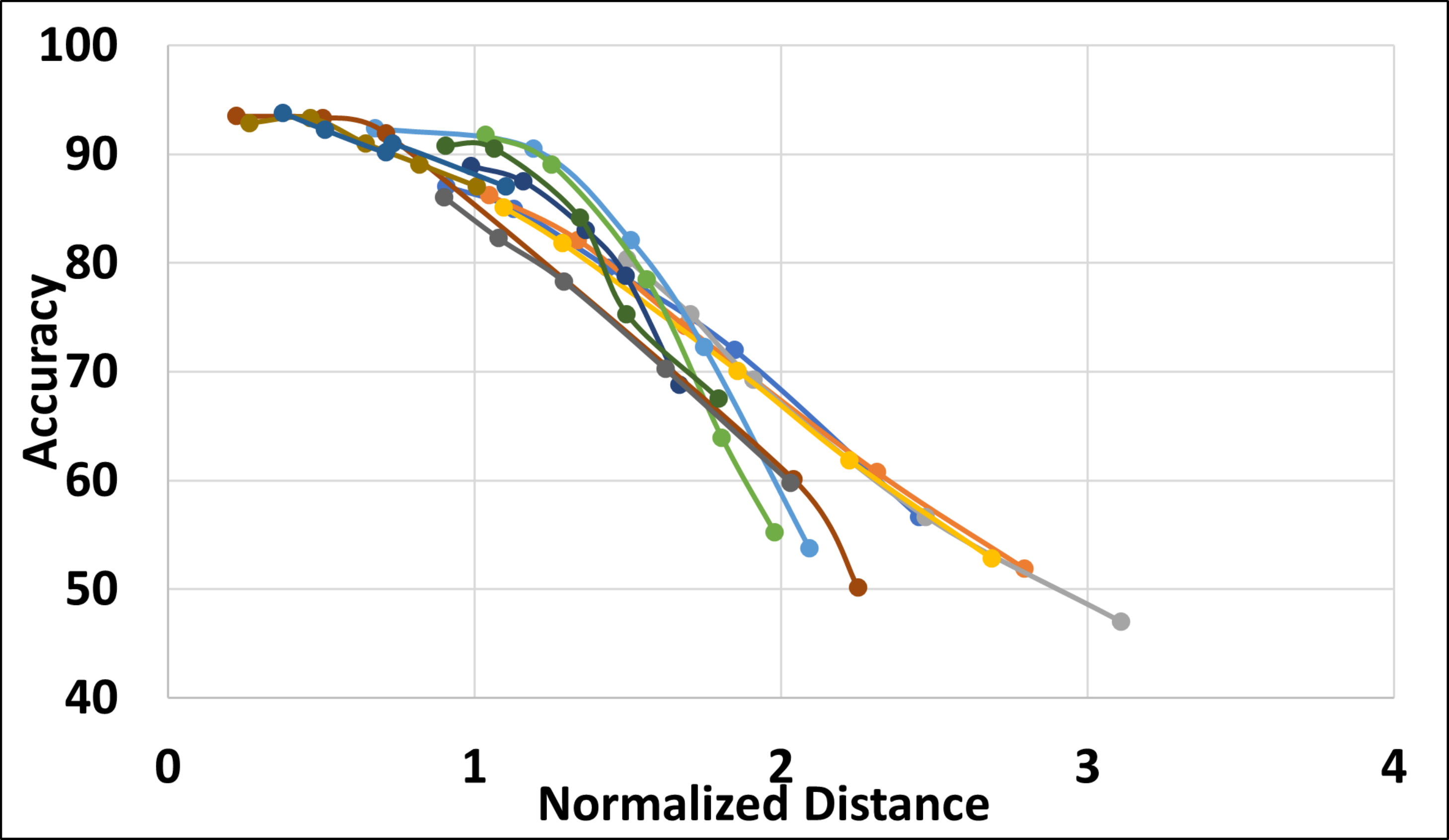}}
  \subfigure[CDAN on PACS ]{\includegraphics[width=0.31\columnwidth]{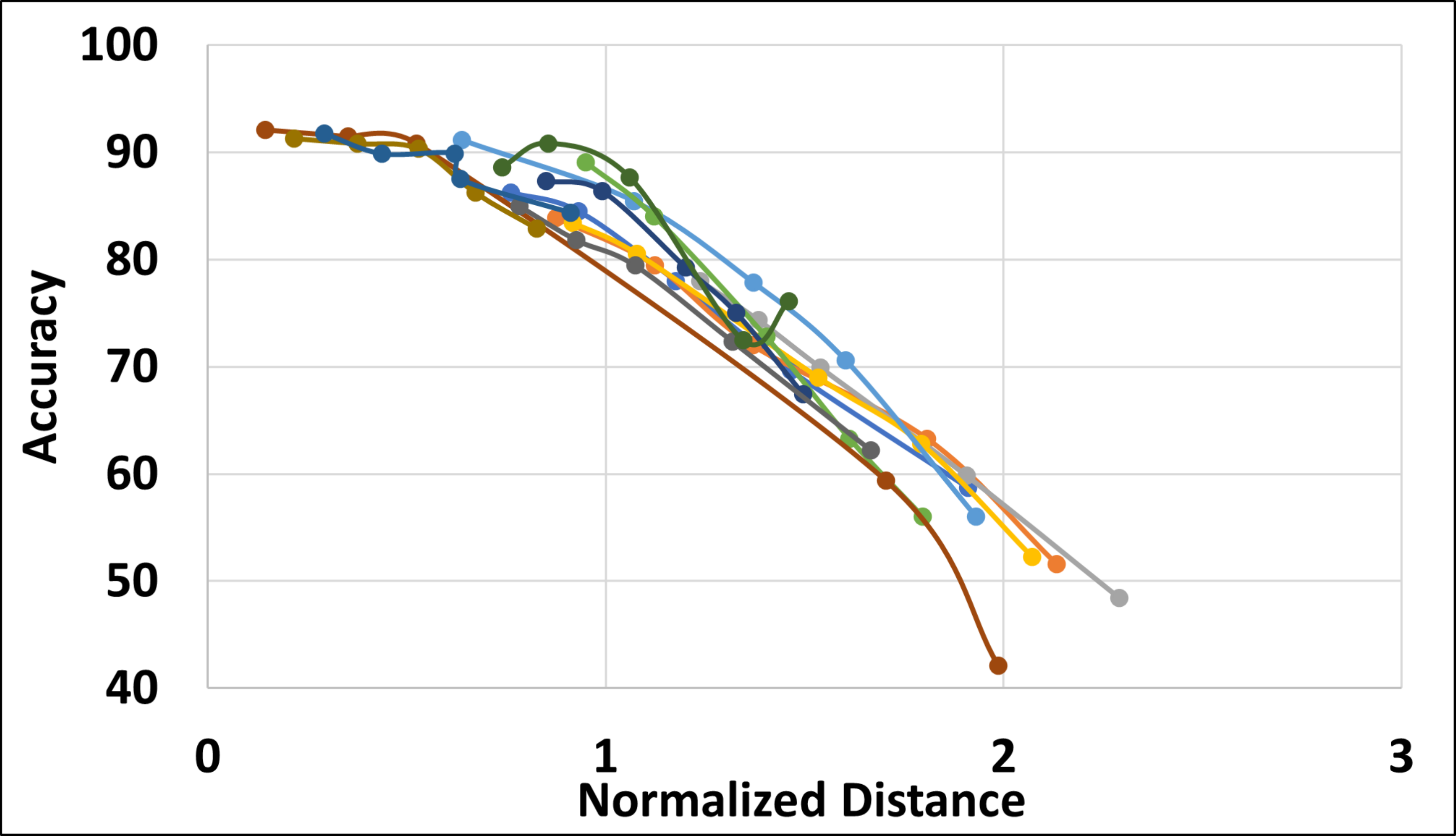}}
  
  \subfigure[DR-DG ($F=0.75$) with WM on PACS ]{\includegraphics[width=0.31\columnwidth]{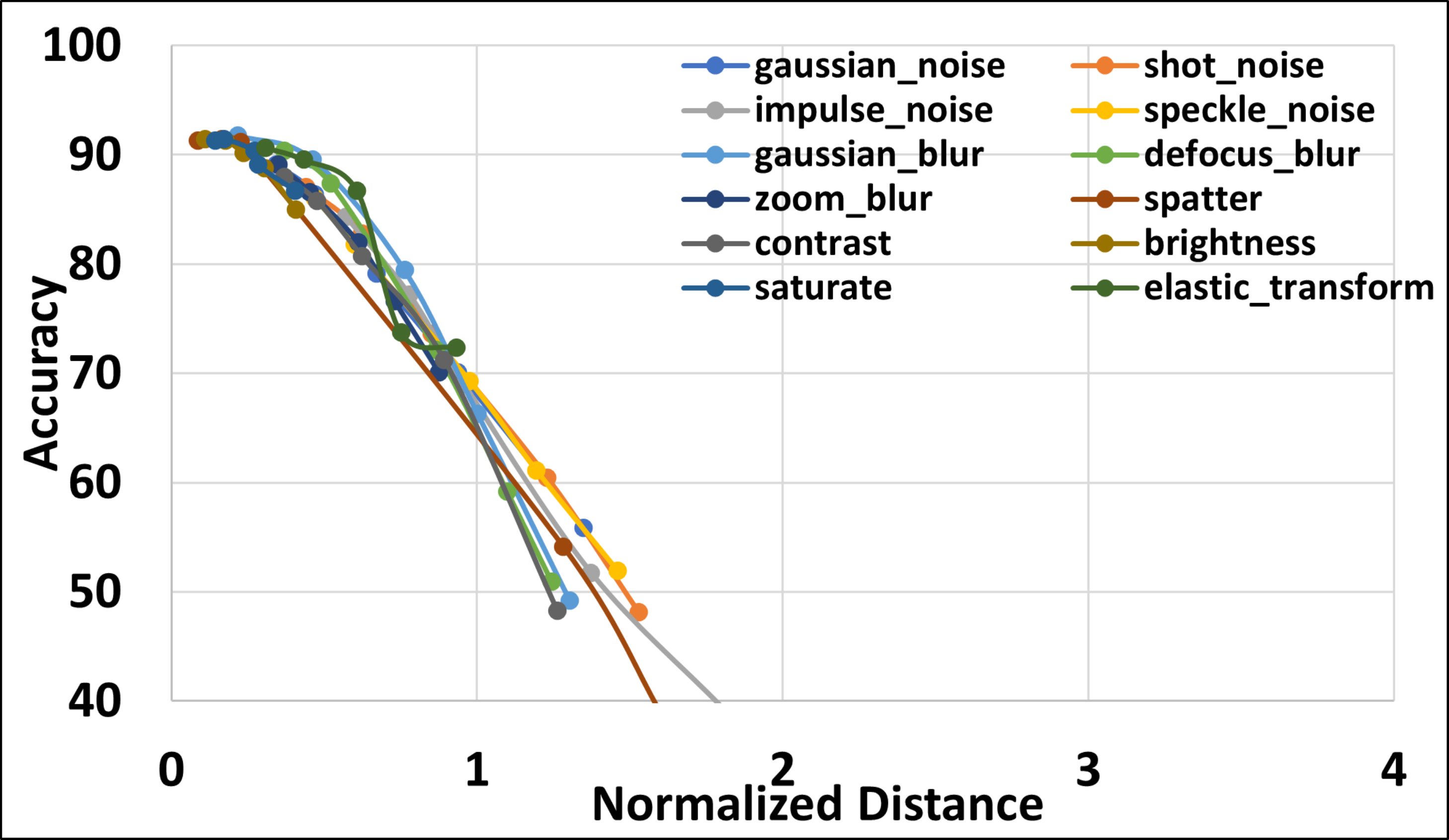}}
  \subfigure[DR-DG ($F=0.75$) with G2DM on PACS ]{\includegraphics[width=0.31\columnwidth]{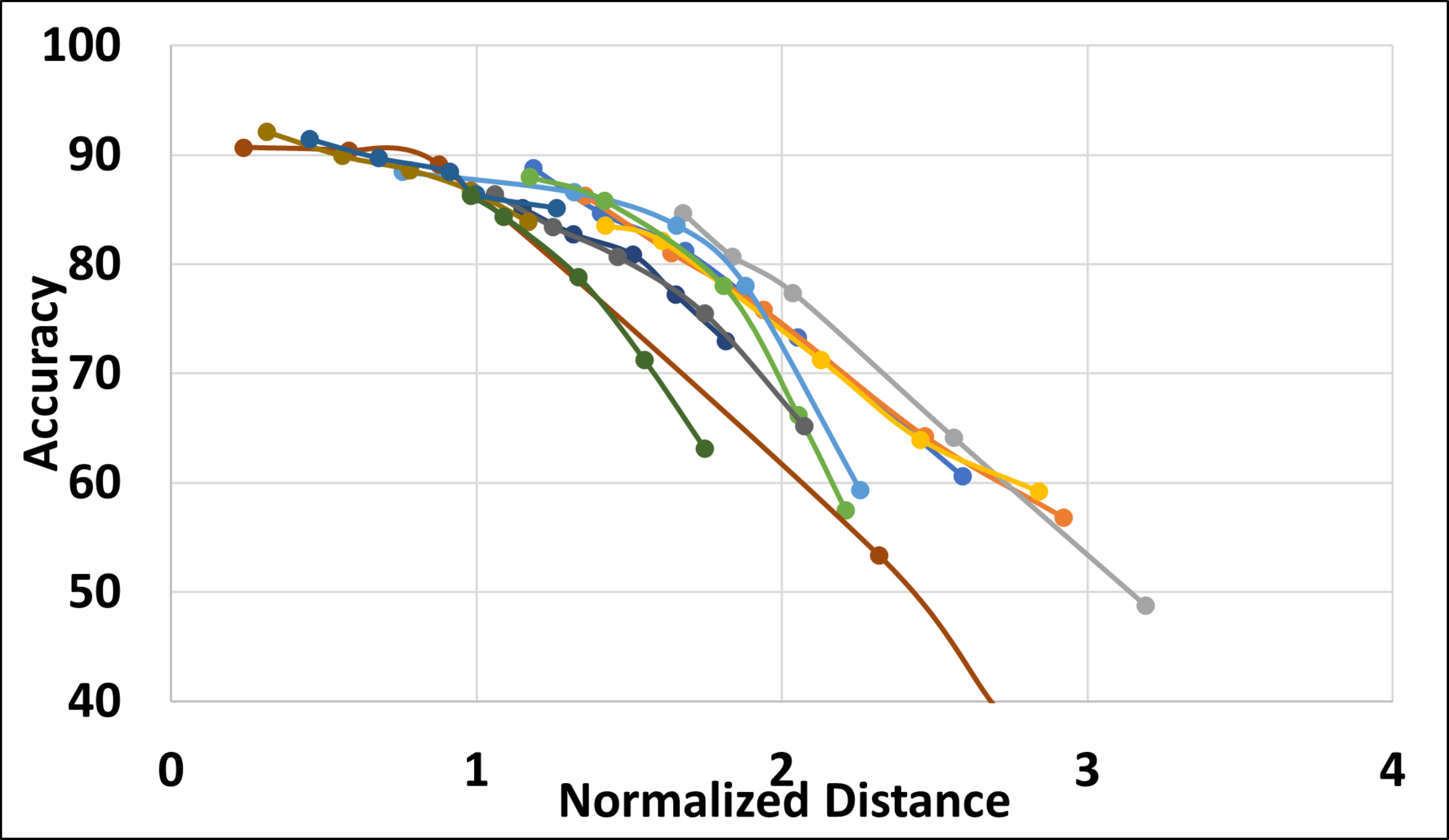}}
  \subfigure[DR-DG ($F=0.5$) with CDAN  on PACS ]{\includegraphics[width=0.31\columnwidth]{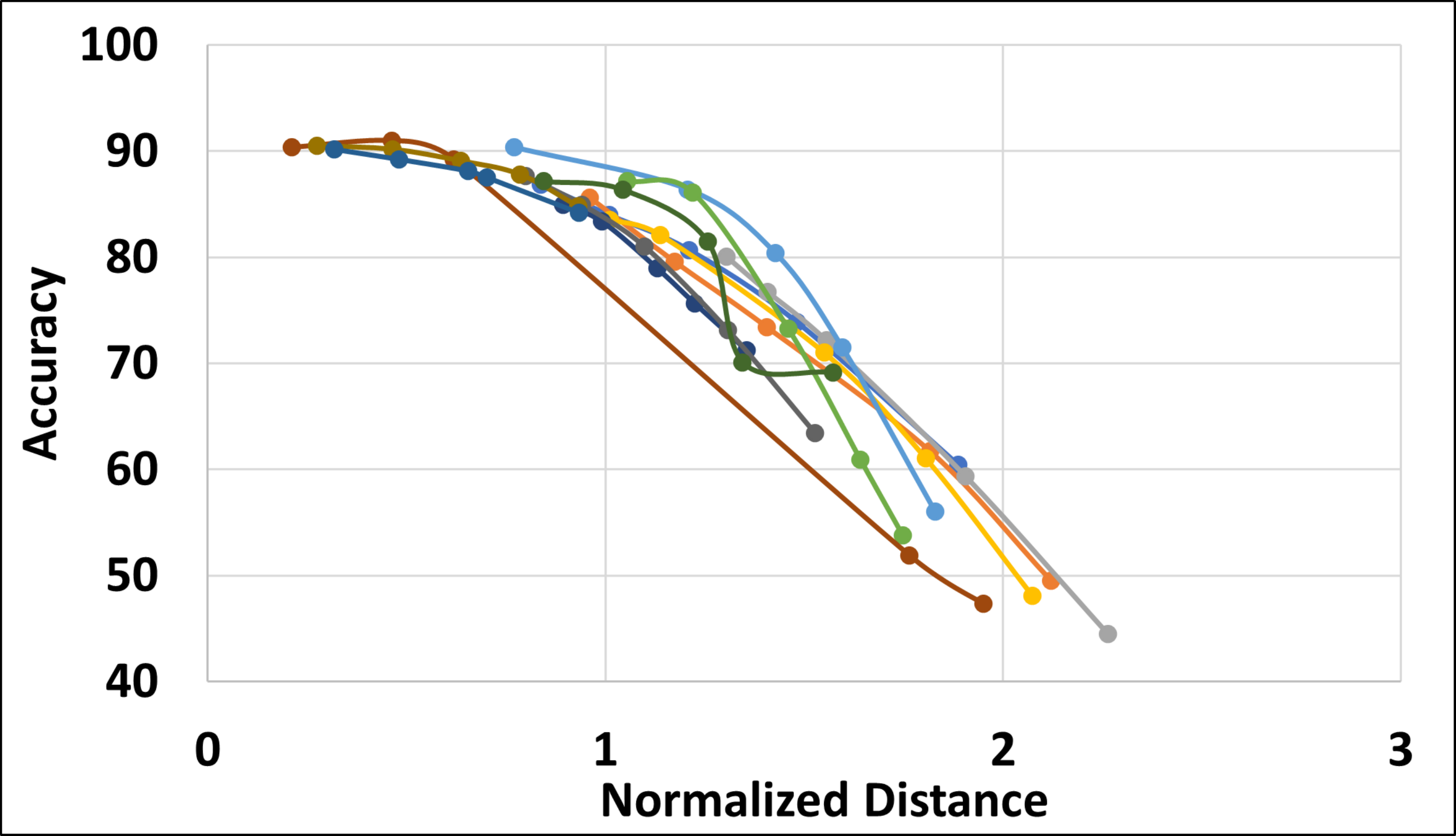}}
  
  \subfigure[WM on VLCS]{\includegraphics[width=0.31\columnwidth]{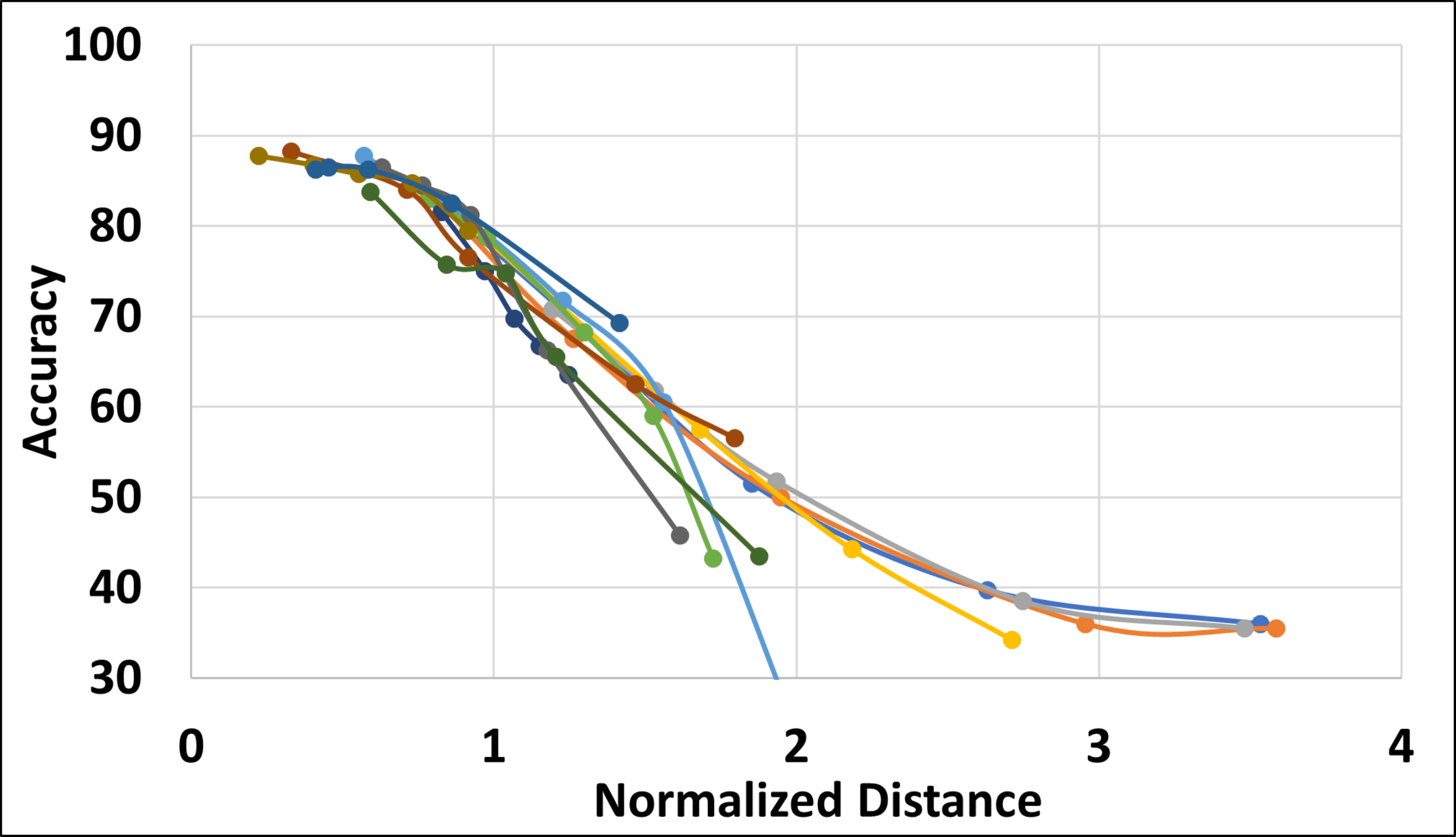}}
  \subfigure[G2DM on VLCS ]{\includegraphics[width=0.31\columnwidth]{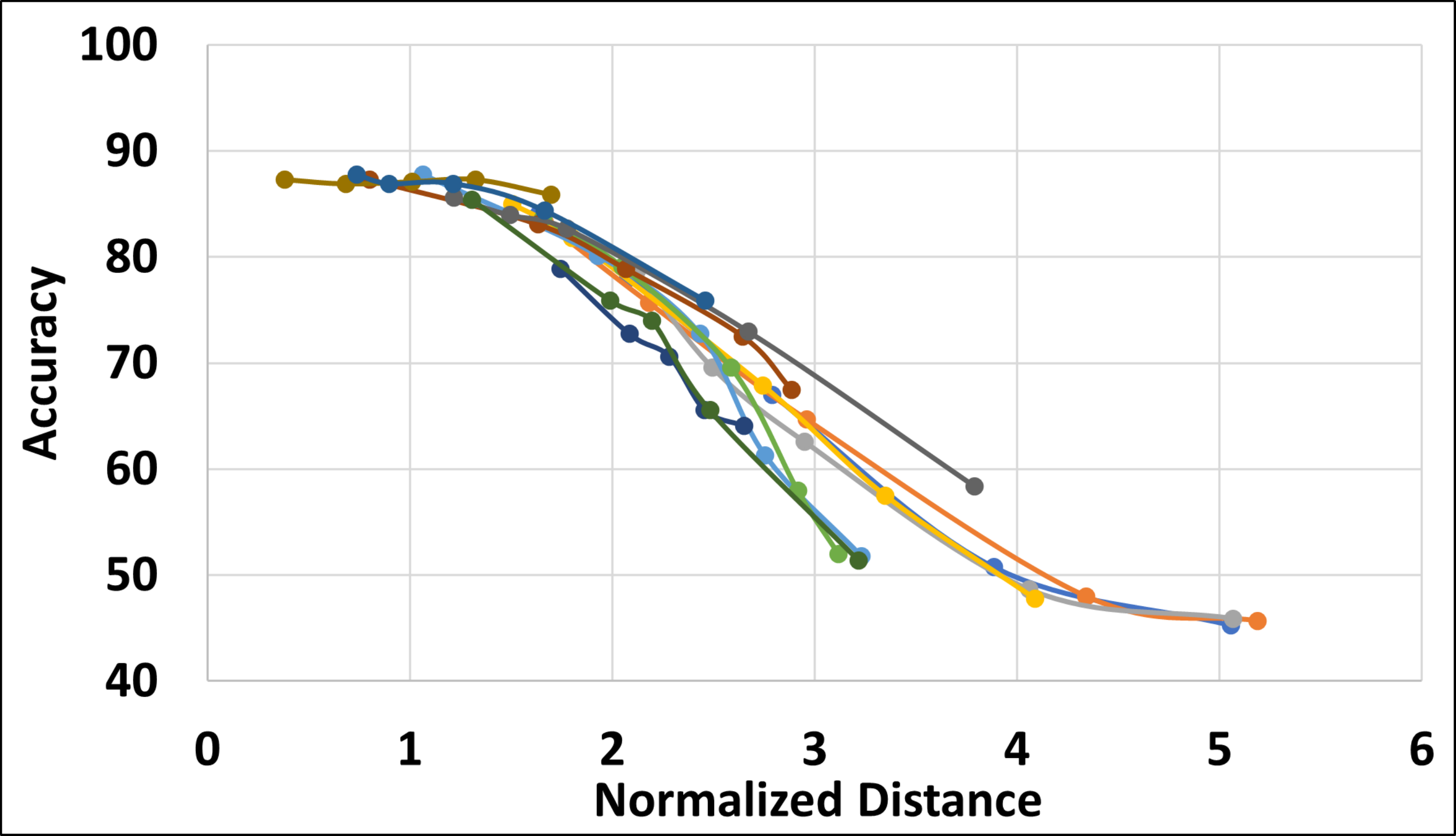}}
  \subfigure[CDAN  on VLCS]{\includegraphics[width=0.31\columnwidth]{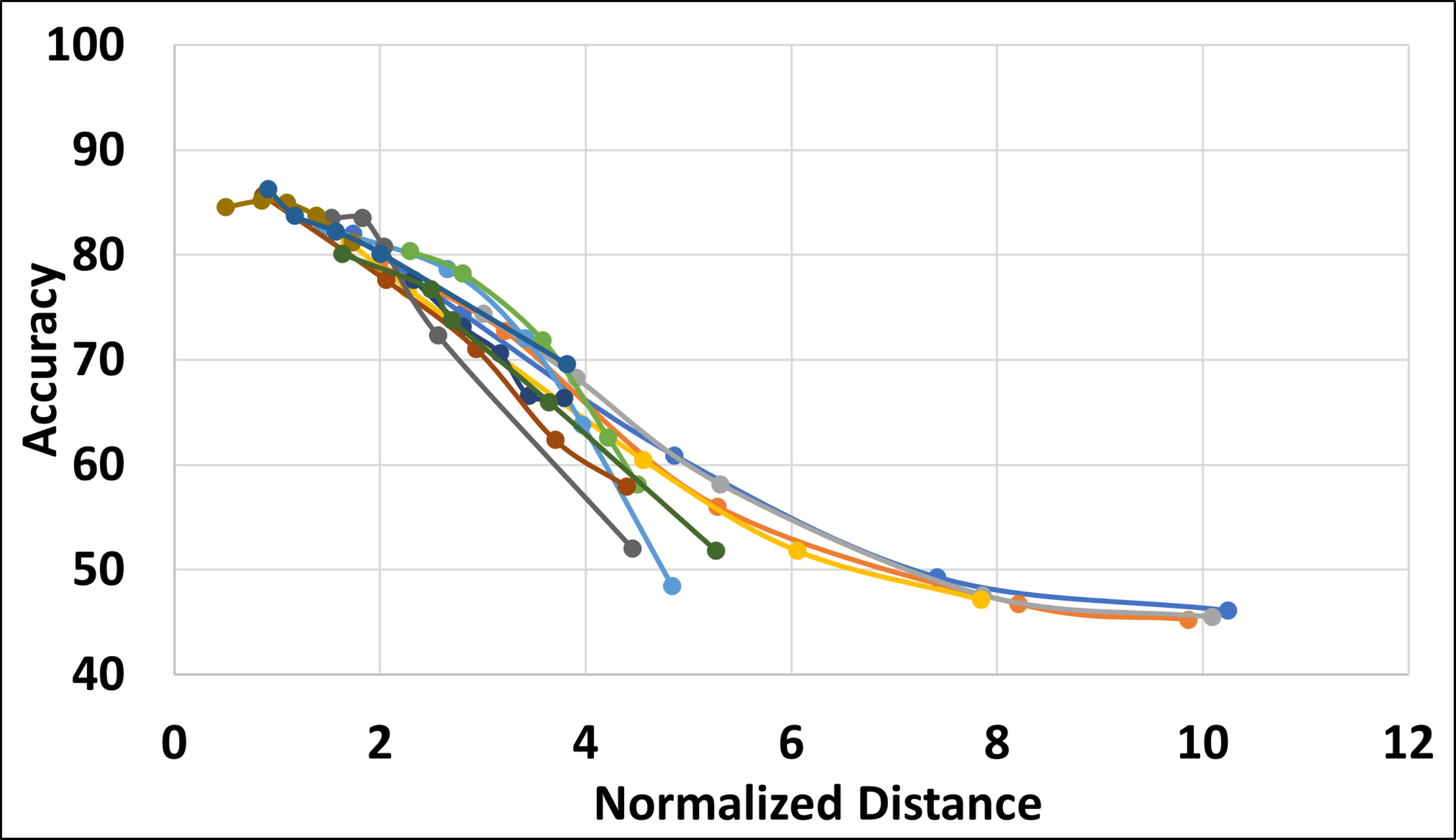}}
  
  \subfigure[DR-DG ($F=0.75$) with WM on VLCS ]{\includegraphics[width=0.31\columnwidth]{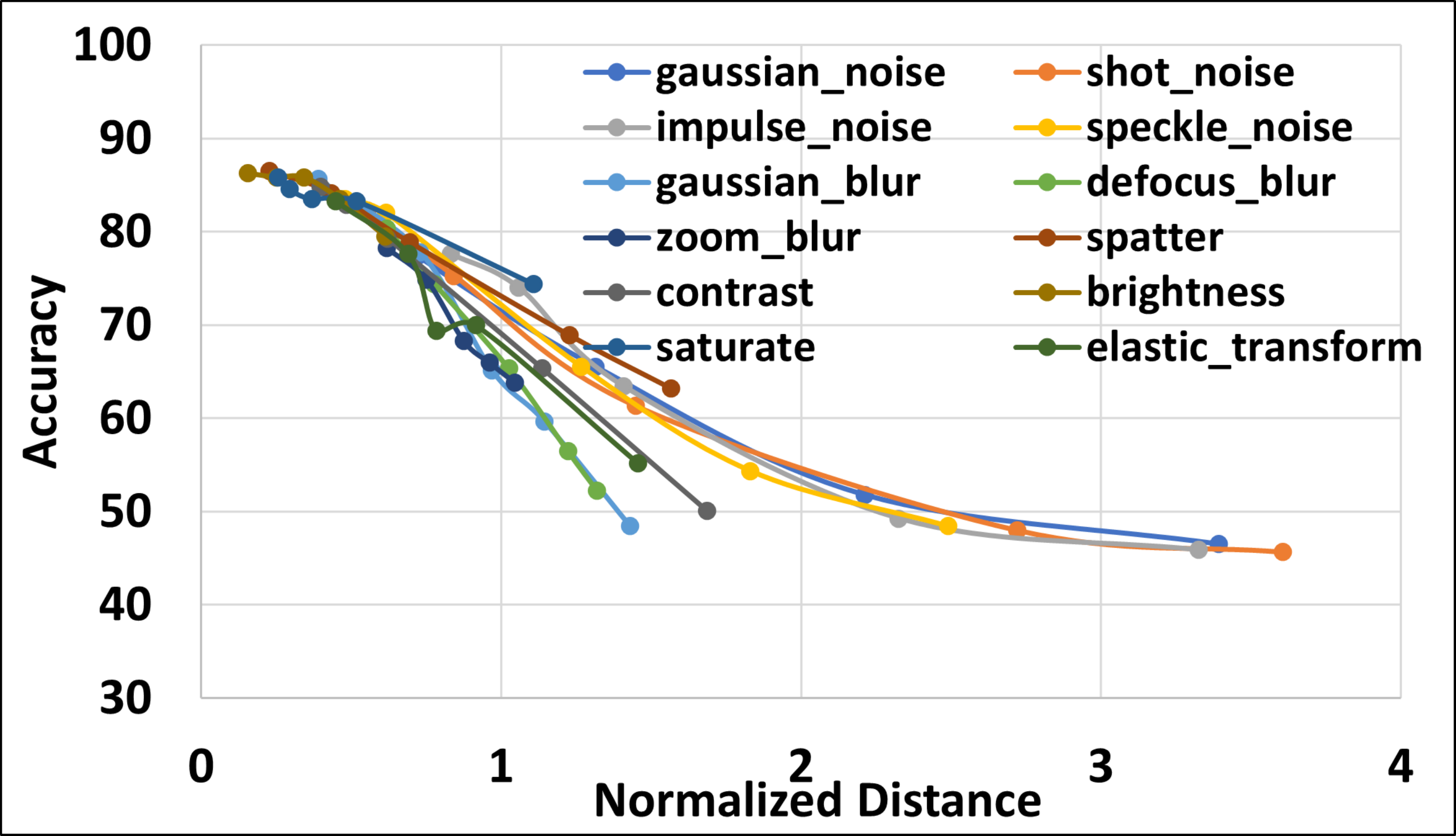}}
  \subfigure[DR-DG ($F=0.4$) with G2DM on VLCS ]{\includegraphics[width=0.31\columnwidth]{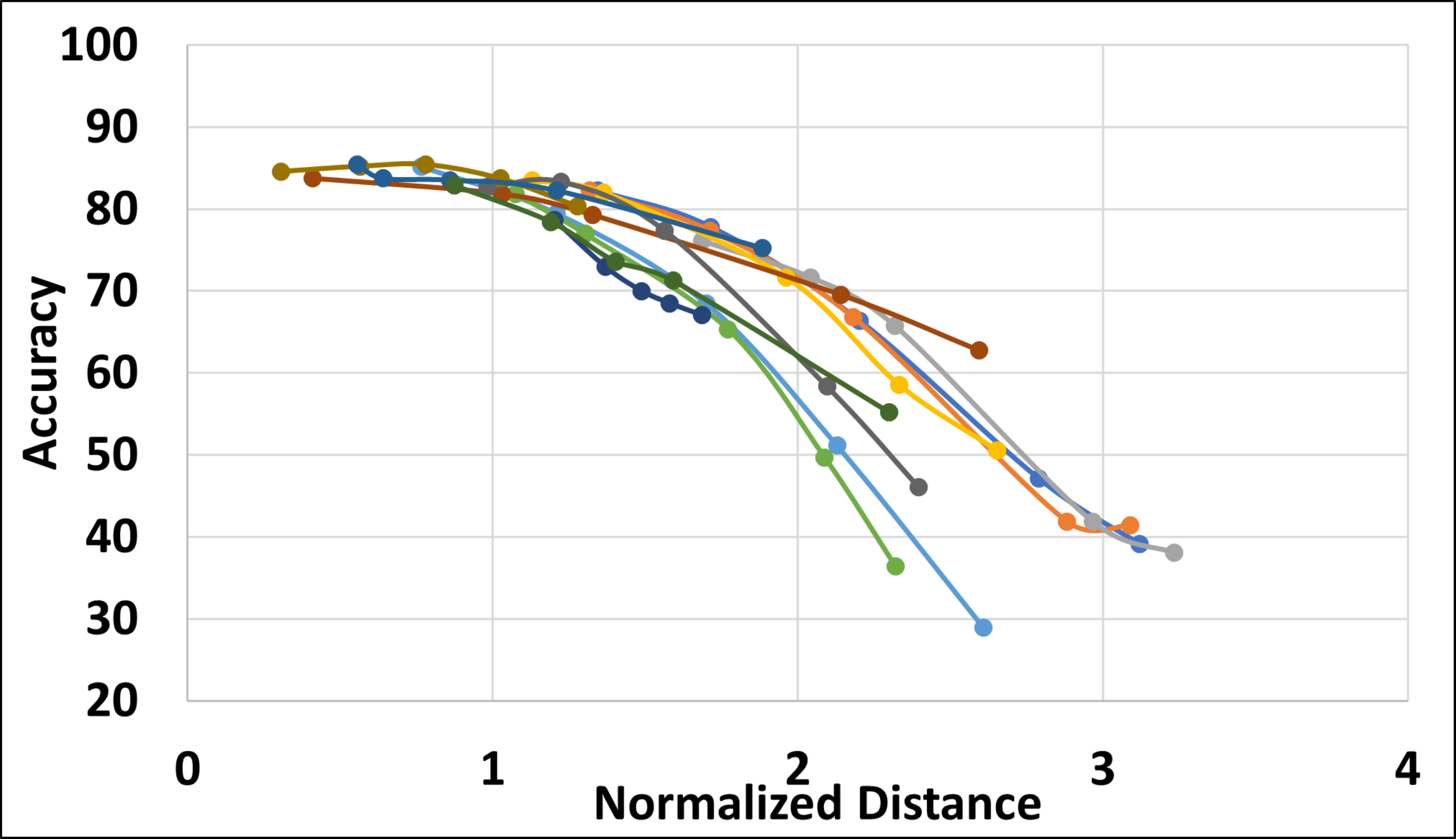}}
  \subfigure[DR-DG ($F=1.0$) with CDAN on VLCS ]{\includegraphics[width=0.31\columnwidth]{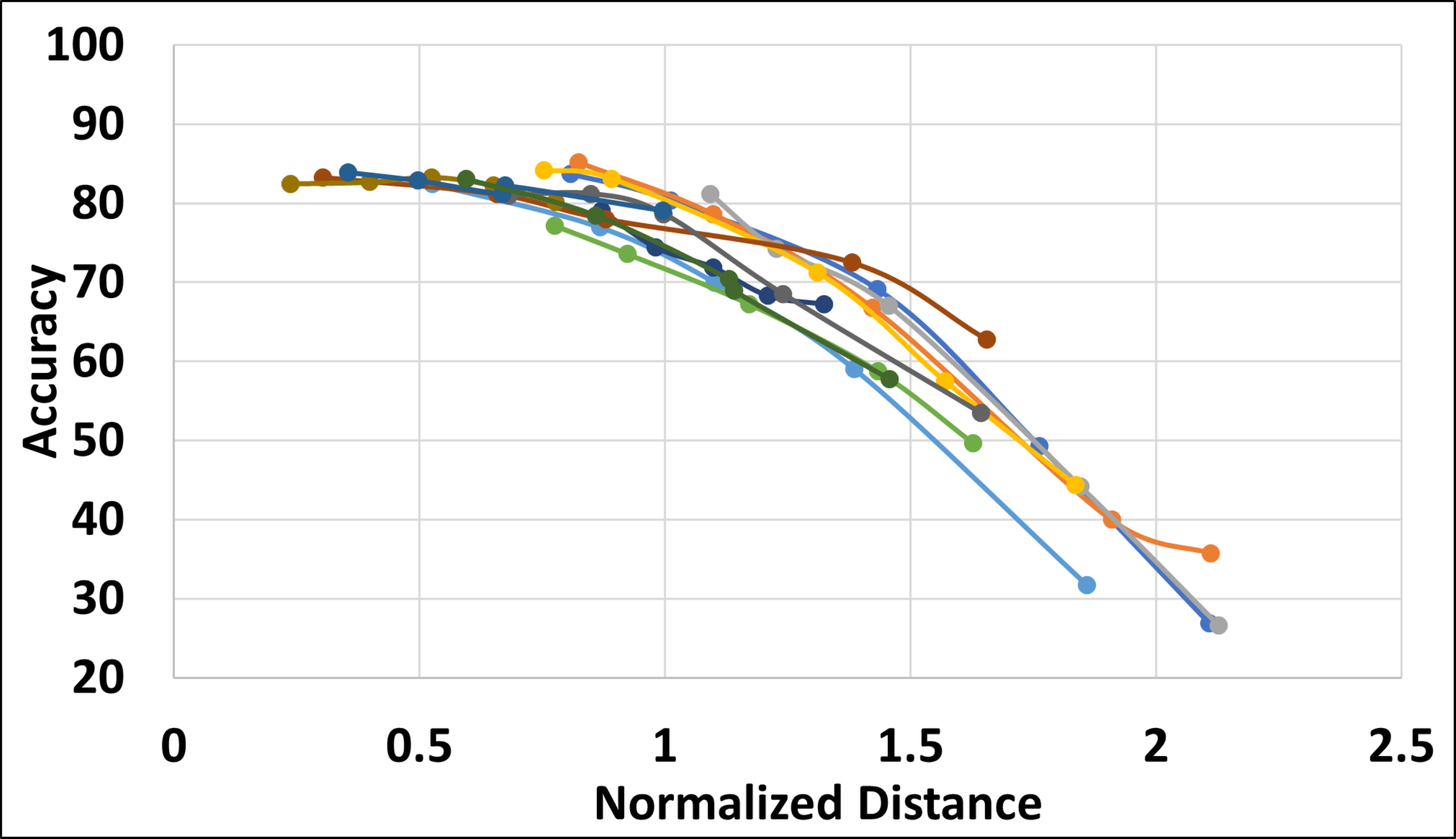}}

  \caption{(Best viewed in color.) Comparison of accuracy of the models on unseen distributions created by adding common corruptions to the source test set trained with Vanilla DG methods (rows 1, 3 and 5) and DR-DG (rows 2, 4, and 6) on R-MNIST (rows 1 and 2), PACS (rows 3 and 4), and VLCS (rows 5 and 6).}
  \label{fig:accuracy_before_after_all_datasets}
\end{figure}

Here, we provide a comparison of the accuracy of the models trained with DR-DG and models trained with vanilla DG methods on distributions generated by adding common corruptions to the test set.
Fig.~\ref{fig:accuracy_before_after_all_datasets} shows that the accuracy of DR-DG models doesn't degrade significantly compared to models trained with vanilla methods on these distributions. 
We emphasize that improving the performance on the distributions of corrupted data is not the primary goal of DR-DG and these results are presented for completeness. 

\if0
\begin{figure}[tb]
  \centering
  \subfigure[WM]{\includegraphics[width=0.32\columnwidth]{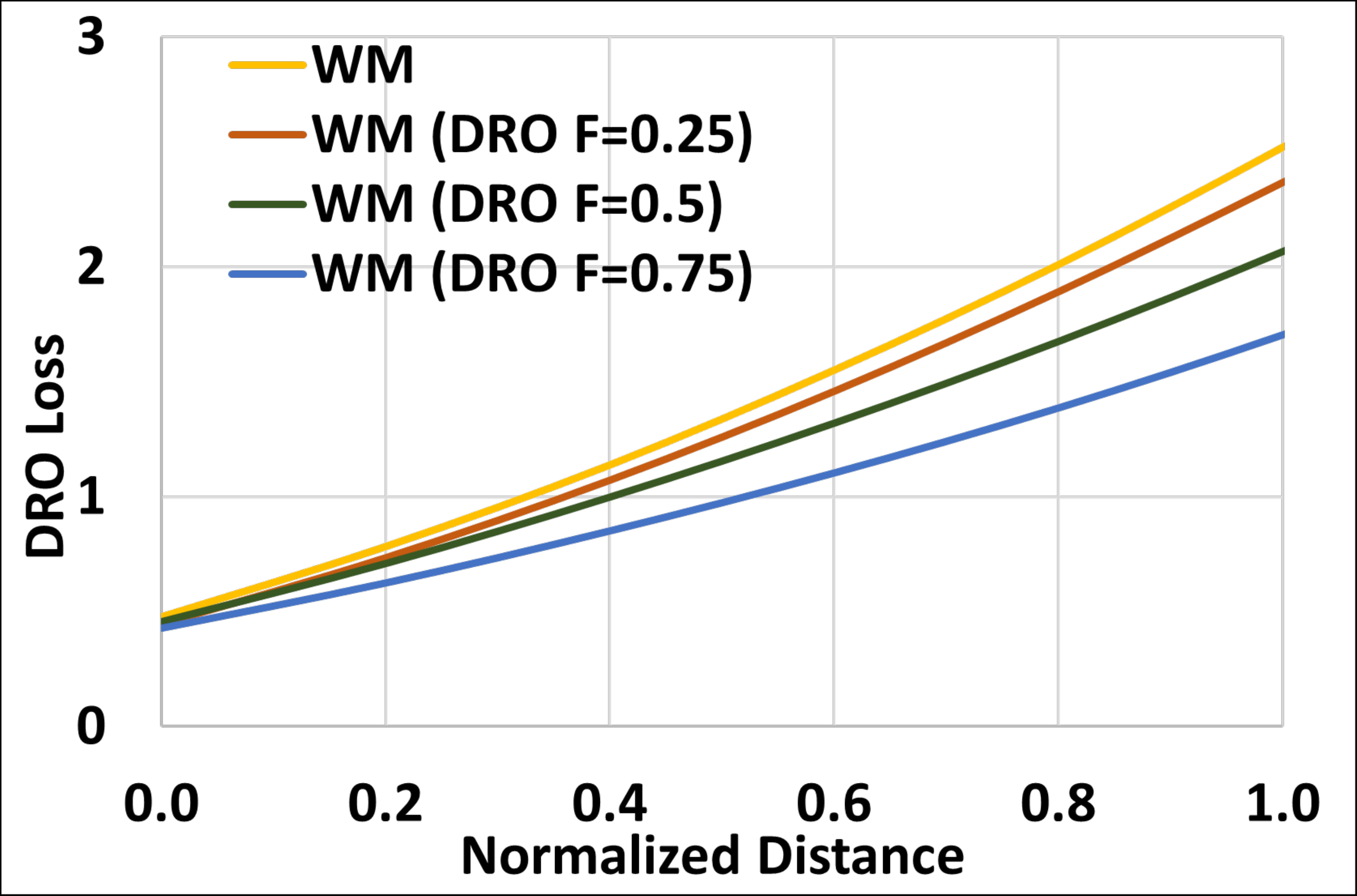}}
  \subfigure[G2DM]{\includegraphics[width=0.32\columnwidth]{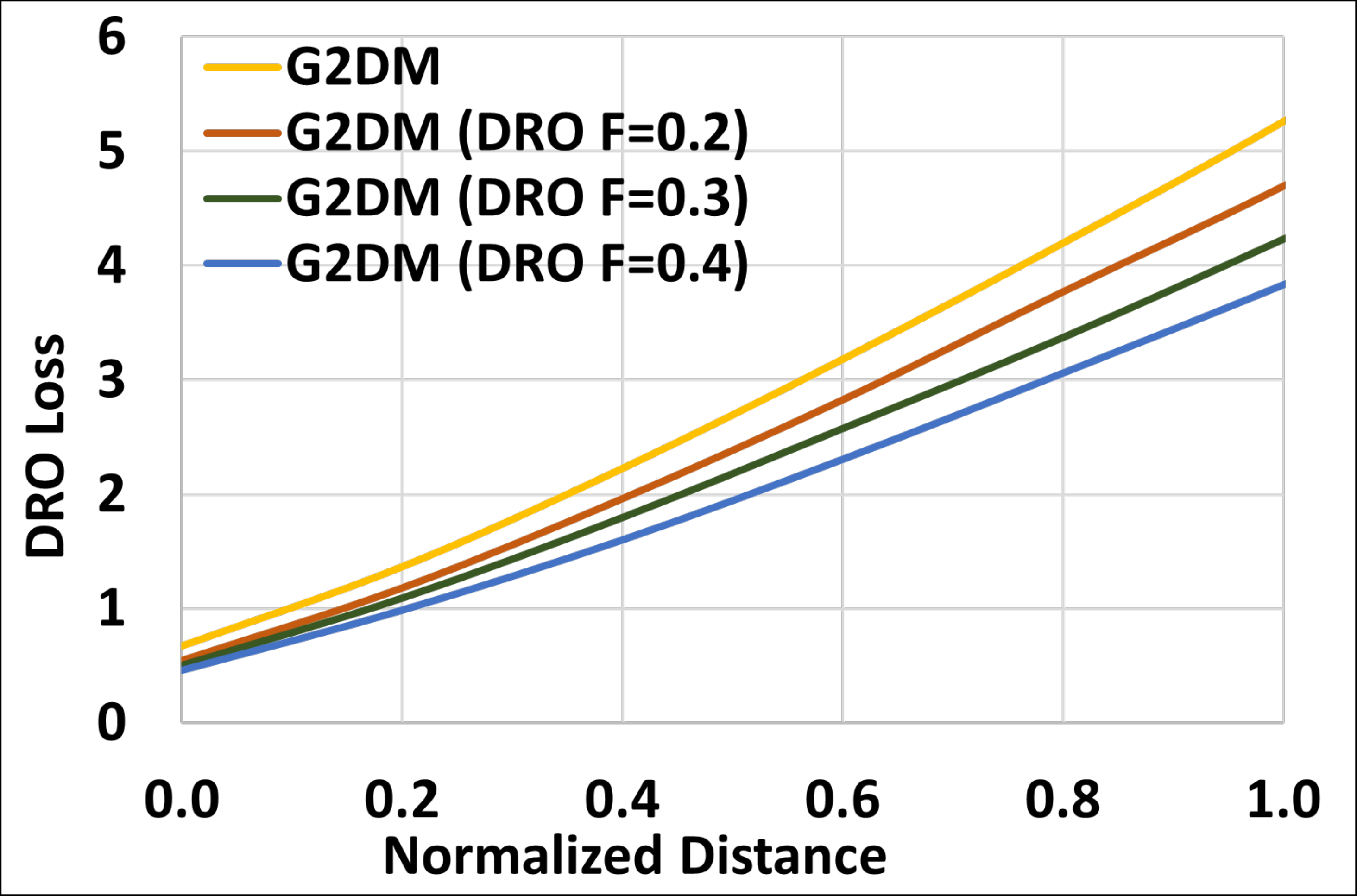}}
  
  \caption{(Best viewed in color.) Improved worst-case loss of models trained with WM and G2DM using DR-DG proposed in Alg.~\ref{alg:dro_training} on VLCS dataset. 
   }
  \label{fig:dro_trained_models_vlcs}
\end{figure}
\fi

\if0
\begin{figure}[tb]
  \centering
  \subfigure[VREX with R-MNIST]{\includegraphics[width=0.32\columnwidth]{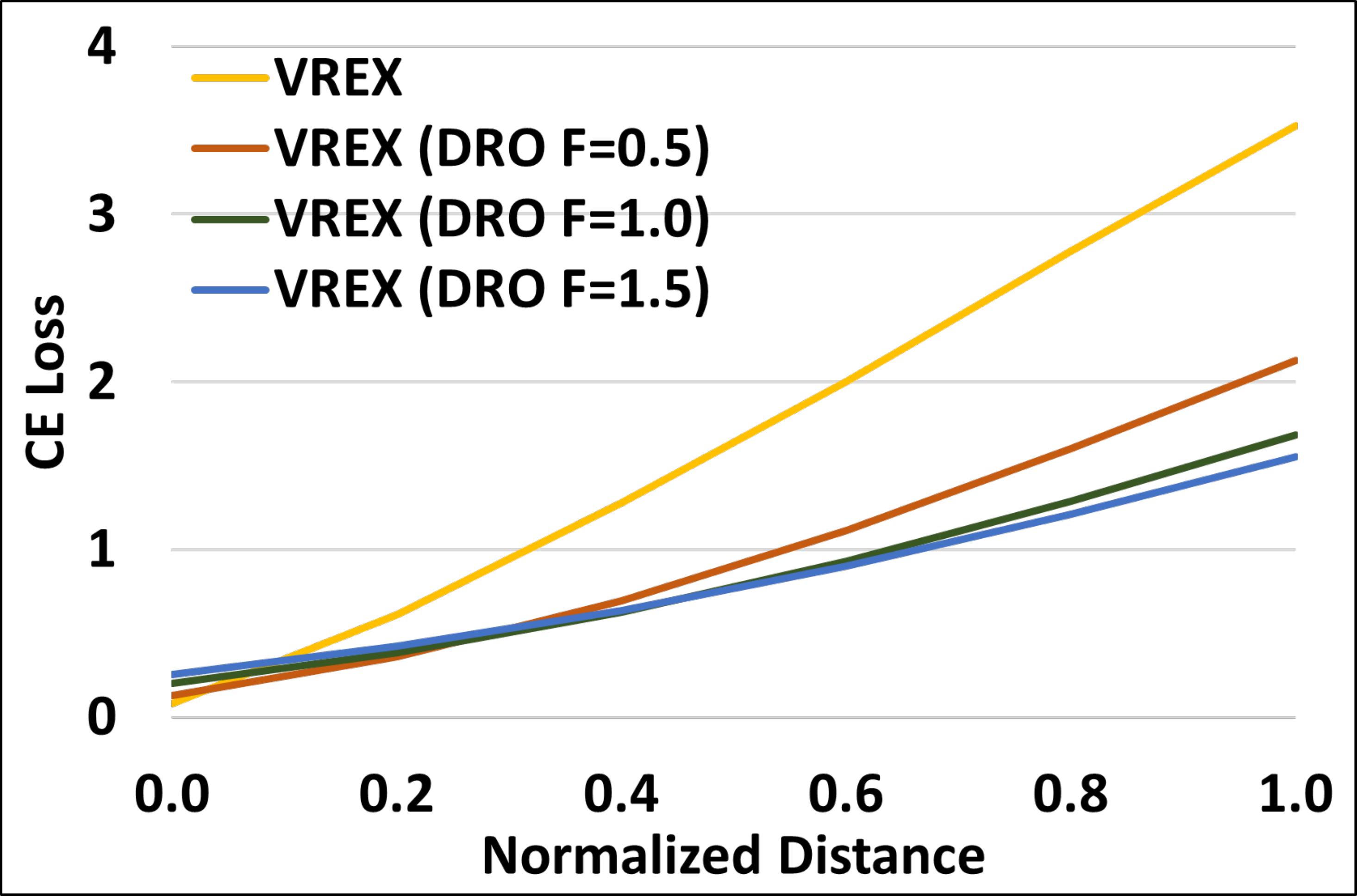}}
  \subfigure[VREX with PACS]{\includegraphics[width=0.32\columnwidth]{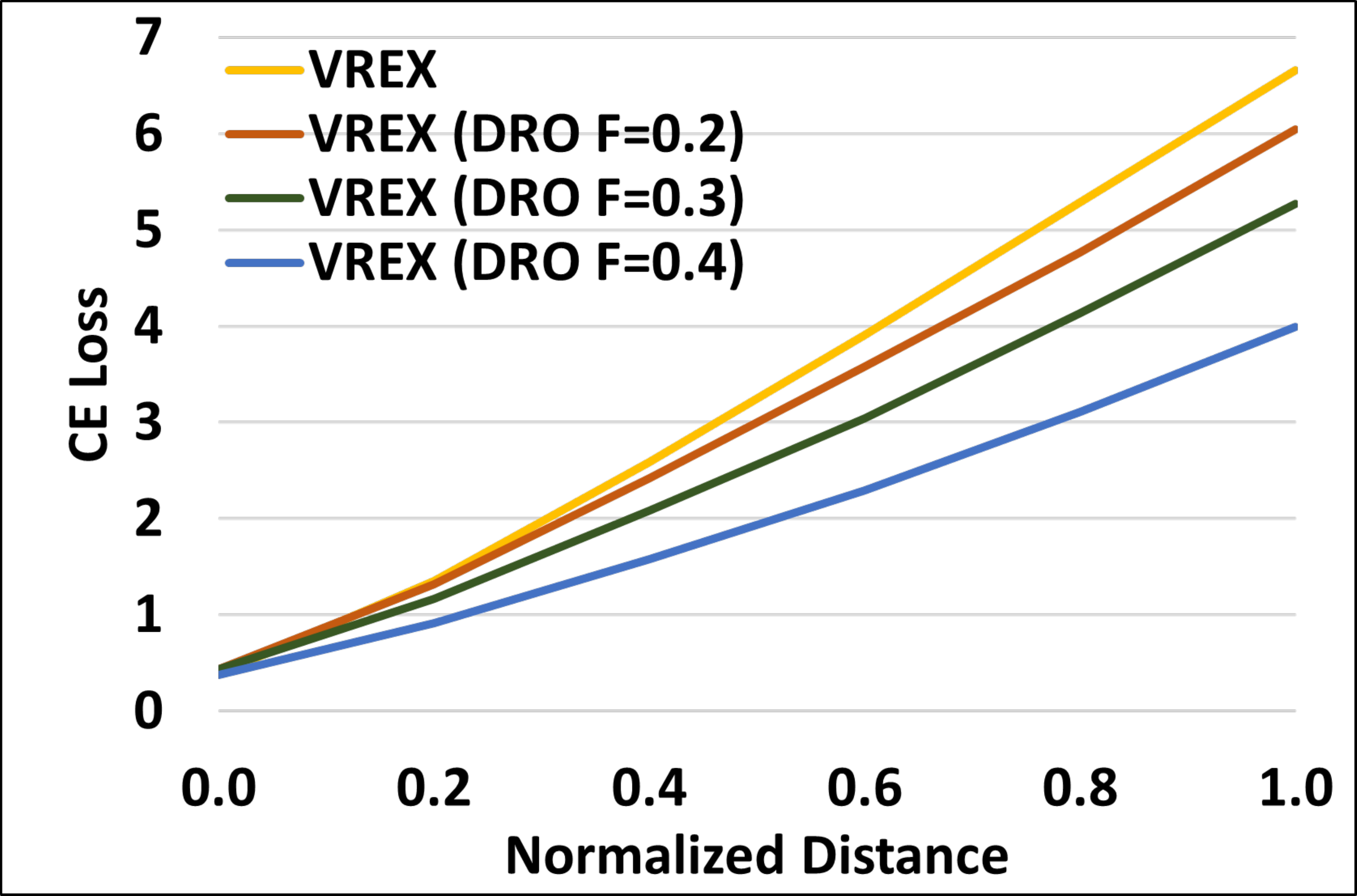}}
  \subfigure[VREX with VLCS]{\includegraphics[width=0.32\columnwidth]{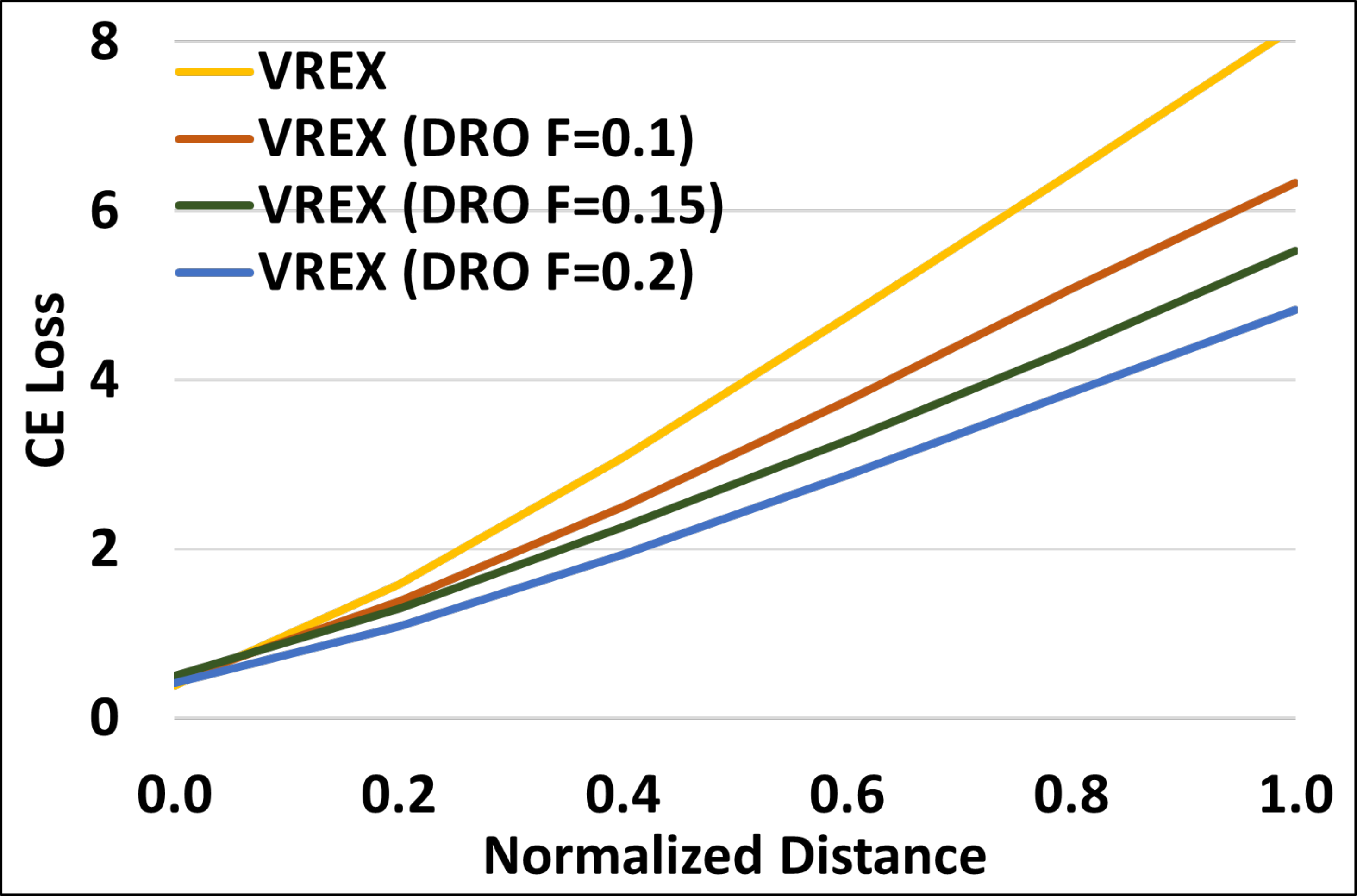}}
  \caption{(Best viewed in color.) Improved worst-case loss of models trained with VREX \cite{long2018conditional} using DR-DG proposed in Alg.~\ref{alg:dro_training} on R-MNIST, PACS and VLCS datasets. 
   }
  \label{fig:dro_trained_models_vrex}
\end{figure}
\fi

%%%%%%%%%%%%%%%%%%%%%%%%%%%%%%%%%%%%%%%%%%%%%%%%%%%%%%%%%%%%%%%%%%%%%%%%
\section{Review of DG algorithms and analyses}
\label{app:dg_algorithms}
%%%%%%%%%%%%%%%%%%%%%%%%%%%%%%%%%%%%%%%%%%%%%%%%%%%%%%%%%%%%%%%%%%%%%%%%

\subsection{DG algorithms}
In this section, we review the DG methods used in this work and describe their objective functions. For detailed descriptions of these algorithms, we refer the reader to original works. 

{\bf G2DM \cite{albuquerque2019generalizing}:} This algorithm relies on using $N_S$ one-vs-all discriminators to distinguish a source domain from other $N_S-1$ source domains. 
By learning a representation that fools the discriminator this method aims to align the source domains in the representation space. 
Mathematically, it solves the following problem 
\[
\min_{g, h}\max_{\tau_1, \cdots, \tau_{N_S}}\;\mathcal{L}(\mathcal{D}_S;h, g) - \sum_{k=1}^{N_S}\mathcal{L}_k(D_k(g(x)), y_k), 
\]
where $\mathcal{L}$ denotes the average loss of a given loss $\ell$ (we used cross-entropy loss) on all points in the the sources, $D_k$ denotes the $k^{th}$ discriminator with parameters $\tau_k$, $y_k$ denotes binary labels such that the points belonging to the $k^{th}$ source domain are labeled as zeros and points from all other domains are labeled as ones and $\mathcal{L}_k$ denotes the classification loss of the $k^{th}$ discriminator. 

{\bf CDAN \cite{long2018conditional}:} Similar to G2DM, this method also utilizes discriminators to align the source distributions.
The discriminators used in this approach are conditioned to use the labels of the source domain data for alignment. 
The overall mathematical objective remains the same as G2DM except for the discriminators in G2DM are replaced by conditional discriminators which use multi-linear conditioning (see \cite{long2018conditional} for more details).  
For CDAN, we used binary discriminators which use a one-vs-all strategy similar to G2DM.

{\bf Wasserstein Matching (WM):} Many previous works have demonstrated the effectiveness of using Wasserstein distance for domain alignment, especially in the domain adaptation literature \cite{shen2018wasserstein, damodaran2018deepjdot}.
Building on that we propose to use Wasserstein matching to reduce the divergence between the source domains in the representation space similar to G2DM and CDAN. 
Mathematically, our objective still remains the same as G2DM, except that the discriminators are now replaced with Wasserstein distance-based terms which consider alignment in a one-vs-all fashion. 
Concretely, our objective becomes 
\[
\min_{g, h}\;\mathcal{L}(\mathcal{D}_S;h, g) + \sum_{k=1}^{N_S}W_2^2(\mathcal{D}_S^k, \mathcal{D}_S \backslash \mathcal{D}_S^k), 
\]
where $\mathcal{D}_S \backslash \mathcal{D}_S^k$ denotes all source domains except the $k^{th}$ source domain. 
To efficiently solve this problem we use the methodology proposed in \cite{damodaran2018deepjdot}, where the coupling is computed batch-wise using a fixed cost matrix and then this coupling is used to optimize the cost matrix. 
The dissimilarity cost between points $(x,y)$ and $(x', y')$ used to populate the cost matrix is computed using the feature distance i.e. $\|g(x) - g(x')\|_2^2$ and label distance i.e. $\|y - y'\|_2^2$ (assuming $y$ are one-hot encoded vector of labels). Thus the overall dissimilarity cost between two points $(x,y)$ and $(x',y')$ is $\|g(x) - g(x')\|_2^2 + \lambda \|y - y'\|_2^2$. In our experiments we set $\lambda=1$. 
%\AM{Refer to Shen and mention the algorithm is similar to theirs.}

{\bf VREX \cite{krueger2021out}:} 
This work proposed to use variance of the risks as a regularizer (VREX) to learn models which can generalize well to unseen domains. In particular, they proposed adding an additional objective to the average error of the source domains, thereby solving  
\[
    \min_{g,h} \mathcal{L}(\mathcal{D}_S;h, g) + \beta \mathrm{Var}(\{ \mathcal{L}(\mathcal{D}^1_S;h, g), \cdots, \mathcal{L}(\mathcal{D}^{N_S}_S;h, g)\}).
\]
In our experiments we fix the parameter $\beta$ to be 1. 
%Unlike previous methods this method does not try to align the domains in the representation space but rather learns a hypothesis 

\subsection{Analyses of DG using distributional divergence}
In this section, we review some of the recent analyses of the DG problem which have demonstrated that generalization to an unseen domain can be studied in terms of the distributional distance between the source and the unseen distributions. Please refer to the original works for detailed descriptions of the assumptions and results. 

{\cite{ben2007analysis}, \cite{shen2018wasserstein}:} 
In one of the early works, it was shown that performance on an unknown target domain can be estimated based on the performance on the source domain and the distance between the marginal distributions and labeling functions of the two domains. 
Let a domain be $(P, f)$, where $P$ denotes the distribution on inputs $\mathcal{X}$ and $f:\mathcal{X}\rightarrow[0,1]$ denotes the labeling function. 
Let $h:\mathcal{X}\rightarrow\{0,1\}$ be a hypothesis and $\mathcal{E}_P(h,f) := \mathbb{E}_{x \sim P}[|h(x) - f(x)|]$ denote the risk of the hypothesis $h$. 
Then it was shown in Theorem 1 of \cite{ben2007analysis} that 
\[
\mathcal{E}_T(h, f_T) \leq \mathcal{E}_S(h, f_S) + d_1(P_S, P_T) + \min\{\mathbb{E}_{P_S}[|f_S(x) - f_T(x)|], \mathbb{E}_{\mathcal{D}_T}[|f_S(x) - f_T(x)|]\},
\]
where $d_1$ denotes the total variation distance.
\cite{shen2018wasserstein}, showed a similar result (Theorem 1 of \cite{shen2018wasserstein}) using type-1 Wasserstein distance for all $K-$Lipschitz continuous hypotheses i.e., 
%{\bf ``Wasserstein Distance Guided Representation Learning''}
%Theorem 1. Under assumptions of Lemma 1, for every $h$ the following holds:
\[
    \mathcal{E}_T(h, f_T) \leq \mathcal{E}_S(h, f_S) + 2 K\cdot W_1(P_S, P_T) + \lambda,
\]
where $\lambda$ is the combined error of the ideal hypothesis $h^\ast$ that minimizes the combined error $\mathcal{E}_S(h, f_S) + \mathcal{E}_T(h, f_T)$.

%{DRO based analyses:}

{\cite{kumar2022certifying}}:
Recent work showed that the accuracy of smoothed classifiers, smoothed using different smoothing functions under distribution shifts can also be bounded using a function dependent on the Wasserstein distance between the source and target domains. 
Theorem 4.1 \cite{kumar2022certifying} shows that for a function $h:\mathcal{X} \times \mathcal{Y} \to [0,1]$, whose smoothed version is defined as $\overline{h}(x, y) := \mathbb{E}_{x'\sim S(x)}[h(x', y)]$,
\[
    |\mathbb{E}_{(x_1, y_1) \sim S}[\overline{h}(x_1, y_1)] - \mathbb{E}_{(x_2, y_2) \sim T}[\overline{h}(x_2, y_2)]| \leq \psi(\rho),
\]
where $\rho$ is the radius of the Wasserstein ball around $S$.

{\cite{sehwag2021robust}}:
Another recent work showed that distribution divergence can also be used to explain the transfer of robustness across the two domains. In particular, Theorem 1 \cite{sehwag2021robust} shows that the difference in the average margin ($Rob(h, P) := \mathbb{E}_{(x,y)\sim P}[\inf_{h(x') \neq y} \|x'-x\|]$) of the classifier $h:\mathcal{X}\rightarrow\mathcal{Y}$ under distribution shift is bounded by conditional Wasserstein distance between the two domains (see the original paper for definition). 
\[
|Rob(h, P_S) - Rob(h, P_T)| \leq W_{cond}(P_S, P_T).
\]
Assuming robust error (RE$_\epsilon$) at a perturbation $\epsilon$ is defined as the probability of having margin ($mar := \inf_{h(x') \neq y}\|x-x'\|$) less than $\epsilon$ i.e., $RE_{\epsilon} := Pr[mar < \epsilon]$ and average margin $\mathbb{E}[mar]$. 
Using Markov's inequality, we have $1 - RE_{\epsilon} \leq \epsilon^{-1}\mathbb{E}[mar]$. Since $1 - RE_{\epsilon} = RA_{\epsilon}$, the robust accuracy at $\epsilon$, we get a bound on the robust error under distribution shift, as follows
%we can see that doing adversarial training which improves the robust accuracy also helps to improve the average margin. 
%(
%Markov's inequality: $P(X\geq a)\leq E[X]/a$ for a nonneg RV $X$ and $a>0$.
%)
%So, according to ``Can Proxy...", we have
\[
|RE_{\epsilon}(P_S)-RE_{\epsilon}(P_T)|\leq \epsilon^{-1} W_{cond}(P_S,P_T).
\]

%Marginal Wass dist is same or smaller than Labeled Wass dist by definition: 
%$W_{mar}((x,y),(x',y')) = W(x,x') \leq W((x,y),(x',y'))$.

Along with these works, a large body of works exists in the DRO literature which explicitly considers their uncertainty set to be based on different divergence measures such as the Wasserstein distance, $f-$divergences such as Jenson Shannon, Kullback–Leibler, Hellinger distance as discussed in the related work section of the main paper. 
This highlights the important role of distributional divergence in certification as we are proposing in the paper to assess the performance of domain generalization methods on new unseen domains. 
\section{Additional discussions}
%%%%%%%%%%%%%%%%%%%%%%%%%%%%%%%%%%%%%%%%%%%%%%%%%%%%%%%%%%%%%%%%%%%%%%%%
\label{app:discussion}
In this section, we discuss certification using different loss functions in Cert-DG and also present a comparison of input space versus representation space certification. 
%Additionally we demonstrate how our DR-DG method guarantees 

%\begin{wrapfigure}[16]{r}{0.65\textwidth}
\begin{figure}[tb]
\vspace{-0.5cm}
  \centering
  %\subfigure[WM with R-MNIST]{\includegraphics[width=0.43\columnwidth]{Images/certification_ce_vanilla_wm_rotatedmnist.pdf}}
  \subfigure[CE loss]{\includegraphics[width=0.31\linewidth]{Images/certification_ce_vanilla_wm_rotatedmnist.pdf}}
  \subfigure[Modified hinge loss]{\includegraphics[width=0.31\linewidth]{Images/certification_hinge_vanilla_wm_rotatedmnist.pdf}}
  \subfigure[0/1 loss]{\includegraphics[width=0.324\linewidth]{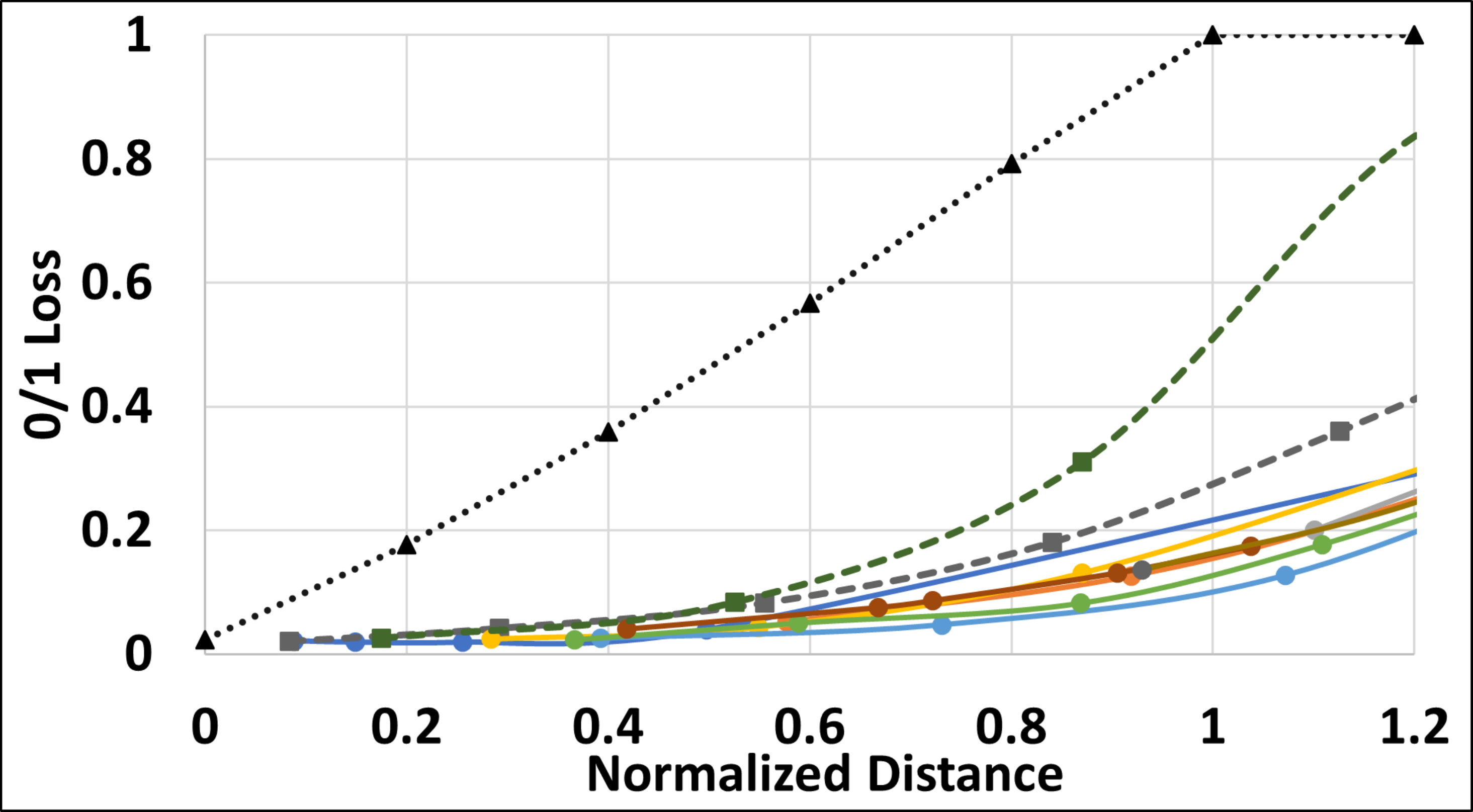}
}
%\caption{Comparison of certified loss and empirical loss on corrupted and adversarial domains of a model trained with WM on R-MNIST. The large gap between the certified and empirical losses indicate the insufficiency of evaluation of DG methods on a few unseen domains to guarantee generalization on other domains. Cross-entropy (left) and modified hinge loss (right) were used.
%Compatibility of our certification procedure with different loss functions such as the cross-entropy and the modified hinge loss. The certified loss provides a meaningful measure of the performance of the models on unseen natural and adversarial domains.
%\BK{provide detail on data/method}
%}
  
  %\subfigure[WM with R-MNIST]{\includegraphics[width=0.44\columnwidth]{Images/certification_hinge_vanilla_wm_rotatedmnist.pdf}}
  %\hfill
  %\subfigure[DR-DG (with WM) on R-MNIST]{\includegraphics[width=0.48\linewidth]{Images/certification_misclassification_dro_wm_rotatedmnist.pdf}}
 %\vspace{-0.3cm}
  \caption{(Best viewed in color.) Comparison of certified loss and empirical loss on corrupted and adversarial domains of a model trained with WM on R-MNIST. The models trained with Vanilla WM on R-MNIST are certified using Cert-DG with three loss functions. %This demonstrates the compatibility of our certification framework with different loss functions. 
}
  \label{fig:certification_various_losses}
\end{figure}
%\end{wrapfigure}

\subsection{Certification using different loss functions}%Other losses}
\label{app:other_losses}
As discussed in the main paper, the proof of strong duality (Eq.~\ref{Eq:strong_duality}) has been shown to hold under different assumptions as well as loss functions. Here we show that Cert-DG can be used with the hinge loss and the 0/1 loss; the latter can be used to certify the accuracy of the DG models.
%when theAlong with the cross entropy loss we discuss the compatibility of Cert-DG with different loss functions.

%Why choose CE loss? Because it's differentiable and it's what models-to-be-certified are trained with. However, as we discussed in the previous section, other losses are also possible and may even be more useful under different conditions.

\subsubsection{Hinge loss}
% Closed-form approach
When the loss is piece-wise affine $l(z) = \max_{k\leq K}\; [a_k^Tz + b_k]$ such as the hinge loss,
%it is shown that the complexity of the optimization is linear with $N$ \AM{What is N?}, and 
the dual problem can be written as a simple linear programming (Corollary 5.1, Remark 6.6 \cite{mohajerin2018data}, Remark 9 \cite{gao2016distributionally}), and therefore the optimum  
$\inf_{\gamma\geq0}\left[\gamma \rho^2 + E[\phi_\gamma(z_0,y_0)]\right]$ 
can be computed accurately and efficiently.
% GD approach
Furthermore, we try to find the optimum for an unconstrained saddle-point problem using SGD in our approach. 
However, the vanilla hinge loss requires modification to be used with SGD, because the loss gradient is zero ($\nabla_z \ell(z)=0$) for correctly classified point $z$,
and therefore SGD does not progress. We address this by using the modified hinge loss (similar to leaky ReLU) 
\begin{equation}
    \ell_\mathrm{modified\_hinge}(t) = \max \{0,\;1 - t\} - \alpha \max \{0,\;t-1\}
\end{equation}
where $t$ denotes the difference between the logit of the true class and the maximum logit of the other classes i.e., $t=h_{y}(x) - h_{other}(x)$ for a point $(x, y)$ and $\alpha$ is a small constant such as 0.1. Different from the vanilla hinge loss, the modified loss can have negative values for correctly classified points. This may be undesirable when the objective is loss minimization but has no impact when the objective is loss maximization as is required for certification.

\subsubsection{0/1 loss}
Here we describe how our certification method can be used to certify the accuracy of the models trained with DG methods. 
We consider 0/1 loss for certifying accuracy. 
Using the 0/1 loss, the solution to the robust surrogate loss, Eq.~\ref{eq:rep-surrogate}, can be computed in closed form as shown below: %maximization 
%Accuracy can also be certified. 
%The maximum value Eq~{} is computable in closed form as follow:
\begin{equation}
\sup_{z}\;\{ \ell(z) - \gamma c(z,z_0) \}
= \max\{0,\;1-\gamma \|z_0-z_{adv}\|_2^2\},
\if0
=\left\{
\begin{array}{cc}
     %1-\gamma \|z_0-z_{adv}\|, & \mathrm{if}\; \gamma \|z_0-z_{adv}\| \leq 1\\
     1-\gamma \|z_0-z_{adv}\|_2^2, & \mathrm{if}\;   1 - \gamma \|z_0-z_{adv}\|_2^2 \geq 0\\
     0, & \mathrm{otherwise}
\end{array}
\right.
\fi
\end{equation}
where $(z_0,y_0)$ is the initial point and $z_{adv}=\arg\min_z\;\|z - z_0\|_2^2\;\;\mathrm{s.t.}\;\;h(z)\ne y_0$.
Therefore the objective for certification is 
\begin{equation}
%\gamma \rho^p +\frac{1}{N}\sum_i (1-\gamma \|z_0-z_{i,adv}\|) I[\gamma \leq \|z_i-z_{i,adv}\|^{-1}]    
\inf_{\gamma \geq 0}\;\left\{ \gamma \rho^2 +\frac{1}{N}\sum_i \max\{0,\;1-\gamma \|z^i_0-z^i_{adv}\|\}\right\}.
\end{equation}

The objective function is therefore a non-negative, piece-wise linear, non-continuous,
and convex function 
of $\gamma$.
The non-continuity of the function prevents us from using SGD to optimize over $\gamma$. 
However, minimization over $\gamma$ is just a one-dimensional convex optimization even though it's non-differentiable. 
This problem can be efficiently solved using simple scalar minimization methods (such as ``scipy.optimize.minimize\_scalar'') based on the bisection method to solve for the optimal $\gamma$.
For practically solving the certification problem we compute $z_{adv}$ using CW attack \cite{carlini2017towards} and then solve the minimization problem over $\gamma$ using the $minimize\_scalar$ function with the bounded solver and bound the value of $\gamma$ to be in [1E-20, 100]. 
We present certification results on a model trained with WM on R-MNIST in Fig.~\ref{fig:certification_various_losses}(c).
The certified loss at normalized distance of 1 is the highest since our definition of normalized distance use the adversarial distribution which is the nearest distribution (to the sources in the representation space) with zero accuracy. 

While our method can be used for certifying the model performance with different loss functions such as the cross-entropy loss, hinge loss, or the misclassification (0/1) loss efficiently, we consider only cross-entropy loss for DR-DG.
This choice is due to the differentiability of the worst-case loss and also the practical observations that most deep learning models perform better when trained with the cross-entropy loss compared to other losses such as hinge loss.

\subsection{Input-space vs representation-space certification}
\if0
What to say here.
Pros and cons of certification in input space vs rep space
    
\begin{itemize}
    \item Difficulty of input space certification (experiment results)
    \item Easiness of certifying in rep space (vs input space)
    \item Certifying on intermediate layers
    \item Why it is okay to certify in rep space only for DG?
    \item Adversarial dist in rep vs input space (potential overestimation of the worst-case loss) - experiment?
    \item PGD attack in rep vs input space
    \item Summary of why we use rep space in cert-DG and DR-DG.
\end{itemize}
\fi

In this section, we discuss the differences in certifying the worst-case loss considering the Wasserstein-ball in the input space (Eq.~\ref{eq:surrogate_loss}) versus in the representation space (Eq.~\ref{eq:rep-surrogate}) and discuss why certification in the representation space is easier both from a theoretical and a practical standpoint.

Certification guarantees when distance between the distributions is measured in the input space are vacuous since the distributional distance between the source distributions and the unseen distributions can be large (i.e., when $c((x_0,y_0), (x_1,y_1)) = \|x_0 - x_1\|_2^2 + \infty \cdot I[y_0\neq y_1]$ in the Wasserstein distance).
For example, many previous works have shown that DG methods perform reasonably well on the R-MNIST dataset.
But, we observe that the Wasserstein distance measured in the input space between the source distributions and the well-performing target distribution is very large. 
In particular, models trained with WM on rotation angles $0^\circ, 15^\circ, 30^\circ, 45^\circ$, and 60$^\circ$ yield $\sim$95\% accuracy on a domain with rotation angle 75$^\circ$. 
But the normalized distance of the 75$^\circ$ domain from the sources is roughly 5, which is too far from the sources to provide any meaningful guarantees. 
Thus, the high performance of DG methods is uncertifiable in the input space i.e. as depicted in Fig.~\ref{fig:method_explanation}, there could be other distributions ($P_{T_{bad}}$) at the same distance from the source with a worse loss.
In contrast, the representation space learned by DG methods is much more useful for certification since the distance between the distributions is explicitly reduced by either using Wasserstein matching or discriminator-based approaches or DR-DG. 
Although there are distinct trade-offs between different methods of alignment, unseen distributions in general lie much closer to the sources compared to their distances in the input space. 
As long as DG algorithms can achieve this, certification using the representation-space distance is preferable compared to the input-space distance.

A major advantage of computing the worst-case loss by considering the Wasserstein ball in the representation space is the ease of solving the robust surrogate in Eq.~\ref{eq:rep-surrogate}) compared to solving the version in Eq.~\ref{eq:surrogate_loss}, which considers the ball in the input space. As discussed extensively in \cite{sinha2017certifying}, finding the maximizer of the robust surrogate loss is in general an NP-hard problem for networks with ReLU activations. 
Even if the ReLU activations are changed to smoother ones, there are still problems with provably maximizing the robust surrogate loss which has been shown to be possible only when additional assumptions such as bounded Lipschitz gradient w.r.t to the input and the model parameters are imposed.
These assumptions are prohibitive and restrict certification only for small to medium size problems. 
However, when certifying in the representation space, we don't need these assumptions since representation space is usually followed by a single softmax layer, and maximization of the robust surrogate is much easier both theoretically and practically.
In this work we restrict the certification only in the second to last layer. Although certification can be applied to other layers but additional assumptions are needed to guarantee convergence of the certification procedure. Some of these assumptions, as discussed earlier, could be prohibitive for large neural networks or may require changes to the architecture. Moreover, as we consider layers closer to the input, it becomes more difficult for DG algorithms to reduce the distance between the sources, thereby leading to poor/vacuous certification guarantees.
%without affecting the performance. 

\begin{figure}[tb]
  \centering
  \subfigure[WM]{\includegraphics[width=0.4\columnwidth]{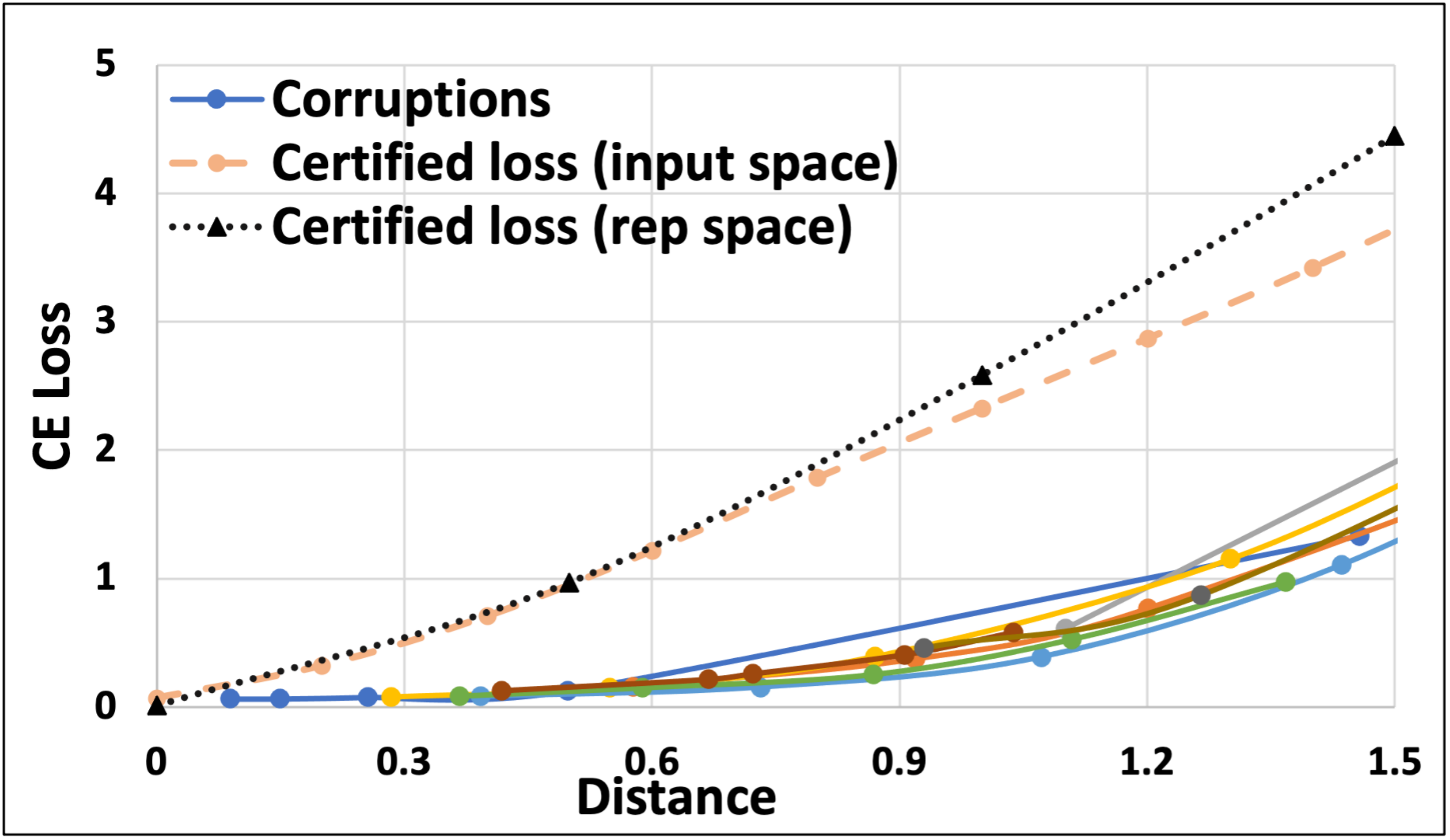}}
  \subfigure[G2DM]{\includegraphics[width=0.4\columnwidth]{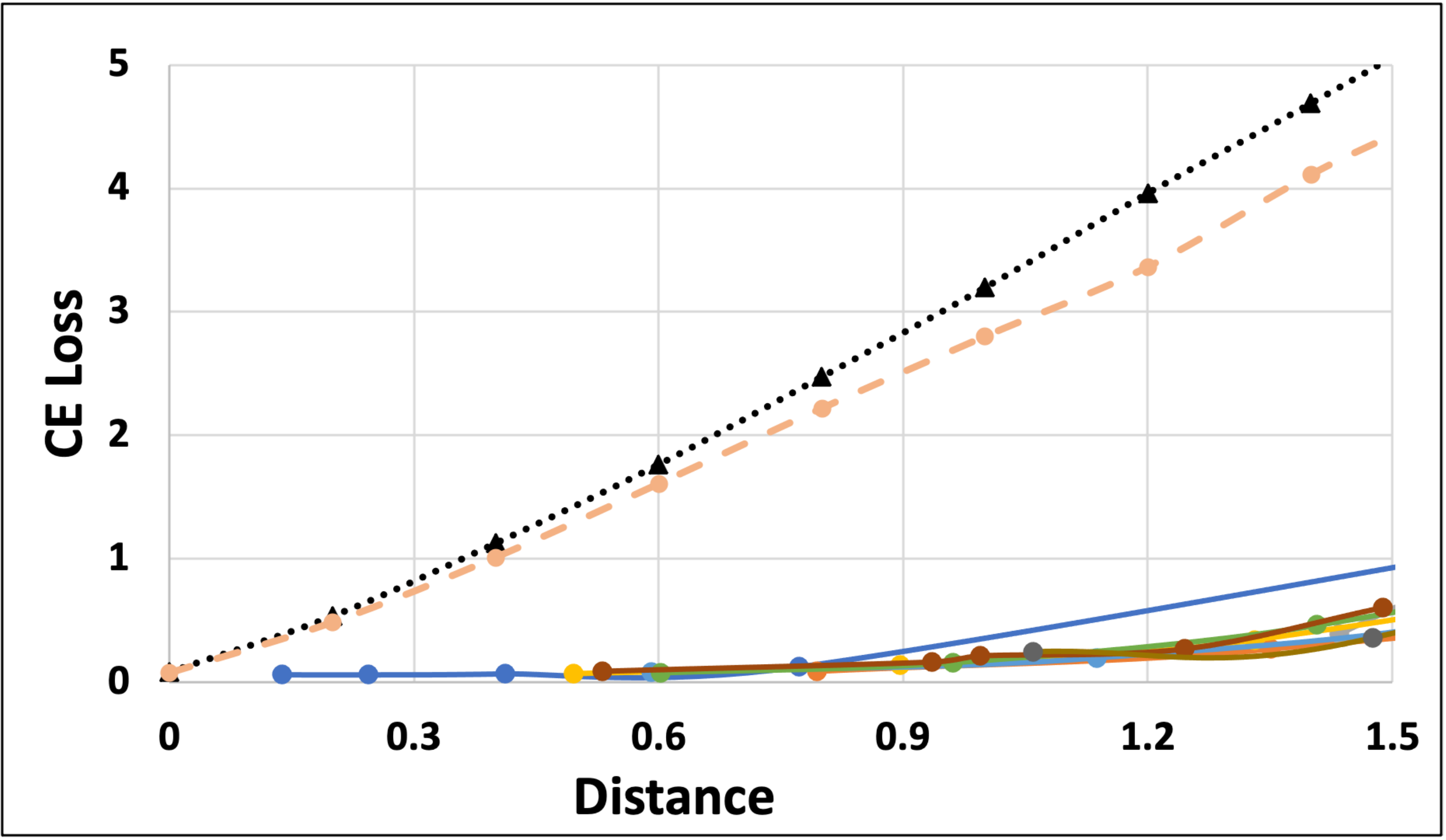}}
  
  \subfigure[DR-DG ($F$ = 0.5) with WM ]{\includegraphics[width=0.4\columnwidth]{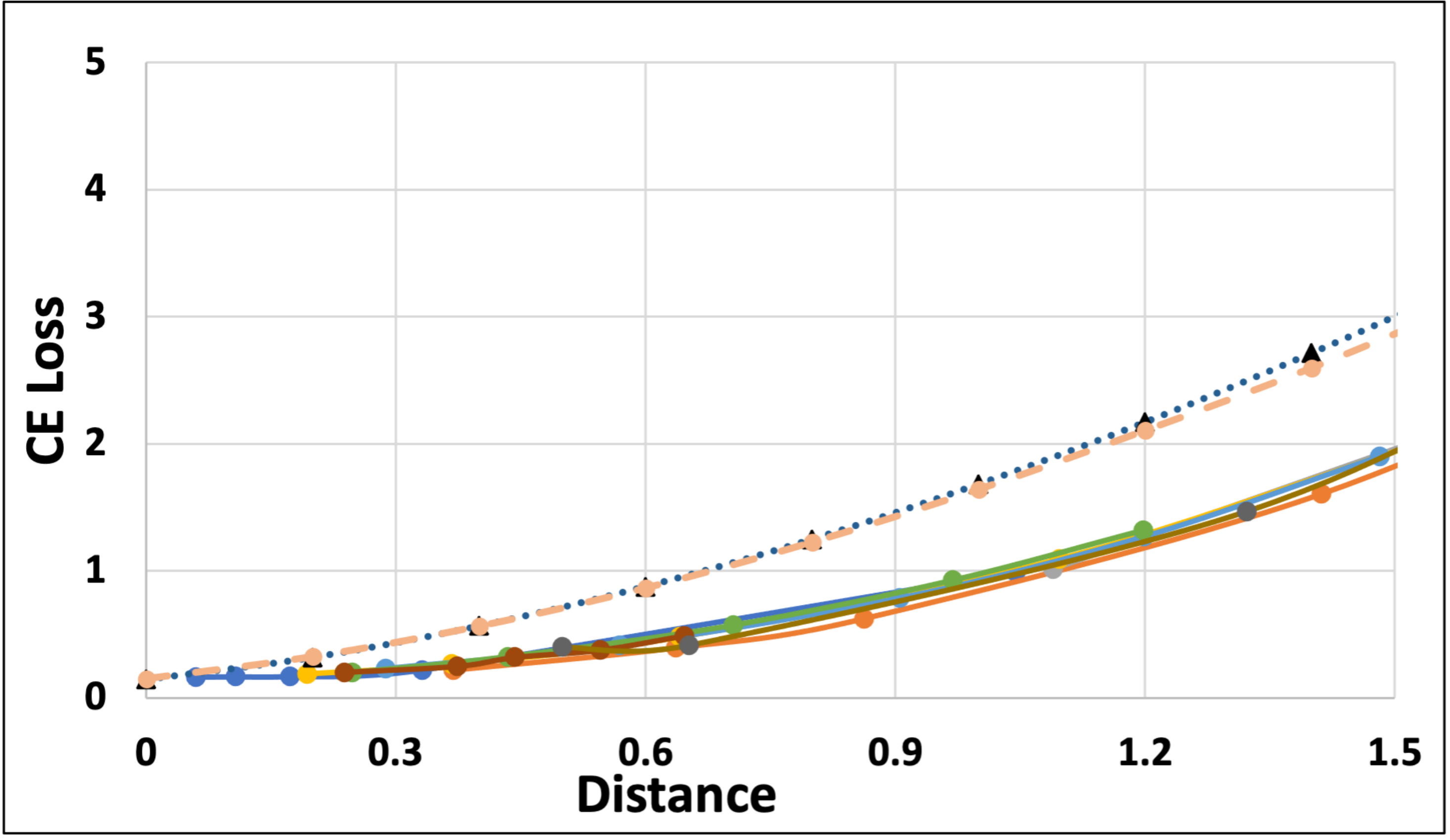}}
  \subfigure[DR-DG ($F$ = 1.5) with G2DM]{\includegraphics[width=0.4\columnwidth]{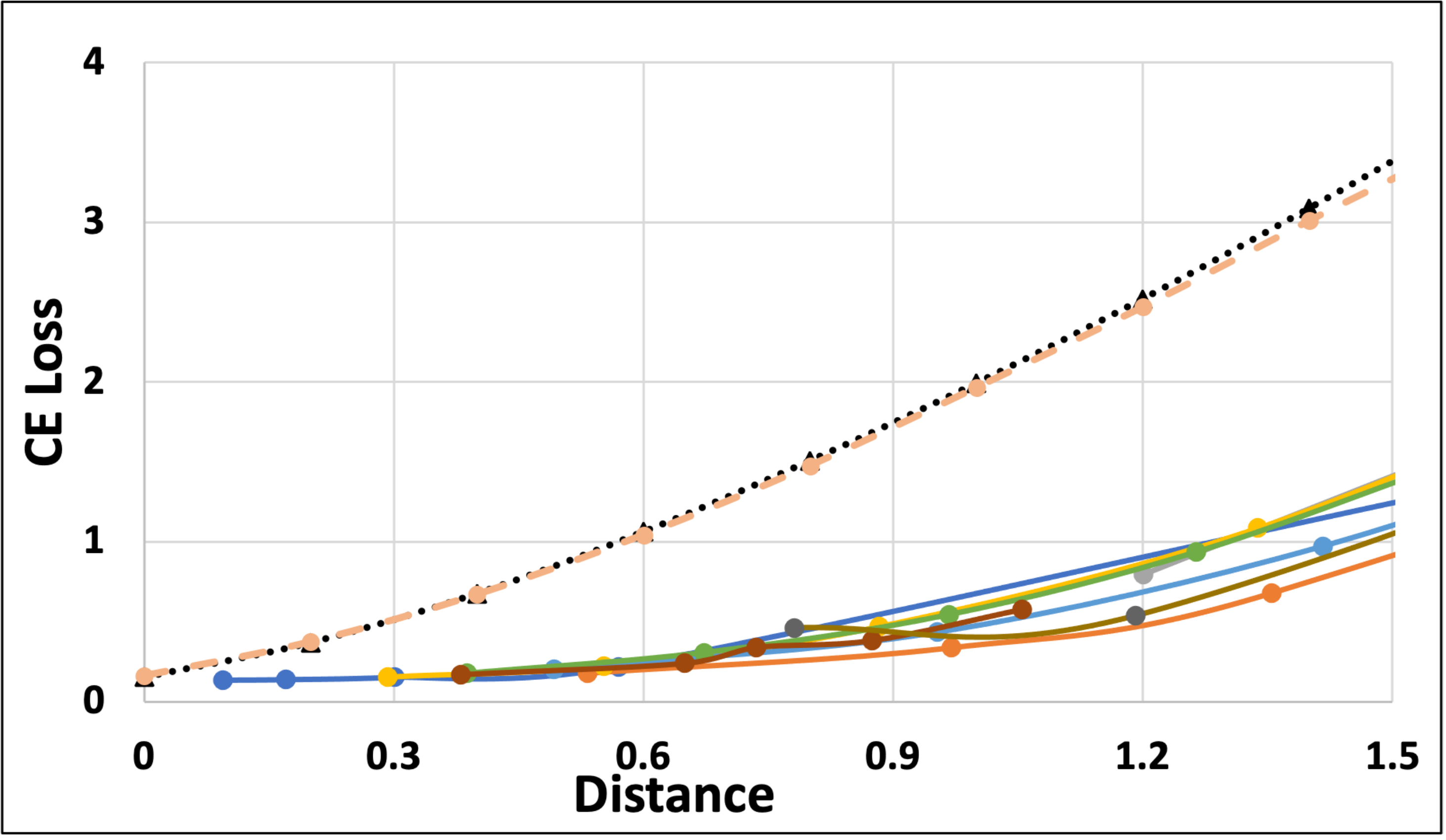}}
  
  \caption{(Best viewed in color.) Comparison of the certified (worst-case) loss computed in the representation space (by using Eq.~\ref{eq:rep-surrogate} as the surrogate loss) versus the certified loss computed in the input space (by using Eq.~\ref{eq:surrogate_loss} as the surrogate loss) with WM and G2DM trained on R-MNIST dataset.}
  \label{fig:input_vs_rep_space_certification}
\end{figure}

A potential disadvantage of our certification method is that it considers the uncertainty set to be distributions in the representation space. This may overestimate the worst-case loss since this space can have a worst-case distribution that is not a push-forward distribution, i.e., one that is not generated through the representation map.
Considering only push forward distributions is possible by adding additional constraints to our uncertainty set but it significantly increases the computational complexity of both Cert-DG and DR-DG since the maximizer of the robust surrogate must be computed through the representation map (Eq.~\ref{eq:surrogate_loss}). 
Although solving Eq.~\ref{eq:surrogate_loss}, restricts the worst-case distribution to be the push forward distributions, we find the worst-case loss to be very similar in both cases, as shown in Fig.~\ref{fig:input_vs_rep_space_certification}. 
The gap between the worst-case losses computed becomes even smaller when the models are trained by DR-DG. This small gap demonstrates that our representation space based certification is not significantly overestimating the loss of the worst-case distribution compared to the input space certification. 
We thus prefer solving Eq.~\ref{eq:rep-surrogate} due to its ease of computation compared to Eq.~\ref{eq:surrogate_loss} which gives us the ability to use Cert-DG and DR-DG on even large-scale problems.

\subsection{Point-wise adversarial robustness versus distributional robustness}
\label{app:discussion_pointwise}
Point-wise adversarial robustness focuses on ensuring the predictions of a data point remains constant when it is perturbed by a specific amount (usually within an $\ell_p$ ball). 
In contrast, distributional robustness deals with how the performance of the models changes on average when faced with a distribution different from the one used during training. 
Specifically, distributional robustness does not focus on how the model behaves for a specific point from the distribution but rather focuses on how the model behaves on average on points from a distribution.
In this context, the worst-case and average-case point-wise robustness of the models can be evaluated using the following objectives
\[
\mathbb{E}_{(x,y) \in Q}[\max_{\delta \in \Delta(x)} [\ell(h(g(x+\delta)), y)]] \; \mathrm{and} \; \mathbb{E}_{(x,y) \in Q}[\mathbb{E}_{\delta \in T(x)} [\ell(h(g(x+\delta)), y)]]
\]
where $\Delta(x)$ denotes the permissible perturbations set around a point $x$, $T(x)$ is the distribution of permissible perturbations and $Q$ is the source distribution. 
Notice the above problems are different from the problem in Eq.~\ref{Eq:dro}, which focuses on generating a worst-case distribution in the Wasserstein ball around the source. 
The above problems can be approximately solved using PGD \cite{madry2017towards} attack and by evaluating the model's performance on known transformations in $T(x)$ but solving for the worst-case distribution as required in Eq.~\ref{Eq:dro} is not possible directly since the problem is infinite-dimensional. 
Methodology to solve the problem in Eq.~\ref{Eq:dro} is the focus of our work and has been discussed extensively in Sec.~\ref{sec:certification}.
Moreover, our work focuses on solving the problem certifiably and providing a guarantee on the performance of the model on the worst-case distribution in the Wasserstein ball around the source distribution in the representation space. 
Thus, our work can be considered as a distributional analog of point-wise certified robustness works such as those based on randomized smoothing or interval bound propagation (for adversarial robustness) which aim to provide a certified radius within which performance of the model remains constant (for that point).

Our certification algorithm, Cert-DG, aims to compute the loss of the worst-case distribution in the Wasserstein ball in the representation space. 
Since we consider the distribution in the representation space, we are able to solve the robust surrogate loss in Eq.~\ref{eq:rep-surrogate}, without requiring any prohibitive assumptions, which are needed if the worst-case distribution is considered in the input space \cite{sinha2017certifying}.
Similar to point-wise certification guarantees \cite{cohen2019certified}, models trained with empirical risk minimization (ERM) achieve small certified distributional robustness which implies that the loss of the worst-case distribution even close to the source distribution can be high. 
However, this gap does not imply that the certification results are not tight but shows the existence of distributions that could lead to a much higher loss.
Although, by imposing additional assumptions on the target distributions it may be possible to obtain better certification results but it would limit the kind of distribution shift that the model is certified for and is less general than the setting we consider. 
To make models provably more robust than the models trained with ERM, we proposed the DRO-based training approach, DR-DG.
The point-wise analog of which are approaches proposed in \cite{salman2019provably, zhai2020macer} which provided new training methods that can make models certifiably more robust.
Similar to their results we observe that DR-DG lowers the worst-case loss of the models under distribution shift without significantly lowering the empirical performance of the models.
Thus, making the models provably robust to unseen distributions which is crucial for enabling their deployment in critical applications.

%%%%%%%%%%%%%%%%%%%%%%%%%%%%%%%%%%%%%%%%%%%%%%%%%%%%%%%%%%%%%%%%%%%%%%%%
\section{Experimental details}
\label{app:experimental_details}
%%%%%%%%%%%%%%%%%%%%%%%%%%%%%%%%%%%%%%%%%%%%%%%%%%%%%%%%%%%%%%%%%%%%%%%%
All codes are written in Python using Tensorflow/Keras and were run on Intel Xeon(R) W-2123 CPU with 64 GB of RAM and dual NVIDIA TITAN RTX. Dataset details and model architectures used are described below.

\subsection{Dataset description}
We have used three popular DG benchmark datasets in our work as described below.

{\bf Rotated MNIST (R-MNIST) \cite{ghifary2015domain}:} This is a variant of the popular MNIST dataset where domains are created by rotating images at different angles. 
The rotation angles present in the datasets are $\{0, 15, 30, 45, 60, 75\}$. 
Domains in our dataset comprise of 2000 images from each rotation angle. 
We use 1000 images from the domain for our model training and reserve the other 1000 images for testing. 
For evaluating the performance of models trained with DG we use the images from the test set (although for unseen domains distinguishing training and test set is not required). 

{\bf PACS \cite{li2017deeper}:} This dataset contains images for different styles from four domains Art, Cartoons, Photos, and Sketches. 
It contains 9991 images belonging to 7 different classes.
For evaluating the performance of DG methods on source domains we hold out 10\% of the data from each source domain.
However, for evaluating performance on the unseen domains use all the data from those domains.

{\bf VLCS \cite{fang2013unbiased}:}
This dataset contains images from four domains Caltech101, LabelMe, SUN09,
VOC2007.
It contains 10729 images belonging to 5 different classes.
Similar to PACS, we evaluate the performance of DG methods on source domains using a held-out dataset comprising 10\% of the data from each source domain and use all the data from unseen domains for evaluating the performance of the models. 

\subsection{Model description}
The representation network for R-MNIST uses a convolutional neural network similar to the one described in Table 7 \cite{gulrajani2020search} and we fine-tune a Resnet-50 model for PACS and VLCS.
Additionally, for PACS and VLCS we add 2 additional fully connected layers on top of the 512-dimensional output of the Resnet-50 to reduce the size of the output dimension to 128.
This is needed since estimating Wasserstein distance in a high dimensional space required a large number of samples. Due to the limited size of these datasets and even smaller size of the test sets, we reduce the dimension to estimate the Wasserstein distance better.
For discriminators used in G2DM and CDAN, we use two fully connected layers and feed in the output of the representation network. For G2DM we do not use the random projection layers in the network and for CDAN we use multi-linear conditioning.
Our classifier just comprises of a fully connected layer on top of the representation network.

\subsection{Corruptions used in Figs.~\ref{fig:high_variability_of_dg_a} and~\ref{fig:high_variability_of_dg_b}}
\label{app:fig_1_corruptions}
To evaluate the performance of models trained with DG methods on unseen domains in relation to their distance from the source domains, we chose to add common corruptions to the test set of the source domain data.
The ability to change the severity of the corruption allows us to create multiple unseen domains at different distances from the sources. 
We emphasize that the use of common corruptions is just an easy and efficient way of generating multiple unseen domains and only covers a small subset of all possible unseen domains.
Moreover, our task here is not to show that the performance of models trained with different DG methods deteriorates on these corrupted domains but it is rather to highlight the high variability in their performance even when distributions lie at the same distance. 
For R-MNIST we use the corruptions from the MNIST-C dataset \cite{mu2019mnist}  (\texttt{shot noise, impulse noise, glass blur, shear, scale, translate, fog, spatter, elastic transform}) and for PACS and VLCS we use corruptions from the Imagenet-C dataset \cite{hendrycks2018benchmarking} (\texttt{Gaussian noise, shot noise, impulse noise, speckle noise, Gaussian blur, defocus blur, zoom blur, spatter, contrast, brightness, saturate, elastic transform}).

\subsection{PGD adversarial attacks}
\label{app:adversarial_attacks}
We consider adversarial attacks using PGD \cite{madry2017towards} to demonstrate the existence of distributions beyond common corruptions which can degrade the performance of DG methods as predicted by our certification in Fig.~\ref{fig:certification_vanilla_wm_rotatedmnist}.
We consider two variants of this attack. 
The first is the usual PGD attack where the adversarial examples are crafted in the input space by maximizing the loss of the perturbed examples (denoted by the grey dashed line with square markers in Fig.~\ref{fig:certification_vanilla_wm_rotatedmnist}). The distance between the adversarial example and the clean example is measured in the input space in this case.
The second variant is the PGD attack created in the representation space. 
In this attack, the distortion between the clean and adversarial examples is measured in the representation space and the perturbation is applied to the representation of the clean examples, i.e., \(\max_{z'} \ell(h(z'), y) \; \mathrm{s.t.} \;\|z' - z\|_2 \leq \epsilon_g\) where $z'$ is the adversarial example in the representation space for the point $(z=g(x),y)$ and $\epsilon_g$ is the bound on the distortion in the representation space. 
The result of this attack is shown as the green dashed line in Fig.~\ref{fig:certification_vanilla_wm_rotatedmnist}).

\subsection{Implementation details of Cert-DG and DR-DG}
Here we briefly describe some of the implementation details for our certification and training algorithms.
Cert-DG requires to compute distributions which have an average distortion of $\rho^2$. 
Solving the certification problem requires maximization of the robust surrogate loss Eq.~\ref{eq:rep-surrogate} for every point.
We solve the problem batch-wise and keep track of the perturbation applied to the points. 
Since the network parameters are fixed during certification, using the previously saved perturbation as the starting point helps the algorithm converge faster with minimal tuning of the hyperparameters such as the learning rates in Alg.~\ref{alg:certification}.
%Thus creating distributions perturbed on average by $\rho^2$.
For DR-DG, since the representation space changes we find that keeping track of the perturbations is not as helpful for some DG methods specially on PACS and VLCS where the representation is fine-tuned starting from a pre-trained ResNet-50 models.
Thus, for G2DM, CDAN and VREX, we opt to increase the number of maximization steps per batch rather than using the previous perturbation on VLCS and PACS. 
On the other hand, for WM on these datasets we use a fewer maximization steps and keep track of the perturbations since we empirically found that the representation changes gradually when trained WM and using the previous perturbations help generate distributions perturbed by $\rho^2$ just as in Cert-DG. 
Due to the simplicity of the R-MNIST dataset we do not observe any difference in behavior of the algorithms while training with DR-DG and thus we chose to keep track of the perturbations applied to the points to solve the problem more efficiently. 
The main advantage of keeping track of the perturbation is that it helps eliminate the problem of doing a large number of steps to generate distributions with perturbation $\rho^2$ which can slow down DR-DG training.

Our distance computations and DR-DG algorithm rely on estimating the adversarial distribution in the representation space. 
We rely on the Cleverhans implementation of the CW-attack \cite{carlini2017towards} for generating this distribution.
We use 1000 points from the test set and compute the adversarial examples for these points and measure $\rho_{adv}$ (in Fig.~\ref{fig:method_explanation}) using the original and the perturbed points. 
Since we work in the representation space estimating the adversarial distribution is computationally easy and we require approximately 2 minutes for computing points which achieve greater than 95\% attack success on models trained with vanilla DG methods.

\end{document}